\documentclass[10pt]{article}
\usepackage{graphicx}
\usepackage{amsmath}
\usepackage{amssymb}
\usepackage{algorithmic}
\usepackage{algorithm}

\DeclareMathOperator*{\argmin}{arg\,min}
\numberwithin{equation}{section}
\usepackage{fancyhdr} 
\author{Brian K. Vogel \\
brian@brianvogel.com}

\title{Positive factor networks: A graphical framework for modeling non-negative sequential data}

\begin{document}
\maketitle

\begin{abstract}
We present a novel graphical framework for modeling non-negative sequential data with hierarchical structure. Our model corresponds to a network of coupled non-negative matrix factorization (NMF) modules, which we refer to as a positive factor network (PFN). The data model is linear, subject to non-negativity constraints, so that observation data consisting of an additive combination of individually representable observations is also representable by the network. This is a desirable property for modeling problems in computational auditory scene analysis, since distinct sound sources in the environment are often well-modeled as combining additively in the corresponding magnitude spectrogram. We propose inference and learning algorithms that leverage existing NMF algorithms and that are straightforward to implement. We present a target tracking example and provide results for synthetic observation data which serve to illustrate the interesting properties of PFNs and motivate their potential usefulness in applications such as music transcription, source separation, and speech recognition. We show how a target process characterized by a hierarchical state transition model can be represented as a PFN. Our results illustrate that a PFN which is defined in terms of a single target observation can then be used to effectively track the states of multiple simultaneous targets. Our results show that the quality of the inferred target states degrades gradually as the observation noise is increased. We also present results for an example in which meaningful hierarchical features are extracted from a spectrogram. Such a hierarchical representation could be useful for music transcription and source separation applications. We also propose a network for language modeling.

\end{abstract}

\section{Introduction}

We present a graphical hidden variable framework for modeling non-negative sequential data with hierarchical structure. Our framework is intended for applications where the observed data is non-negative and is well-modeled as a non-negative linear combination of underlying non-negative components. Provided that we are able to adequately model these underlying components individually, the full model will then be capable of representing any observed additive mixture of the components due to the linearity property. This leads to an economical modeling representation, since a compact parameterization can explain any number of components that combine additively. Thus, in our approach, we do not need to be concerned with explicitly modeling the maximum number of observed components nor their relative weights in the mixture signal.

To motivate the approach, consider the problem of computational auditory scene analysis (CASA), which involves identifying auditory ``objects'' such as musical instrument sounds, human voice, various environmental noises, etc, from an audio recording. Speech recognition and music transcription are specific examples of CASA problems. When analyzing audio, it is common to first transform the audio signal into a time-frequency image, such as the spectrogram (i.e., magnitude of the short-time Fourier transform (STFT)). We empirically observe that the spectrogram of a mixture of auditory sources is often well-modeled as a linear combination of the spectrograms of the individual audio sources, due to the sparseness of the time-frequency representation for typical audio sources. For example, consider a recording of musical piece performed by a band. We empirically observe that the spectrogram of the recording tends to be well-approximated as the sum of the spectrograms of the individual instrument notes played in isolation. If one could construct a model using our framework that is capable of representing any individual instrument note played in isolation, the model would then automatically be capable of representing observed data corresponding to arbitrary non-negative linear combinations of the individual notes. Likewise, if one could construct a model under our framework capable of representing a recording of a single human speaker (possibly including a language model), such a model would then be capable of representing an audio recording of multiple people speaking simultaneously. Such a model would have obvious applications to speaker source separation and simultaneous multiple-speaker speech recognition. We do not attempt to construct such complex models in this paper, however. Rather, our primary objective here will be to construct models that are simple enough to illustrate the interesting properties of our approach, yet complex enough to show that our approach is noise-robust, capable of learning from training data, and at least somewhat scalable. We hope that the results presented here will provide sufficient motivation for others to extend our ideas and begin experimenting with more sophisticated PFNs, perhaps applying them to the above-mentioned CASA problems. 

An existing area of research that is related to our approach is non-negative matrix factorization (NMF) and its extensions. NMF is a data modeling and analysis tool for approximating a non-negative matrix $X$ as the product of two non-negative matrices $W$ and $H$ so that the reconstruction error between $X$ and  $W H$ is minimized under a suitable cost function. NMF was originally proposed by Paatero as \emph{positive matrix factorization} \cite{paatero_1994}. Lee and Seung later developed robust and simple to implement multiplicative update rules for iteratively performing the factorization \cite{Lee_seung}. Various sparse versions of NMF have also been recently proposed \cite{hoyer_sparse_NMF}, \cite{cichockiNMF}, \cite{nsNMF2006}. NMF has recently been applied to many applications where a representation of non-negative data as an additive combination of non-negative basis vectors seems reasonable. Such applications include object modeling in computer vision, magnitude spectra modeling of audio signals \cite{wang_mag_spectrogram_NMF}, and various source separation applications \cite{NMFParis}. The non-negative basis decomposition provided by NMF is, by itself, not capable of representing complex model structure. For this reason, extensions have been proposed to make NMF more expressive. Smaragdis extended NMF in \cite{NMFParis} to model the temporal dependencies in successive spectrogram time slices. His NMF extension, which he termed \emph{Convolutive NMF}, also appears to be a special case of one of our example models in Section~\ref{sec:sparseHierarchicalModel}.  We are unaware of any existing work in the literature that allow for the general graphical representation of complex hidden variable models, particularly sequential data models, that is provided by our approach, however. 

A second existing area of research that is related to our approach is probabilistic graphical models \cite{jordan_graphical_models} and in particular, dynamic Bayesian networks (DBNs) \cite{Murphy_Thesis}, \cite{Friedman1999}, which are probabilistic graphical models for sequential data. We note that the Hidden Markov Model (HMM) is a special case of a DBN. DBNs are widely used for speech recognition and other sequential data modeling applications. Probabilistic graphical models are appealing because they they can represent complex model structure using an intuitive and modular graphical modeling representation. A drawback is that the corresponding exact and/or approximate inference and learning algorithms can be complex and difficult to implement, and overcoming tractability issues can be a challenge.

Our objective in this paper is to present a framework for modeling non-negative data that retains the non-negative linear representation of NMF, while also supporting more structured hidden variable data models with a graphical means for representing variable interdependencies analogously to that of the probabilistic graphical models framework. We will be particularly interested in developing models for sequential data consisting of the spectrograms of audio recordings. Our framework is essentially a modular extension of NMF in which the full graphical model corresponds to several coupled NMF sub-models. The overall model then corresponds to a system of coupled vector or matrix factorization equations. Throughout this paper, we will refer to a particular system of factorizations and the corresponding graphical model as a \emph{positive factor network (PFN)}. We will refer to the dynamical extension of a PFN as a \emph{dynamic positive factor network (DPFN)}. Given an observed subset of the PFN model variables, we define inference as solving for the values of the hidden subset of variables and learning as solving for the model parameters in the system of factorization equations. 

Note that our definition of inference is distinct from the probabilistic notion of inference. In a PFN, inference corresponds to solving for actual values of the hidden variables, whereas in a probabilistic model inference corresponds to solving for probability distributions over the hidden variables given the values of the observed variables. Performing inference in a PFN is therefore more analogous to computing the MAP estimates for the hidden variables in a probabilistic model. One could obtain an analogous probabilistic model from a PFN by considering the model variables to be non-negative continuous-valued random vectors and defining suitable conditional probability distributions that are consistent with the non-negative linear variable model. Let us call this class of models \emph{probabilistic PFNs}.  Exact inference is generally intractable in such a model since the hidden variables are continuous-valued and the model is not linear-Gaussian. However, one could consider deriving algorithms for performing approximate inference and developing a corresponding EM-based learning algorithm. We are unaware of any existing algorithms for performing tractable approximate inference in a probabilistic PFN. It is possible that our PFN inference algorithm may also have a probabilistic interpretation, but exploring the idea further is outside the scope of this paper. Rather, in this paper our objective is to to develop and motivate the inference and learning algorithms by taking a modular approach in which existing NMF algorithms are used and coupled in a way that seems to be intuitively reasonable. We will be primarily interested in empirically characterizing the performance of the proposed inference and learning algorithms on various example PFNs and test data sets in order to get a sense of the utility of this approach to interesting real-world applications.

We propose general joint inference and learning algorithms for PFNs which correspond to performing NMF update steps independently (and therefore potentially also in parallel) on the various factorization equations while simultaneously enforcing coupling constraints so that variables that appear in multiple factorization equations are constrained to have identical values. Our empirical results show that the proposed inference and learning algorithms are fairly robust to additive noise and have good convergence properties. By leveraging existing NMF multiplicative update algorithms, the PFN inference and learning algorithms have the advantage of being straightforward to implement, even for relatively large networks. Sparsity constraints can also be added to a module in a PFN model by leveraging existing sparse NMF algorithms. We note that the algorithms for performing inference and learning in PFNs should be understandable by anyone with a knowledge of elementary linear algebra and basic graph theory, and do not require a background in probability theory. Similar to existing NMF algorithms, our algorithms are highly parallel and can be optimized to take advantage of parallel hardware such as multi-core CPUs and potentially also stream processing hardware such as GPUs. More research will be needed to determine how well our approach will scale to very large or complex networks.

The remainder of this paper has the following structure. In Section~\ref{sec:pfn_main_section}, we present the basic PFN model. In Section~\ref{sec:main_factored_model}, we present an example of how a DPFN can be used to represent a transition model and present empirical results. In Section~\ref{sec:main_hierarchical_state_model}, we present an example of using a PFN to model sequential data with hierarchical structure and present empirical results for a regular expression example. In Section~\ref{sec:target_tracking}, we present a target tracking example and provide results for synthetic observation data which serve to illustrate the interesting properties of PFNs and motivate their potential usefulness in applications such as music transcription, source separation, and speech recognition. We show how a target process characterized by a hierarchical state transition model can be represented by a PFN. Our results illustrate that a PFN which is defined in terms of a single target observation can then be used to effectively track the states of multiple simultaneous targets in the observed data. In Section~\ref{sec:hierarchicalSeqDecomp} we present results for an example in which meaningful hierarchical features are extracted from a spectrogram. Such a hierarchical representation could be useful for music transcription and source separation applications. In Section~\ref{sec:main_language_model}, we propose a DPFN for modeling the sequence of words or characters in a text document as an additive factored transition model of word features. We also propose slightly modified versions Lee and Seung's update rules to avoid numerical stability issues. The resulting modified update rules are presented in Appendix~\ref{appendix_nmf}.

\section{Positive Factor Networks}
\label{sec:pfn_main_section}

In this section, we specify the basic data model and present a graphical representation. We then propose inference algorithms for solving for the hidden variables and learning the model parameters. 

\subsection{Model specification}
\label{sec:model_specification}

We now specify a data model for a set of non-negative continuous vector-valued variables $\{x_i: i = 1, \dots, N\}$ where the dimension of each $x_i$, in general, can be distinct. We will refer to the set $\{x_i\}$ as the \emph{model variables}. We assume that a subset $X_E$ of the model variables is observed and the rest of the variables comprise the hidden subset, $X_H$. 

The model is specified by a system of $Q$ of non-negative factorization equations where the $j$'th equation is given by:

\begin{align}
x_{f(j,0)} =& \sum_{k=1}^{P_j} W^j_k x_{f(j,k)} \notag \\
=&  \left[ \begin{array}{cccc} W^j_1 & W^j_2 & \dots & W^j_{P_j} \end{array} \right] \left[ \begin{array}{c} x_{f(j,1)}\\
x_{f(j,2)} \\
\vdots \\
x_{f(j,{P_j})} \end{array} \right] \notag \\
=& W^j x^j
\label{eq:factorization_equation}
\end{align}

where $P_j \geq 1$ for all $j \in \{1, \dots, Q\}$ and all $W^j_k$ are non-negative matrices with a possibly distinct number of columns for each $k$. The function $f(j,k)$ maps each $(j,k)$ above into a corresponding index $i \in \{1, \dots, N\}$ and satisfies $f(j,k) = f(m,n)$ implies $j \ne m$. Thus, $x_{f(j,k)}$ refers to one of the model variables $x_i$, and it is possible for a given variable $x_i$ to appear in multiple equations above but only at most once in any given equation. The matrix $W^j$ is defined as the horizontal concatenation of the non-negative $W^j_k$ matrices.

Since there are no constant terms in the above equations, the system corresponds to a a homogeneous system of linear equations, subject to a non-negativity constraint on the variables \{$x_i$\}, which can be written as:

\begin{align}
A y = 0 \text{ , where } y = \left[ \begin{array}{c} x_1\\
x_2 \\
\vdots \\
x_N \end{array} \right] \geq 0
\end{align}

Note that $A$ will contain both negative and positive values, however, since all of the terms are grouped together on one side. It follows that our model satisfies the linearity property subject to non-negativity constraints (non-negative superposition property): if $y_1$ and $y_2$ are solutions to the system, then $\alpha_1 y_1 + \alpha_2 y_2$ is also a solution, for any choice of scalars $\alpha_1 > 0, \alpha_2 > 0$.

\subsection{Graphical representation}
We now develop a graphical representation to facilitate the visualization of the local linear relationships between the model variables in the various factorization equations. Hidden variables correspond to shaded nodes in the graph, and observable variables correspond to unshaded nodes. We construct the graphical model such that the $j$'th factorization equation corresponds to a subgraph with $x_{f(j,0)}$ as the child node and the $\{x_{f(j,k)} : k = 1, \dots, P_j\}$ as the parent nodes. Thus, a child node variable is a linear function its parent nodes in the subgraph. We then arrive at the complete graphical model by superimposing the subgraphs corresponding to the various factorization equations. We allow the same variable $x_i$ to correspond to the child node in multiple subgraphs (i.e., distinct factorization equations), provided that its parents do not overlap between the subgraphs. That is, we do not allow a single arc to correspond to multiple linear relationships. We annotate arcs with dash marks where necessary, to disambiguate subsets of parent nodes that correspond to the same factorization equation. Arcs annotated with the same number of dash marks connecting a child node to its parents denote a subset of parent nodes that corresponds to a single factorization equation. The pseudocode in Algorithm~\ref{alg:create_graph} outlines a procedure for creating a graphical model representation from a system of factorization equations. 

\begin{algorithm}
\caption{Create a directed graph from a system of factorizations.}
\label{alg:create_graph}
\begin{tabbing}
{\bf for} \= {$j$ = 1 to $Q$}  \\
\> // for each factorization equation $x_{f(j,0)} = \sum_{k=1}^{P_j} W^j_k x_{f(j, k)}$ \\
\> Create a node for the corresponding variable $x_{f(j, 0)}$, if it does not already exist. \\
\> $dashCount \gets$ 1 + maximum dash count on any existing arc from a parent node of $x_{f(j, 0)}$ to $x_{f(j, 0)}$. \\
\> {\bf for} \= {$k$ = 1 to $P_j$} \\
\> \> {\bf if} \= a node corresponding to $x_{f(j, k)}$ does not already exist \\
\> \> \> Create a node corresponding to $x_{f(j, k)}$ \\
\> \> \>  {\bf if} \= $x_{f(j, k)}$ is observed \\
\> \> \> \> Shade the node corresponding to $x_{f(j, k)}$ \\
\> \> \> {\bf end} \\
\> \> {\bf end} \\
\> \> Create a directed arc from the $x_{f(j,k)}$ node to the $x_{f(j, 0)}$ node. \\
\> \> Annotate the arc with the number of dash marks given by the current value of $dashCount$. \\
\> \> Annotate the arc with $W^j_k$ \\
\> {\bf end} \\
{\bf end} 
\end{tabbing}
\end{algorithm}

As an example, consider the set of variables \{$x_1, x_2, \dots, x_{12}$\} that are described by the following system of factorization equations:

\begin{align}
\label{eqn:factor1}
x_1 =& W_1 x_5 + W_2 x_6 \\
\label{eqn:factor2}
x_2 =& W_3 x_6 + W_4 x_7 \\
\label{eqn:factor3}
x_2 =& W_5 x_8 + W_6 x_9 \\
\label{eqn:factor4}
x_2 =& W_7 x_{10} \\
x_3 =& W_8 x_{10} \\
\label{eqn:factor6}
x_3 =& W_9 x_{11} \\
\label{eqn:factor7}
x_4 =& W_{10} x_{11} \\
x_6 =& W_{11} x_{12} \\
x_{10} =& W_{12} x_{12} \\
\label{eqn:factor10}
x_{11} =& W_{13} x_{12}
\end{align}

\begin{figure}
\centering
\includegraphics[width=60ex]{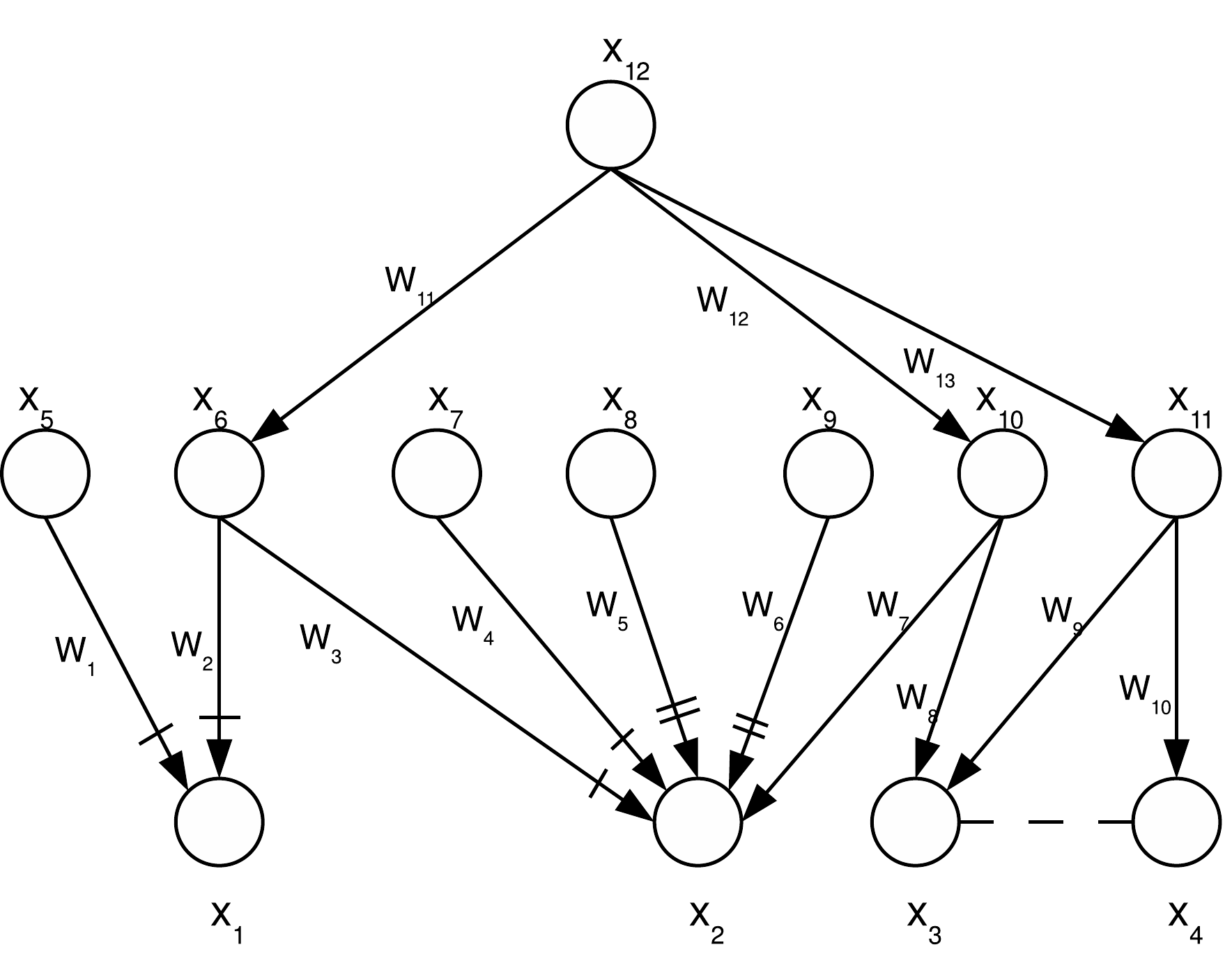}
\caption{A graphical model corresponding to the example system of factorizations in Equations~\ref{eqn:factor1} - \ref{eqn:factor10}.}
\label{fig:sampleModel1}
\end{figure}

Figure~\ref{fig:sampleModel1} shows the graphical model associated with the above system of factorizations. Note that nodes $x_3$ and $x_4$ are connected by a dashed line. This is an optional annotation that specifies that the corresponding connected variables have \emph{forced factored co-activations}, or simply \emph{forced co-activations}. So far the only constraint we have placed on the parameter matrices \{$W_i$\} is that they be non-negative. However, in some models we might wish to place the additional constraint that the columns $\{w_n\}$ of $W_i$ are normalized in some way. For example, we might consider requiring that each $w_n$ have column sum = 1. Consider the subgraph corresponding to Equations (\ref{eqn:factor6}) and (\ref{eqn:factor7}). The dashed line connecting $x_3$ and $x_4$ specify that these variables are factorized in terms of a common parent ($x_{11}$ in this case) and that the columns of the corresponding \{$W_i$\} ($W_9$ and $W_{10}$ in this case) are normalized to have unit sum. This constraint ensures that these three variables will have equal column sums. Thus, if any one of these variables is observed in the model, then we can infer that all three must have the same column sum. An activation of the parent $x_{11}$ implies that its children $x_3$ and $x_4$ will then be activated with the same column sum (activated together), hence the term forced co-activations.

Consider the subgraph corresponding to Equation (\ref{eqn:factor1}). An activation in $x_1$ (i.e., $x_1$ is nonzero) could be explained by just one of the parents being nonzero. Now consider the subgraph consisting of $x_2$ and its parents, corresponding to Equations (\ref{eqn:factor2}),  (\ref{eqn:factor3}), and (\ref{eqn:factor4}). In this case, an activation of $x_2$ corresponds to at least one of $x_6$ and $x_7$ being activated, at least one of $x_8$ and $x_9$ being activated, as well as $x_{10}$ being activated.

\begin{figure}
\centering
\includegraphics[width=25ex]{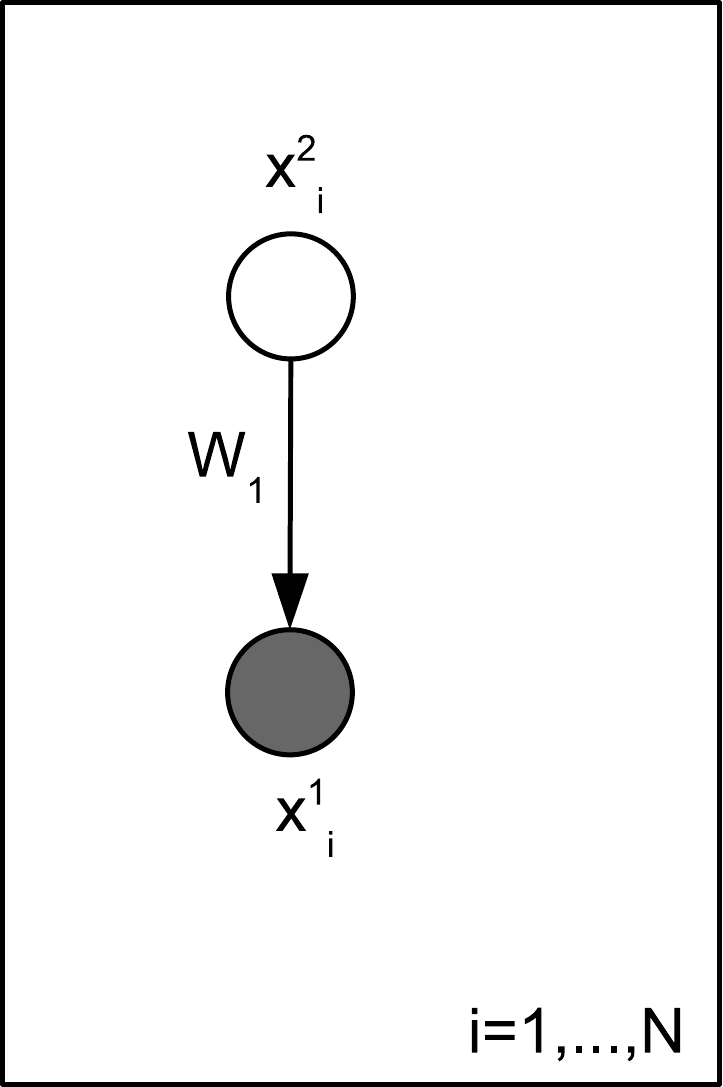}
\caption{An example of using plate notation, corresponding to $N$ observable variables \{$x^1_i$\} and $N$ hidden variables \{$x^2_i$\}. This model corresponds to standard NMF.}
\label{fig:simpleNmfPlate}
\end{figure}

We borrow the notion of a \emph{plate} from the probabilistic graphical models formalism \cite{jordan_graphical_models}, in which plates are used to represent replicated random variables. A plate consists of a box that is drawn around the replicated variables, with the replication count specified in a corner. Figure~\ref{fig:simpleNmfPlate} shows an example of using the graphical plate notation. This graphical model corresponds to the following system of factorization equations:

\begin{align}
\label{eqn:nmf_vector}
x^1_1 =& W_1 x^2_1 \notag \\
x^1_2 =& W_1 x^2_2 \notag \\
\vdots \notag \\
x^1_N =& W_1 x^2_N
\end{align}

Letting $X^1 = \left[ \begin{array}{cccc} x^1_1 & x^1_2 & \dots & x^1_N \end{array} \right]$, and letting $X^2 = \left[ \begin{array}{cccc} x^2_1 & x^2_2 & \dots & x^2_N \end{array} \right]$, we can then write the system of factorization equations more compactly as the matrix equation:

\begin{align}
\label{eqn:nmf_matrix}
X^1 = W_1 X^2
\end{align}

Note that this corresponds to standard NMF, since the observed $X^1$ matrix is factored as the product of two non-negative matrices.

\subsection{Algorithms for Inference and Learning}
\label{sec:inference_and_learning}

Typically, a subset of the model variables $X_E = \{x_{E_n}: n = 1, \dots, N_E\}$ is observed, and we are interested in solving for the values of the hidden variables $X_H = \{x_{H_n}: n = 1, \dots, N_H\}$, which we refer to as the \emph{inference problem}. We are also interested in solving for the values of the model parameters $\theta = \{W^1, W^2, \dots, W^Q\}$, which we refer to as the \emph{learning problem}. The observed variables may deviate from the modeling assumptions and/or contain noise so that an exact solution will not be possible in general. We will thus seek an approximate solution by optimizing a reasonable cost function.

Given a subset of observable variables and the model parameters, we define the inference solution as the values of the hidden variables that minimize some reasonable cost function $g(\theta, X_H, X_E)$. That is, we consider $X_E$ and $\theta$ to be fixed and solve for $X_H$:

\begin{align}
X_H = \argmin_{X_H} g(\theta, X_H, X_E)
\end{align}

The learning problem corresponds to solving for $\theta$ given observations $X_E$. The joint learning and inference problem corresponds to minimizing $g(\theta, X_H, X_E)$ jointly for $\theta$ and $X_H$ given $X_E$: 

\begin{align}
(\theta, X_H) = \argmin_{\theta, X_H} g(\theta, X_H, X_E)
\end{align}

The cost function should have the property that its value approaches zero as the approximation errors of the system of factorization equations approach zero. One possibility is the following function, specified as the sum of the squared approximation errors of the factorization equations (\ref{eq:factorization_equation}):

\begin{align}
g(\theta, X_H, X_E) =& \sum_{j = 1}^Q ||x_{f(j,0)} -W^j x^j ||^2 \notag \\
=& \sum_{j = 1}^Q ||x_{f(j,0)} -W^j \left[ \begin{array}{c} x_{f(j,1)}\\
x_{f(j,2)} \\
\vdots \\
x_{f(j,{P_j})} \end{array} \right] ||^2 
\end{align}

Other possibilities could involve using the generalized KL-divergence (\ref{eq:kl_div}), for example. However, we will not develop inference and learning algorithms by directly optimizing any particular cost function, and so will not be too concerned with its exact form. Rather, we propose algorithms that seem intuitively reasonable and for which we have observed good empirical convergence properties on test data sets, but do not offer any proof of convergence, even to a local minimum of any particular cost function. We will only make use of a particular cost function in order to quantify the empirical performance of our algorithms.

We require that the graphical model correspond to a directed acyclic graph (DAG). Since there are no cycles, the graph can be arranged so that it flows from top to bottom. We then rename each of the model variables $\{x_n : n=1, \dots, N\}$ to $x^l_i$ where $l = 1,\dots, L$ denotes the level of the variable (vertical index), and $i$ is its position within the level (horizontal index). We then have a graph with $L$ levels, such that the top level nodes (level $L$) have no parents and the bottom level nodes (level 1) have no children. A level is defined such that no pair of nodes with a parent-child relationship are allowed to be in the same level. That is, for any pair of variables $(x^l_i, x^l_j), i \ne j$ in the same level $l$, we disallow that $x^l_i$ is the parent of $x^l_j$ or vice versa.  If an arc exists between a higher level variable and a lower level variable, then the higher level variable must be the parent of the lower level variable. That is, we require that for any two variables $(x^m_i, x^n_j), \text{ s.t }. m > n$, $x^m_i$ cannot be a child of $x^n_j$. For example, the graph in Figure~\ref{fig:sampleModel1} corresponds to an $L = 3$ level graph with the following renamed variables. Variables $(x_1, \dots, x_4)$ would be renamed to $(x^1_1, \dots, x^1_4)$. Variables $(x_5, \dots, x_{11})$ would be renamed to $(x^2_1, \dots, x^2_7)$. Variable $x_{12}$ would be renamed to $x^3_1$.

The inference and learning algorithm will require a set of \emph{local variables} \{$v^j_k$\}. We use the term local variable because a given $v^j_k$ is associated only with the $j$'th factorization equation, unlike a model variable $x_{f(j,k)}$ which is allowed to appear in multiple factorization equations. Specifically, a distinct $v^j_k$ will be associated with  each allowable combination of $j$ and $k$ that appears in the factorization equations in (\ref{eq:factorization_equation}). Thus, several distinct $v^j_k$ may be associated with a given model variable and this will be the case when a given model variable appears in more than one factorization equation. Replacing the $x_{f(j,k)}$ with the associated $v^j_k$, we then have $Q$ factorization equations $FactorSystem = \{eq_j : j = 1, \dots, Q\}$  where equation $eq_j$ is given by:

\begin{align}
v^j_0 =& \sum_{k=1}^{P_j} W^j_k v^j_k \notag \\
=&  \left[ \begin{array}{cccc} W^j_1 & W^j_2 & \dots & W^j_{P_j} \end{array} \right] \left[ \begin{array}{c} v^j_1\\
v^j_2 \\
\vdots \\
v^j_{P_j} \end{array} \right] \notag \\
=& W^j v^j
\label{eq:factorization_equation333}
\end{align}

We say that the above equations are in a \emph{consistent state} if each model variable $x_{f(j,k)}$ and all of its associated local variables \{$v^j_k$\} have the same value. Otherwise, we say that the equations are in an \emph{inconsistent state}. Note being in a consistent state does not imply that the corresponding $x_{f(j,k)}$ actually constitute a solution to the system.

The basic idea of our approach is to learn a generative model of data by iteratively performing inference and learning in a bottom-up pass through the network and then perform data generation through activation propagation in a top-down pass. In the inference and learning pass, parent node values (activations) and parameters are updated based on the values of their child node variables. These inference and learning updates are performed locally using NMF algorithms. Once the top level nodes have been updated, we then propagate values downward to the lowest level nodes in a data generation pass. This is performed by computing the new child node values as the value of the right hand side of the corresponding factorization equations in which they appear. Throughout this process, the multiple $v^j_k$ variables that correspond to a single model variable $x_{f(j,k)}$ are repeatedly replaced by their mean value in order to put the system of factorization equations back into a consistent state. This process of a bottom-up inference and learning step followed by a top-down value propagation (data generation) step is iterated until convergence. 

Algorithm~\ref{alg:inference_learning1} and the corresponding procedures in Algorithm~\ref{alg:upStepDownStep} and Algorithm~\ref{alg:averagingProcedures} show the pseudocode for the basic inference and learning algorithm that was used to obtain all of the empirical results in this paper. We start by initializing the hidden variables $X_H$ to small random positive values. We then make the system consistent by copying the value of each model variable $x_{f(j,k)}$ into each of its corresponding local variables $v^j_k$. The only distinction between hidden and observed variables from the perspective of the learning and inference algorithm is that model variables in the observed set $X_E$ are never modified in the algorithm. In computing the mean values of the variables, the new mean for the model variables is computed only from the subset of the local variables that were updated in the corresponding inference or value propagation step. After performing the value propagation step, the updated values of all lower level variables in the model are a function of the top level variables.

When a parameter matrix $W$ is tied across multiple vector factorization equations, the corresponding equations can be merged into a single matrix factorization equation, as we did in going from (\ref{eqn:nmf_vector}) to (\ref{eqn:nmf_matrix}). This merging will also be possible in the dynamic models that we present starting in Section~\ref{sec:main_factored_model}. In this case, the learning update of the upStep() procedure is simply modified to perform a single NMF left update instead of multiple left up steps. Appendix~\ref{appendix_inf_learn_2} describes a similar inference and learning algorithm in which the mean values of the variables are computed differently.

\begin{algorithm}
\caption{Perform inference and learning}
\label{alg:inference_learning1}
\begin{tabbing}
Initialize hidden variables to random positive values \\
// Main loop \\
{\bf repe}\={\bf at}  \\
\> // Bottom-to-top inference and learning \\
\> {\bf for} \= $l$ = 1 to $L -1$ \\
\> \> upStep($l$) \\
\> \> averageParents($l$) \\
\> {\bf end} \\
\> // Top-to-bottom value propogation \\
\> {\bf for} $l$ = $L -1$ downto 1 \\
\> \> downStep($l$) \\
\> \> averageChildren($l$) \\
\> {\bf end} \\
{\bf until} convergence
\end{tabbing}
\end{algorithm}

\begin{algorithm}
\caption{upStep() and downStep() procedures}
\label{alg:upStepDownStep}
\begin{tabbing}
// Using the values of the child variables at level $l$, update the values of the parent variables and \\
// update the parameter matrices by performing NMF update steps. \\
// Let $X_l$ denote the set of model variables \{$x^l_1, x^l_2, \dots, x^l_{LevelCount_l}$\} corresponding to the level $l$ nodes.\\
// Let $FactorSystem_l$ denote the subset of the factorization equations from (\ref{eq:factorization_equation333})\\
// \{$eq_j: $ such that $v^j_0$ in $eq_j$ corresponds to an $x^l_i \in X_l$\}.  \\
// Let $duplicationSetChild(i,l)$ = \{$j : eq_j \in FactorSystem_l$ and $v^j_0$ corresponds to $x^l_i$\}. \\
{\bf upSt}\={\bf ep}($l$) \\
\> {\bf for} \={\bf each} $j \in FactorSystem_l$ \\
\> \> {\bf if} \= learning is enabled \\
\> \> \> Learning update: Using $v^j_0 = W^j v^j$, perform a left NMF update on $W^j$, using, e.g. (\ref{eqn:left_nmf_update}) \\
\> \> {\bf end} \\
\> \> Inference update: Using $v^j_0 = W^j v^j$, perform a right NMF update on $v^j$, using, e.g. (\ref{eqn:right_nmf_update}) \\
\> {\bf end} \\
{\bf end} \\
\\
// Using the values of the parent variables, update the values of the level $l$ child variables by performing \\
// value propagation. \\
{\bf down}\={\bf step}($l$) \\
\> {\bf for} \={\bf each} $j \in FactorSystem_l$ \\
\> \> // Perform value propagation \\
\> \> $v^j_0 \gets W^j v^j$ \\
\> {\bf end} \\
{\bf end} 
\end{tabbing}
\end{algorithm}

\begin{algorithm}
\caption{averageChildren() and averageParents() procedures}
\label{alg:averagingProcedures}
\begin{tabbing}
// Let $X_l$ denote the set of model variables \{$x^l_1, x^l_2, \dots, x^l_{LevelCount_l}$\} corresponding to the level $l$ nodes.\\
//  Let $duplicationCountChild(i,l)$ = $|duplicationSetChild(i,l)|$.\\
// Let $duplicationSet(i,l)$ = \{$(j,k) : eq_j \in FactorSystem$ and $v^j_k$ corresponds to $x^l_i$\}. \\
// Update the value of each $x^l_i \in X_l$ as the mean value of the corresponding $v^j_0$ \\
// that appear in $FactorSystem_l$. Then set all $v^j_k$ corresponding to $x^l_i$ to this value as well.\\
{\bf aver}\={\bf ageChildren}($l$) \\
\> {\bf for} \= $i$ = 1 to $LevelCount_l$ \\
\> \> // For each (child) variable $x^l_i \in X_l$ \\
\> \> {\bf if} \=$x^l_i$ is hidden \\
\> \> \> meanValue = $\frac{1}{duplicationCountChild(i,l)} \sum_{j \in duplicationSetChild(i,l)} v^j_0$ \\
\> \> \> {\bf for} \= {\bf each} $(j,k) \in duplicationSet(i,l)$ \\
\> \> \> \> $v^j_k \gets meanValue $ \\
\> \> \> {\bf end} \\
\> \> \> $x^l_i \gets meanValue $ \\
\> \> {\bf else if} $x^l_i$ is observed \\
\> \> \> {\bf for each} $(j,k) \in duplicationSet(i,l)$ \\
\> \> \> \> $v^j_k \gets x^l_i $ \\
\> \> \> {\bf end} \\
\> \> {\bf end} \\
\> {\bf end} \\
{\bf end} \\
\\
// Let $X_{l+1}$ denote the set of model variables \{$x^{l+1}_1, x^{l+1}_2, \dots, x^{l+1}_{LevelCount_{l+1}}$\} corresponding to the level $l+1$ nodes.\\
// Note that it is possible for a level $l$ variable to have some or all of its parents in a level higher than $l+1$, in\\
// which case the set $X_{l+1}$ will only represent a proper subset of the parents of level $l$. \\
// Let $duplicationSetParent(i,l+1)$ = \{$(j,k) :  eq_j \in FactorSystem_l$ and $v^j_k, k \geq 1$ corresponds to $x^{l+1}_i$\}. \\
// Let $duplicationCountParent(i,l+1) = |duplicationSetParent(i,l+1)|$. \\
// For each $x^{l+1}_i \in X_{l+1}$, update the corresponding $v^j_k, k \geq 1$ to their mean value.\\
{\bf aver}\={\bf ageParents}($l$) \\
\> {\bf for} \= $i$ = 1 to $LevelCount_{l+1}$ \\
\> \> // For each (parent) variable $x^{l+1}_i \in X_{l+1}$ \\
\> \> {\bf if} \=$x^{l+1}_i$ is hidden \\
\> \> \> meanValue = $\frac{1}{duplicationCountParent(i,l+1)} \sum_{(j,k) \in duplicationSetParent(i,l+1)} v^j_k$ \\
\> \> \> {\bf for} \= {\bf each} $(j,k) \in duplicationSet(i,l+1)$ \\
\> \> \> \> $v^j_k \gets meanValue $ \\
\> \> \> {\bf end} \\
\> \> \> $x^{l+1}_i \gets meanValue$ \\
\> \> {\bf else if} $x^{l+1}_i$ is observed \\
\> \> \> {\bf for each} $(j,k) \in duplicationSet(i,l+1)$ \\
\> \> \> \> $v^j_k \gets x^{l+1}_i $ \\
\> \> \> {\bf end} \\
\> \> {\bf end} \\
\> {\bf end} \\
{\bf end} 
\end{tabbing}
\end{algorithm}

Note that the only distinction the algorithm makes between $X_E$ and $X_H$ is that the values of the $X_E$ variables are not updated during inference. If we wish to make some previously hidden variables observed or vice versa, is is then trivial to modify the algorithm to handle this by simply enabling or disabling updates on those variables. 

We also note that typically, some kind of normalization of the parameter matrices $W$ will be performed during the left NMF update steps. For example, one could normalize the column of $W$ to have unit sum. 

Note also that inference and learning are performed jointly. We can perform inference only by simply disabling the learning updates. If a subset of the parameters $\theta$ are known, then the learning updates can be disabled for those parameters.

The computational cost of performing a single iteration of the inference and learning algorithm is given by the sum of the costs of performing the learning and inference NMF updates and the value propagation multiplication on each factorization equation. The number of iterations to convergence can depend on several factors, such as the longest path in the graph, NMF algorithm employed, etc.

\section{Factored sequential data models}
\label{sec:main_factored_model}

We now consider PFNs for modeling sequential data, which we will refer to as dynamic positive factor networks (DPFNs). A DPFN provides for the modeling of a data sequence as an additive combination of realizations of an underlying process model. Analogously to a dynamic Bayesian network (DBN) \cite{Friedman1999}, a DPFN is created by horizontally replicating a PFN subgraph. Thus, a DPFN is simply a PFN that has a regular repeating graphical structure. The corresponding parameters are also typically replicated. We only require that the particular subgraph that is replicated correspond to a valid PFN. We will refer to the subgraph that is replicated as corresponding to a time slice. Although we use terminology that implies a time series, such an interpretation is not required in general. For example, the data could correspond to a biological sequence, or the characters or words in a text document, for example.

Our models will make use of an non-negative linear state representation. Intuitively, one can think of the model as supporting the simultaneous representation of any additive combination of allowable realizations of an underlying state transition model, or more generally, an underlying dynamic process model. A state variable in the model is defined such that the dimensionality of the variable corresponds to the number of possible states. We allow these variables to be general non-negative vectors, so that the non-zero-valued components of the variable represent a (positive) weight or strength of the corresponding state. Any given non-zero valued state variable then corresponds to some superposition of states. A given pair of state variables in adjacent time slices is then factored in terms of a set of basis vectors that specify the allowable transitions, and a corresponding encoding vector. The factored state representation seems to allow significant representational power, while making use of a compact parameter set.

In this section we present a factored state transition model and present empirical results to illustrate the performance of the inference and learning algorithms.

\subsection{Model}
\label{sec:fact_state_tran_model}

\begin{figure}
\centering
\includegraphics[width=60ex]{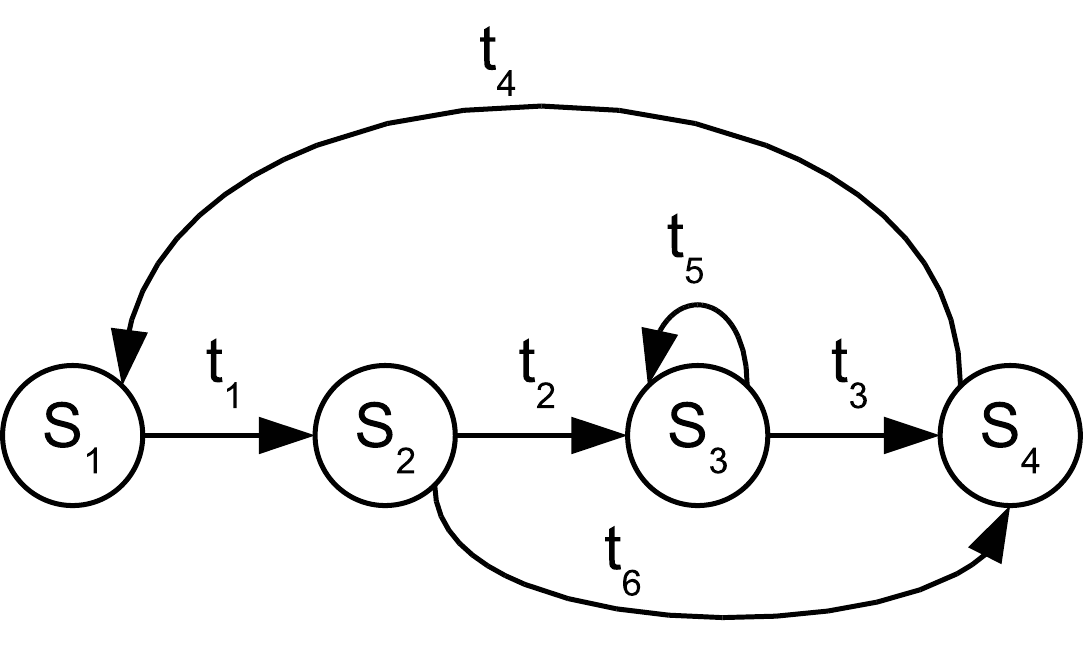}
\caption{A state transition diagram for a 4-state automaton.}
\label{fig:fsm1}
\end{figure}

Consider a transition model with $M$ states and $R$ possible state transitions. Figure~\ref{fig:fsm1} shows an example state transition diagram for a 4-state automaton. This transition model contains $M = 4$ states, labeled $S_1, \dots, S_4$ and $R = 6$ state transitions, labeled $t_1, \dots, t_6$. 

\begin{figure}
\centering
\includegraphics[width=60ex]{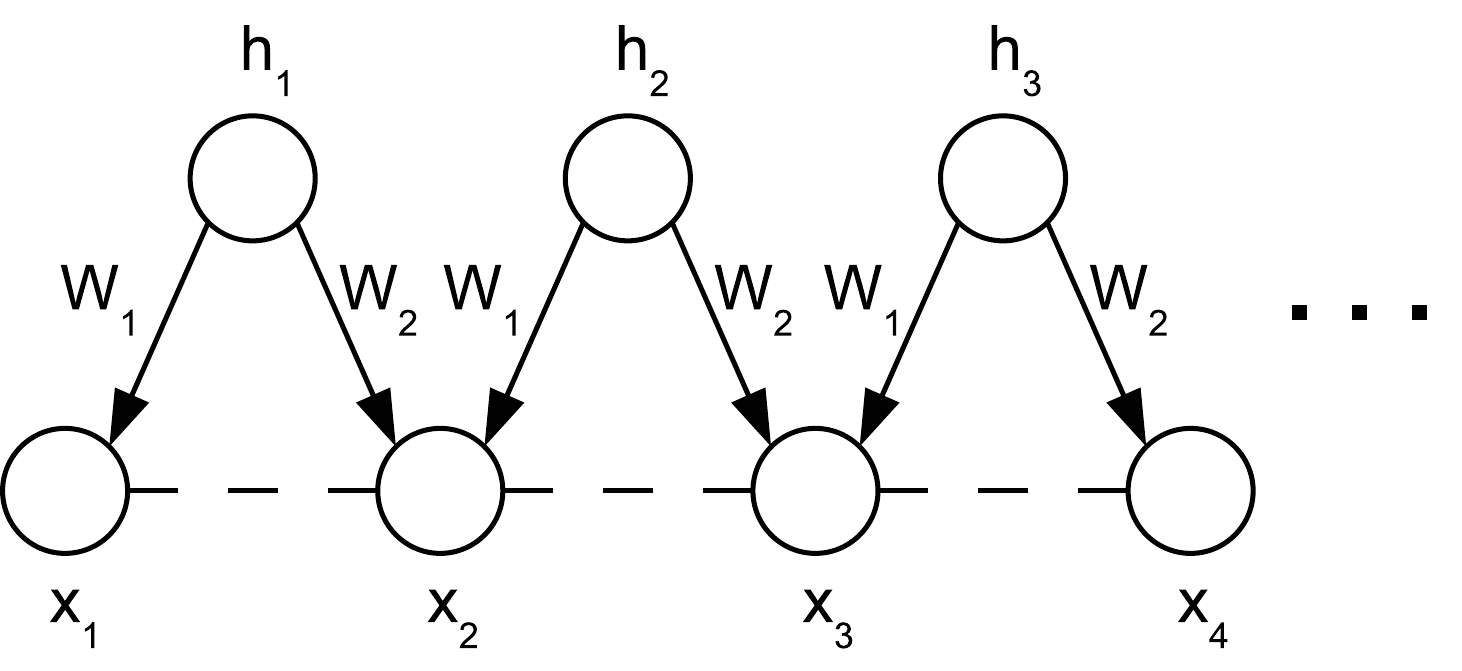}
\caption{The DPFN corresponding to the factorized state model in Equation (\ref{eqn:key_factorization}). The first four time slices are shown. The state at time slice $t$ is represented by $x_t$.  We place constrains on states at adjacent time slices such that the states $x_t$, $x_{t+1}$ are represented as an additive combination of the allowable state transition vectors. $h_t$ represents the encoded transitions for $x_t, x_{t+1}$.}
\label{fig:1layerDynamic}
\end{figure}

Figure~\ref{fig:1layerDynamic} shows the DPFN that we will use to model a process that evolves according to the transition model given in Figure~\ref{fig:fsm1}. This DPFN corresponds to a system of $T-1$ factorization equations where for each $t \in 1, \dots, T-1$ we have:

\begin{align}
\left[ \begin{array}{c} 
x_t\\
x_{t+1}  \end{array} \right] =&\ W h_t \notag \\
 =&\ \left[ \begin{array}{c}
W_1 \\
W_2 \end{array} \right] h_t 
\label{eqn:single_timeslice_factorization}
\end{align}

Since the same parameter matrix $W$ appears in each equation, the above system of $T-1$ vector factorization equations can be expressed as a single matrix factorization equation:

\begin{align}
\label{eqn:expanded_key_factorization}
\left[ \begin{array}{ccccc}
x_1 & x_2 & x_3 & \dots & x_{T-1} \\
x_{2} & x_{3} & x_{4} & \dots & x_{T} \end{array} \right] =&\ W \left[ \begin{array}{ccccc} h_1 & h_2 & h_3 & \dots & h_{T-1} \end{array} \right] \notag \\
=&\ \left[ \begin{array}{c}
W_1 \\
W_2 \end{array} \right] \left[ \begin{array}{ccccc} h_1 & h_2 & h_3 & \dots & h_{T-1} \end{array} \right]
\end{align}

Let us define $x_{c_t}$ as the vertical concatenation of any two adjacent state variables $x_t$, $x_{t+1}$ so that:

\begin{align}
x_{c_t} = \left[ \begin{array}{c} 
x_t\\
x_{t+1}  \end{array} \right]
\end{align}

We define the $2 M$ x T-1 matrix $X_c$ as the following horizontal concatenation of the time-ordered $x_{c_t}$ vectors:

\begin{align}
X_c =& \left[ \begin{array}{cccccc} x_{c_1} & x_{c_2} & x_{c_3} & \dots & x_{c_{T-1}} \end{array} \right] \notag \\
=& \left[ \begin{array}{ccccc}
x_1 & x_2 & x_3 & \dots & x_{T-1} \\
x_{2} & x_{3} & x_{4} & \dots & x_{T} \end{array} \right]
\end{align}

We define the $R$ x T-1 matrix $H$ as the following horizontal concatenation of the time-ordered \{$h_t$\} vectors:

\begin{align}
H = \left[ \begin{array}{cccccc} h_1 & h_2 & h_3 & \dots & h_{T-1} \end{array} \right]
\end{align}

Using the above matrix definitions, we can then write the matrix factorization equation (\ref{eqn:expanded_key_factorization}) more compactly as:

\begin{align}
X_c = W H
\label{eqn:key_factorization}
\end{align}

We now illustrate how this DPFN can be used to represent the transition model in Figure~\ref{fig:fsm1}. We let vector $x_t \in  \mathbb{R}^M$ represent the model state at time $t$. The sequence, $\{x_t : t = 1, \dots, T\}$ is then modeled as a realization of the transition model specified in the state transition diagram. The parameter matrix $W$ specifies the transition model, and the vector $h_t$ represents an encoding of $(x_t, x_{t+1})$ in terms of the basis columns of $W$.

The parameter matrix $W$ is constructed directly from the state transition diagram as follows. For each state $S_i$, we define a corresponding \emph{state basis vector} $s_i \in  \mathbb{R}^M$ such that the $i$'th component is 1 and all other components are zero. States $S_1, \dots, S_4$ then correspond to the following state basis vectors, respectivley:

\begin{align}
s_1=& \left[ \begin{array}{c} 
1\\
0\\
0\\
0  \end{array} \right], 
s_2 = \left[ \begin{array}{c} 
0\\
1\\
0\\
0  \end{array} \right], 
s_3 = \left[ \begin{array}{c} 
0\\
0\\
1\\
0  \end{array} \right], 
s_4 = \left[ \begin{array}{c} 
0\\
0\\
0\\
1  \end{array} \right]
\end{align}

For each transition $t_k$ in the state transition diagram that represents a transition from state $S_i$ to state $S_j$, we define the corresponding \emph{transition basis vector} $w_k$ as the vertical concatenation of the corresponding state basis vectors $s_i$ on top of $s_j$. Each of the $R$ transitions $t_k$ will then have a corresponding $2 M$ x 1 transition basis vector $w_k$ given by:

\begin{align}
w_k =  \left[ \begin{array}{c} 
s_i\\
s_j  \end{array} \right]
\end{align}

For example, transition $t_2$ in the diagram, which represents a transition from state $S_2$ to state $S_3$ has a corresponding transition basis vector $w_2$ given by:

\begin{align}
w_2 =  \left[ \begin{array}{c} 
s_2\\
s_3  \end{array} \right] = \left[ \begin{array}{c} 
0\\
1\\
0\\
0\\ \hline
0\\
0\\
1\\
0  \end{array} \right]
\end{align}

Let the $2 M$ x $R$ \emph{transition basis matrix} $W$ be defined as the horizontal concatenation of the transition basis vectors \{$w_i$\}. Any ordering of the columns is possible. The columns of $W$ then specify the allowable transitions in our model:

\begin{align}
W =& \left[ \begin{array}{ccccc} w_1 & w_2 & w_3 & \dots & w_R \end{array} \right]
\end{align}

We will find it useful to partition $W$ into upper and lower sub-matrices $W_1$ and $W_2$ such that the $M$ x $R$ upper sub-matrix $W_1$ represents the time slice $t-1$ state basis vectors and the $M$ x $R$ lower sub-matrix $W_2$ represents the corresponding time slice $t$ state basis vectors. One possible $W$ corresponding to the transition diagram in Figure~\ref{fig:fsm1} is given by:

\begin{align}
W =& \left[ \begin{array}{cccccc} w_1 & w_2 & w_3 & w_4 & w_5 & w_6 \end{array} \right] \notag \\
=&\ \left[ \begin{array}{c}
W_1 \\
W_2 \end{array} \right] \notag\\
=&\ \left[ \begin{array}{cccccc}
s_1 & s_2 & s_3 & s_4 & s_3 & s_2 \\
s_2 & s_3 & s_4 & s_1 & s_3 & s_4 \end{array} \right] \notag \\
=& \left[ \begin{array}{cccccc}
1 & 0 & 0 & 0 & 0 & 0 \\
0 & 1 & 0 & 0 & 0 & 1 \\
0 & 0 & 1 & 0 & 1 & 0 \\
0 & 0 & 0 & 1 & 0 & 0 \\ \hline
0 & 0 & 0 & 1 & 0 & 0 \\
1 & 0 & 0 & 0 & 0 & 0 \\
0 & 1 & 0 & 0 & 1 & 0 \\
0 & 0 & 1 & 0 & 0 & 1 \end{array} \right] 
\label{eqn:W_non_deterministic}
\end{align}

Consider the following state sequence of length $T=10$, which is a valid sequence under the example transition diagram:

\begin{align}
\label{eq:example_state_seq}
seq_1 = (S_1, S_2, S_3, S_4, S_1, S_2, S_3, S_4, S_1, S_2)
\end{align}

Now suppose we construct a sequence of state variables $\{x_t: t=1,\dots,T\}$ corresponding to this state sequence by setting $x_t$ equal to the corresponding state basis vector $s_i$ for state $S_i$ at time $t$:

\begin{align}
x_1 = s_1 = \left[ \begin{array}{c} 
1\\
0\\
0\\
0  \end{array} \right], 
x_2 = s_2 = \left[ \begin{array}{c} 
0\\
1\\
0\\
0  \end{array} \right],
x_3 = s_3 = \left[ \begin{array}{c} 
0\\
0\\
1\\
0  \end{array} \right],
\dots,
x_{10} = s_2 = \left[ \begin{array}{c} 
0\\
1\\
0\\
0  \end{array} \right]
\label{eqn:simple_state_seq}
\end{align}

Let us define the $M$ x $T$ matrix $X$ as the horizontal concatenation of the state variables $x_t$ as:

\begin{align}
X =& \left[ \begin{array}{ccccc} x_1 & x_2 & x_3 & \dots & x_T \end{array} \right]
\end{align}

We then have the following sequence $X$ corresponding to the state sequence (\ref{eq:example_state_seq}):

\begin{align}
X =& \left[ \begin{array}{cccccccccc} x_1 & x_2 & x_3 & x_4 & x_5 & x_6 & x_7 & x_8 & x_9 & x_{10} \end{array} \right] \notag \\
=& \left[ \begin{array}{cccccccccc} 
1 & 0 & 0 & 0 & 1 & 0 & 0 & 0 & 1 & 0\\
0 & 1 & 0 & 0 & 0 & 1 & 0 & 0 & 0 & 1\\
0 & 0 & 1 & 0 & 0 & 0 & 1 & 0 & 0 & 0\\
0 & 0 & 0 & 1 & 0 & 0 & 0 & 1 & 0 & 0 \end{array} \right]
\label{eqn:fsm1_determ_X1}
\end{align}

Using the above $X$ and (\ref{eqn:key_factorization}), we then have the following model factorization:

\begin{align}
X_c =& W H \notag \\
\left[ \begin{array}{ccccccccc} 
1 & 0 & 0 & 0 & 1 & 0 & 0 & 0 & 1\\
0 & 1 & 0 & 0 & 0 & 1 & 0 & 0 & 0\\
0 & 0 & 1 & 0 & 0 & 0 & 1 & 0 & 0\\
0 & 0 & 0 & 1 & 0 & 0 & 0 & 1 & 0\\ \hline
0 & 0 & 0 & 1 & 0 & 0 & 0 & 1 & 0\\
1 & 0 & 0 & 0 & 1 & 0 & 0 & 0 & 1\\
0 & 1 & 0 & 0 & 0 & 1 & 0 & 0 & 0\\
0 & 0 & 1 & 0 & 0 & 0 & 1 & 0 & 0 \end{array} \right] =&
\left[ \begin{array}{cccccc}
1 & 0 & 0 & 0 & 0 & 0 \\
0 & 1 & 0 & 0 & 0 & 1 \\
0 & 0 & 1 & 0 & 1 & 0 \\
0 & 0 & 0 & 1 & 0 & 0 \\ \hline
0 & 0 & 0 & 1 & 0 & 0 \\
1 & 0 & 0 & 0 & 0 & 0 \\
0 & 1 & 0 & 0 & 1 & 0 \\
0 & 0 & 1 & 0 & 0 & 1 \end{array} \right] 
\left[ \begin{array}{ccccccccc} 
1 & 0 & 0 & 0 & 1 & 0 & 0 & 0 & 1\\
0 & 1 & 0 & 0 & 0 & 1 & 0 & 0 & 0\\
0 & 0 & 1 & 0 & 0 & 0 & 1 & 0 & 0\\
0 & 0 & 0 & 1 & 0 & 0 & 0 & 1 & 0\\
0 & 0 & 0 & 0 & 0 & 0 & 0 & 0 & 0\\
0 & 0 & 0 & 0 & 0 & 0 & 0 & 0 & 0 \end{array} \right]
\end{align}

 Note that given $X_C$ and $W$, only one solution for $H$ is possible. We see that each pair of state vectors $x_{c_t}$ in $X_c$ corresponds to exactly one transition basis vector in the resulting factorization, since the value of each $x_t$ was chosen equal to one of the $s_i$ and $X$ corresponds to a valid state sequence. The columns of $H$ then given an encoding of $X$ in terms of the basis vectors of $W$ so that exactly one component of each column $h_t$ has value 1, corresponding to the transition that explains $x_{c_t}$. We say that a sequence $X$ of state vector observations form an \emph{elementary state sequence} if there exists a factorization $X_C = W H$ such that each column of $H$ contains at most one nonzero component.

Now let us return to the case where the $x_t$ is no longer constrained to be equal to one the state basis vectors. We only require $x_t$ to be a non-negative vector in $\mathbb{R}^M$. Although $W$ is specified in the example above as containing only 0 and 1 valued components, in general we only place the constraint that the matrices of the factorization equation be non-negative. Note that since the model is a PFN, the linearity property from Section~\ref{sec:model_specification} holds, so that any additive combination of solutions is also a solution. That is, any additive mixture of state sequences that are valid under the transition model of $W$ are representable.  We can also see that this property follows directly from (\ref{eqn:key_factorization}). For non-negative scalars $\alpha$, $\beta$, we have:

\begin{align}
X_{c_a} =& W H_a \notag \\
X_{c_b} =& W H_b \notag \\
\Leftrightarrow \alpha X_{c_a} + \beta X_{b_c} =& W (\alpha H_a + \beta H_b)
\end{align}

For the case where multiple components of $x_t$ are positive, the interpretation is that multiple states are simultaneously active at time $t$. For example, consider an example where $x_t = (\alpha, 0, 0, \beta)^T$. That is, at time $t$, the model is in state $S_1$ with magnitude $\alpha$ and is simultaneously in state $S_4$ with magnitude $\beta$. Suppose that $x_{t+1}$ is hidden. We can see from the factorization equation above that the solution is $x_{t+1} = (\beta, \alpha, 0, 0)^T$. That is, our factored transition model specifies that if the model is in state $S_1$ with magnitude $\alpha$ at time $t$ then the model must be in state $S_2$ with magnitude $\alpha$ at time $t+1$. Likewise, if the model is in state $S_4$ with magnitude $\beta$ at time $t$ then the model must be in state $S_1$ with magnitude $\beta$ at time $t+1$. Due to the factored state representation, the model can represent both realizations simultaneously. A sequence of state vectors is modeled as an non-negative linear combination of the allowable transition basis vectors.

Note also that since we do not impart a probabilistic interpretation, all allowable outgoing transitions from a given state can be considered equally likely. For example, suppose $x_{t+1}$ is hidden and that $x_t = (0, \alpha, 0, 0)$, corresponding to the model being in state $S_2$ with magnitude $\alpha$ at time $t$. Since there are multiple outgoing transitions from $S_2$ ($t_2$ and $t_6$), multiple solutions for $x_{t+1}$ are possible. We will explore the issue of performing inference when multiple solutions are possible in the following results sections.

A learning and inference algorithm for this network can be obtained by applying Algorithm~\ref{alg:inference_learning1} to the system of factorizations (\ref{eqn:key_factorization}). Example pseudocode is presented in Appendix~\ref{sec:simpleDyn1_inf_learn}.

\subsection{Empirical results}
\label{sec:dyn_model}

In this section, we perform inference and learning on the dynamic network in Figure~\ref{fig:1layerDynamic}, for two distinct transition models, using synthetic data sets. We present results for the cases of fully and partially observed input sequences. We also present results for sequences that have been corrupted by noise. We first consider the case where the underlying transition model is known. We then present results for the case where the transition model is learned from training data.

\begin{figure}
\centering
\includegraphics[width=60ex]{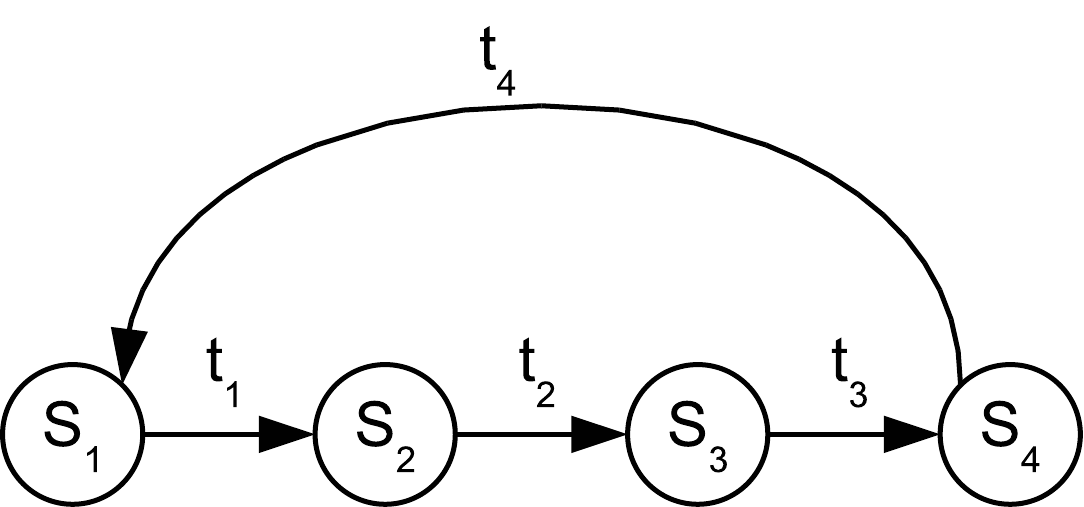}
\caption{A deterministic state transition diagram for a 4-state automaton.}
\label{fig:fsm1_deterministic}
\end{figure}

\subsubsection{Fully observed input sequences under a known transition model}

We start with a simple deterministic transition model, and will later revisit the earlier nondeterministic model from  Figure~\ref{fig:fsm1}.  Figure~\ref{fig:fsm1_deterministic} shows the state transition diagram for a deterministic model that is obtained by removing transitions $t_5$ and $t_6$ from the earlier nondeterministic model. Using the previously outlined procedure, we obtain a parameter matrix $W$ corresponding to this transition diagram as follows. First we construct a transition basis vector $w_k$ corresponding to each transition $t_k$:

\begin{align}
w_1=& \left( \begin{array}{c} s_1\\
s_2 \end{array} \right) ,
w_2= \left( \begin{array}{c} s_2\\
s_3 \end{array} \right) ,
w_3= \left( \begin{array}{c} s_3\\
s_4 \end{array} \right) ,
w_4= \left( \begin{array}{c} s_4\\
s_1 \end{array} \right) ,
\end{align}

The transition basis matrix $W$ is then constructed as the horizontal concatenation of the transition basis vectors (again, the column ordering is arbitrary):

\begin{align}
W = \left[ \begin{array}{cccc} w_1 & w_2 & w_3 & w_4 \end{array} \right]
\end{align}

We start by considering the case where the state sequence $X$ is fully observed, and the transition sequence $H$ is hidden. We then wish to infer the values of $H$ given the observed $X$.  Suppose we wish to specify an input sequence $X$ that corresponds to the following sequence of states: $(S_1, S_2, S_3, S_4, S_1, S_2, S_3, S_4, S_1, S_2)$, which is a valid state sequence under the transition diagram in Figure~\ref{fig:fsm1_deterministic}. This sequence can be represented by setting $X =\alpha \left[ \begin{array}{cccccccccc} s_1 & s_2 & s_3 & s_4 & s_1 & s_2 & s_3 & s_4 & s_1 & s_2 \end{array} \right]$, where $\alpha$ is any positive scalar. For example, the choice $\alpha = 1$ results in:

\begin{align}
X = \left[ \begin{array}{cccccccccc} 
1 & 0 & 0 & 0 & 1 & 0 & 0 & 0 & 1 & 0\\
0 & 1 & 0 & 0 & 0 & 1 & 0 & 0 & 0 & 1\\
0 & 0 & 1 & 0 & 0 & 0 & 1 & 0 & 0 & 0\\
0 & 0 & 0 & 1 & 0 & 0 & 0 & 1 & 0 & 0 \end{array} \right]
\label{eqn:not_used}
\end{align}

\begin{figure}
\centering
\includegraphics[width=60ex]{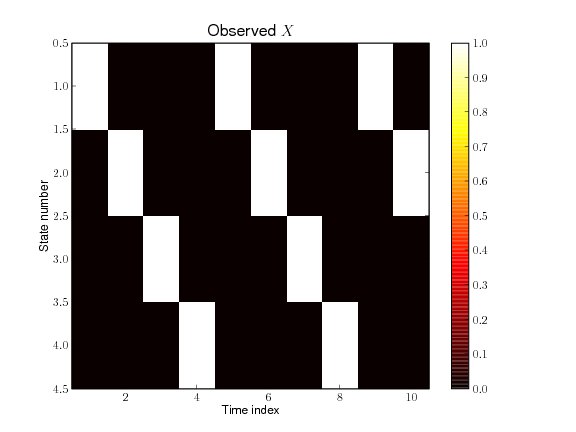}
\caption{An image plot of the observed $X$ from Equation~\ref{eqn:fsm1_determ_X1}, corresponding to the state sequence: $(S_1, S_2, S_3, S_4, S_1, S_2, S_3, S_4, S_1, S_2)$. Row $i$ of $X$ corresponds to state $S_i$. Column $i$ corresponds to the $i$'th time slice. Here and throughout this paper we use the "hot" color map shown on the right, so that components with a minimum component value are displayed in black, components with the maximum component value are displayed in white, and values in between are displayed in shades of red, orange, and yellow.}
\label{fig:fsm1_determ_X1}
\end{figure}

We will be dealing with matrices that are non-negative, typically sparse, and such that many or all of the non-zero components will take on the same value. For visualization purposes, the particular value will typically not be relevant. For these reasons we find that, rather than simply printing out the numerical values of a matrix as above, an image plot can be a more visually appealing way to take in the relevant information. In the remainder of this paper, we will therefore make extensive use of image plots of the matrices that appear in the various PFN factorization equations. In displaying the image plots, we make use of the ``hot'' colormap, so that zero-valued components appear black, small values appear red, larger values yellow, and the maximum valued component(s) in a matrix appear white. Figure~\ref{fig:fsm1_determ_X1} shows an image plot of the observed $X$ from above.

Recall that the network in Figure~\ref{fig:1layerDynamic} corresponds to the matrix factorization Equation (\ref{eqn:key_factorization}), which we reproduce here:

\begin{align}
X_C = W H
\label{eqn:repeatedDPN1}
\end{align}

Recall also that $X_C$ is defined as the vertical stacking of adjacent columns of $X$ and so only $H$ is unknown (hidden). We solve for the values of $H$ by applying the inference algorithm from Appendix~\ref{sec:simpleDyn1_inf_learn}, which is a special case of the general learning and inference algorithm specified in Algorithm~\ref{alg:inference_learning1}. In solving for $H$, this corresponds to first initializing $H$ to random positive values and then iterating the inference algorithm (learning is disabled since $W$ is known) until convergence. For all empirical results in this paper, hidden variables are initialized to random positive values uniformly distributed between 0 and $10^{-6}$. Each iteration of the inference algorithm in this case corresponds to performing an NMF right update step, e.g., using (\ref{eqn:right_nmf_update}).

\begin{figure}
\centering
\includegraphics[width=80ex]{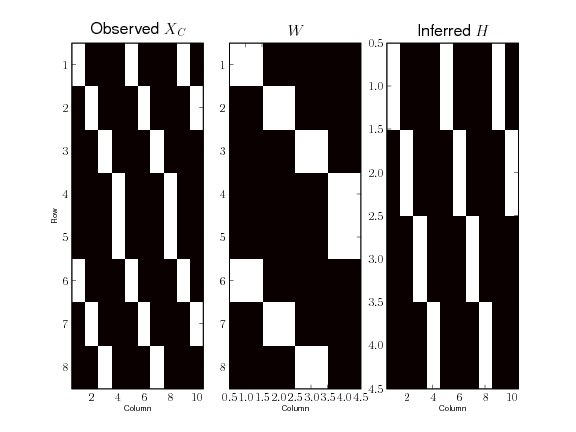}
\caption{A plot of the matrices of the factorization $X_C = W H$, after solving for $H$ Note that the upper half submatrix of $X_c$ is equal to $X$, which corresponds to the state sequence: $(S_1, S_2, S_3, S_4, S_1, S_2, S_3, S_4, S_1, S_2$). Here, $W$ corresponds to the deterministic state transition diagram in Figure~\ref{fig:fsm1_deterministic}}
\label{fig:fsm1_determ_factor}
\end{figure}

Figure~\ref{fig:fsm1_determ_factor} shows a plot of the matrices in the factorization, where $H$ is shown after convergence of the inference algorithm. For this input sequence, a few hundred iterations was sufficient for the RMSE (root mean squared error between $X$ and the reconstruction $W H$) to drop below $10^-4$. The inference computation, implemented in Java and Python, took approximately 1 second to run on a desktop PC with a 3 GHz Core 2 Duo processor, which is representative of the time required to run each of the examples in this section.

\begin{figure}
\centering
\includegraphics[width=80ex]{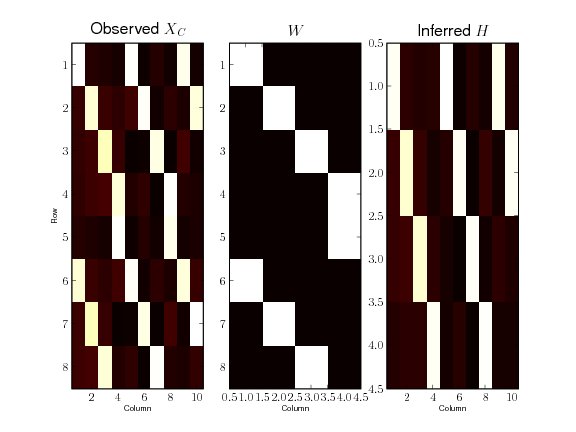}
\caption{A plot of the matrices of the factorization $X_C = W H$, where the observations $X$ are corrupted by additive noise. The inferred hidden transition sequence $H$ is shown after convergence of the inference algorithm. Here, $W$ corresponds to the deterministic state transition diagram in Figure~\ref{fig:fsm1_deterministic}}
\label{fig:fsm1_determ_factor_noise}
\end{figure}

We now add some noise consisting of uniformly distributed values in $[0, 0.1]$ to $X$. Figure~\ref{fig:fsm1_determ_factor_noise} shows the resulting estimate for $H$. The noisy observations can no longer be represented exactly as a linear combination of the transition basis vectors, yielding an RMSE of 0.034. We observe empirically that the approximate factorization found by the NMF updates appears to be relatively insensitive to small amounts of additive noise, and produces an inference result that appears visually similar to applying additive noise to the previous noiseless solution for $H$. 

We now consider an example where the observation sequence corresponds to a mixture of realizations of the underlying transition model. Recall that the if observation sequence $X_a$ and the hidden sequence $H_a$ are one solution to the model, and observation sequence $X_b$ and the hidden sequence $H_b$ are another solution to the model, then the observation sequence $X = X_a + X_b$ and the hidden sequence $H = H_a + H_b$ are also a solution of the model. As an example, consider the observation sequence $X = X_a + X_b$, where $X_a$ and $X_b$ are given by:

\begin{align}
 X_a =& 0.5 \left[ \begin{array}{cccccccccc} s_1 & s_2 & s_3 & s_4 & s_1 & s_2 & s_3 & s_4 & s_1 & s_2 \end{array} \right] \notag \\
 X_b =& 1.0 \left[ \begin{array}{cccccccccc} s_3 & s_4 & s_1 & s_2 & s_3 & s_4 & s_1 & s_2 & s_3 & s_4 \end{array} \right]
 \label{eqn: superposition1}
 \end{align}

\begin{figure}
\centering
\includegraphics[width=60ex]{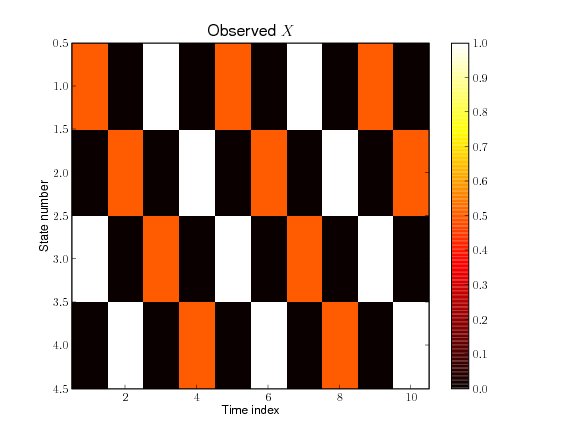}
\caption{An image plot of an observation sequence  $X = X_a + X_b$ consisting of an additive combination of two sequences $X_a$ and $X_b$, which are each a realization of the underlying network transition model. Here, $X_1$ begins in state $s_1$ in the first time slice, and corresponds to the orange components since it has half the magnitude of sequence $X_b$. Sequence $X_b$ begins in state $s_3$ and corresponds to the white components.}
\label{fig:fsm1_determ_X1_superpos}
\end{figure}

\begin{figure}
\centering
\includegraphics[width=80ex]{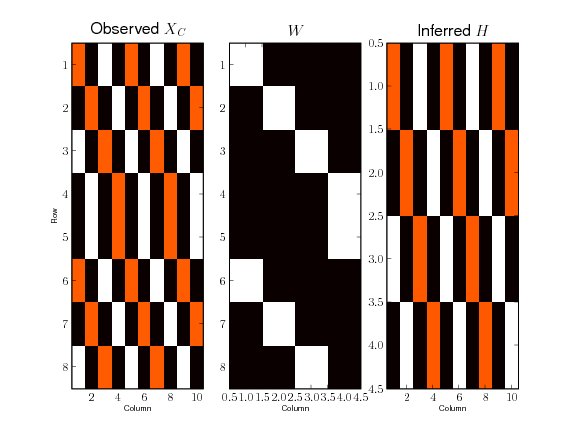}
\caption{An image plot of the matrices of the factorization $X_C = W H$, where the observations are an additive combination of two sequences: $X = X_a + X_b$. $H$ is shown after convergence of the inference algorithm. Here, $W$ corresponds to the deterministic state transition diagram in Figure~\ref{fig:fsm1_deterministic}}
\label{fig:fsm1_determ_factor_superpos}
\end{figure}

Figure~\ref{fig:fsm1_determ_X1_superpos} shows an image plot of the observed sequence $X = X_a + X_b$. Note that both sequences $X_a$ and $X_b$ correspond to a valid realization of the underlying transition model of the network. From the superposition property, we know that the sum sequence $X$ has a corresponding hidden sequence $H$ which satisfies the factorization Equation~\ref{eqn:repeatedDPN1} with equality. The expressiveness of this network is such that any observations and hidden variables corresponding to any non-negative superposition of valid realizations of the underlying transition model specified by $W$ are representable. This does not necessarily mean that our particular choice of inference algorithm will always be able find the corresponding solution, however. We do observe, though, that for this particular network choice, repeated runs of our inference algorithm always converged to an exact solution (specifically, inference was stopped when the RMSE dropped below a certain threshold: $10^-4$). Figure~\ref{fig:fsm1_determ_factor_superpos} shows an image plot of the inferred $H$ along with the other matrices in Equation~\ref{eqn:repeatedDPN1}.

\subsubsection{Partially observed input sequences under a known deterministic transition model}
\label{sec:example_deterministic_fsm}


We now consider the case where the sequence $X$ is partially observed so that we are given the values of $x_t$ for some subset of time slices, and the values of $x_t$ for the remaining time slices are hidden. We would then like to infer the values of all hidden variables in the model, which will consist of $H$ along with the hidden subset of $\{x_t\}$. 

Suppose that $X$ represents a sequence of length 10 where the first time slice $x_1$ is observed and corresponds to state $S_1$. That is, $x_1 = \alpha s_1$. We arbitrarily choose $\alpha = 1$. We then wish to infer the values of the future time slices of $X$ as well as the values of $H$. That is, we wish to perform prediction. We again use the same inference algorithm and initialize all hidden variables to small positive random values.

\begin{figure}
\centering
\includegraphics[width=80ex]{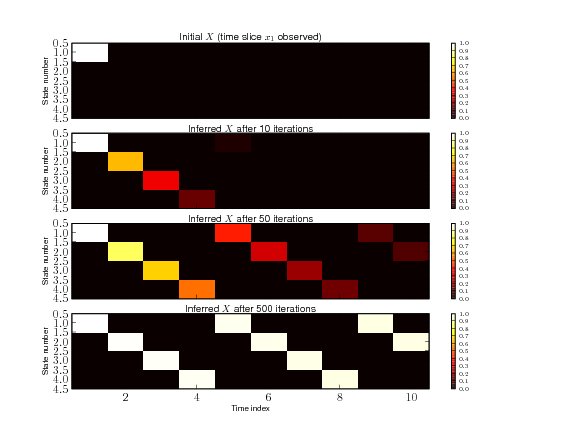}
\caption{A plot of the estimate for $X$ after various iteration counts of the inference algorithm. The top image shows the initial $X$ which only has the first time slice (time index 1 in the figure) set to state $S_1$ (i.e., $x_1 = s_1$). The future time slices $(x_2 \dots x_{10})$ are initialized to small random values, which appear black in the figure since their maximum value of $10^-6$ is small compared to the largest value of 1 in the figure. The next lower figures show the $X$ estimates after 10, 50, and 500 iterations, respectively. We see that 500 iterations is sufficient to effectively reach convergence.}
\label{fig:fsm1_determ_factor_X1_after_various_iterations}
\end{figure}

\begin{figure}
\centering
\includegraphics[width=80ex]{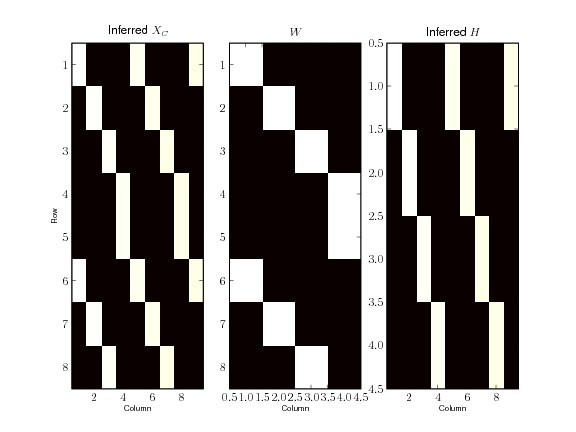}
\caption{An image plot of the matrices of the factorization $X_C = W H$, after solving for $H$ and the hidden time slices $(x_2 \dots x_{10})$ of $X$. Here, $W$ corresponds to the deterministic state transition diagram in Figure~\ref{fig:fsm1_deterministic}}
\label{fig:fsm1_determ_factor_partial}
\end{figure}

Figure~\ref{fig:fsm1_determ_factor_X1_after_various_iterations} shows the initial $X$ and the estimates after 10, 50, and 500 iterations of the inference algorithm. We see that convergence is effectively reached by 500 iterations. Figure~\ref{fig:fsm1_determ_factor_partial} shows the corresponding plots of the matrices of the network factorization equation after 500 iterations. We observe that the inference estimates for $X$ appear to converge outward in time from the observed time slice $x_1$. The local variable averaging step of the inference algorithm causes the transition model to propagate a positive value one time slice forward and backward for each iteration. Note also that given the observed time slice, there is only one possible solution for the hidden variables (and that our inference algorithm successfully finds it) since the transition model is deterministic.

\begin{figure}
\centering
\includegraphics[width=80ex]{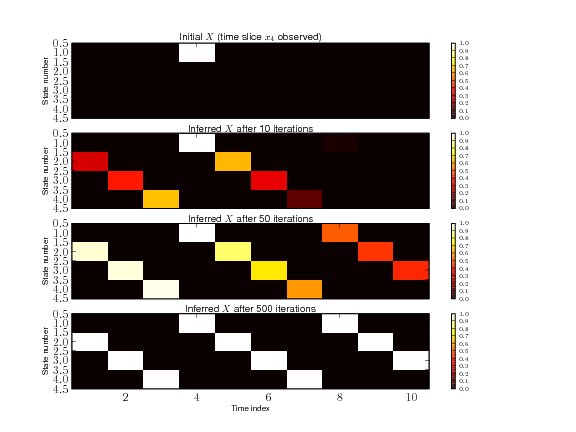}
\caption{A plot of the estimate for $X$ after various iteration counts of the inference algorithm. The top image shows the initial $X$ in which $x_4$ is observed and all other variables are hidden. Here we let $x_4$ correspond to state $S_1$ by setting $x_4 = s_1$. All other variables (i.e, $(x_1,\dots,x_3), (x_5,\dots,x_{10}), (h_1,\dots,h_{10})$) are initialized to small positive random values.  We see that convergence is effectively reached by 500 iterations.}
\label{fig:fsm1_determ_factor_X1_after_various_iterations2}
\end{figure}

Although the first time slice $x_1$ was chosen as the only observed variable above, we could have just as easily chosen any subset of time slices (or even any subset of components) of $X$ and $H$ as the observed subset $X_E$. For example, Figure~\ref{fig:fsm1_determ_factor_X1_after_various_iterations2} shows the estimates of $X$ for various iteration counts for the case where the time slice $x_4$ is observed and all other time slices are hidden. As expected, we observe that the estimates propogate outward (both backward and forward) from $x_4$ with increasing iteration count.

\subsubsection{Partially observed input sequences under a known nondeterministic transition model}

In this section we perform experiments to observe how the inference algorithm performs when multiple solutions for the hidden variables are possible. We will employ a nondeterministic state transition model and supply partial state observations so that multiple solutions will be possible for at least some of the hidden states. We will now use the nondeterministic state transition diagram in Figure~\ref{fig:fsm1}, which corresponds to the transition basis matrix $W$ from Equation~\ref{eqn:W_non_deterministic}.

\begin{figure}
\centering
\includegraphics[width=60ex]{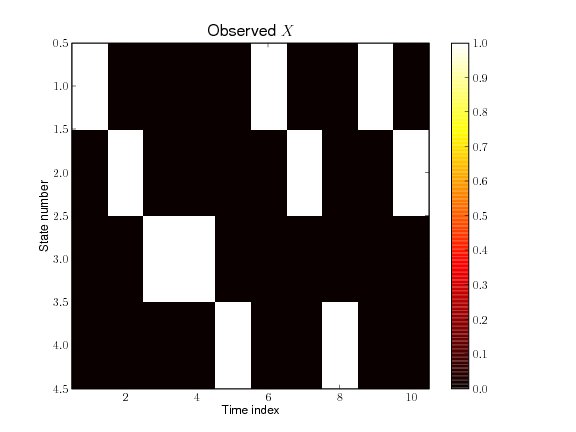}
\caption{An image plot of $X$ corresponding to the state sequence: $(S_1, S_2, S_3, S_3, S_4, S_1, S_2, S_4, S_1, S_2)$. This is a valid realization under the transition diagram in Figure~\ref{fig:fsm1}.}
\label{fig:fsm1_nondeterm_X1}
\end{figure}

\begin{figure}
\centering
\includegraphics[width=80ex]{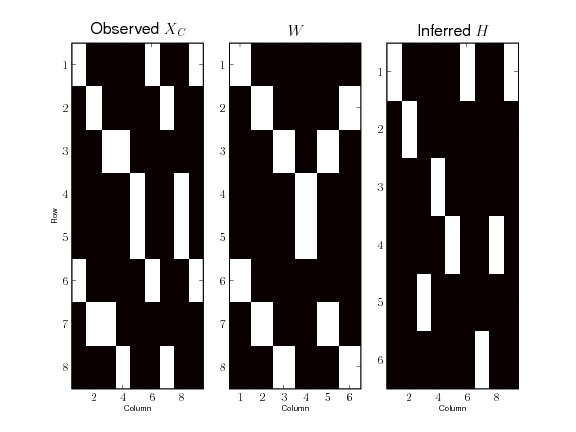}
\caption{An image plot of the matrices of the factorization $X_C = W H$, after solving for $H$. Note that $X$ is equal to the upper half submatrix of $X_C$ and corresponds to the state sequence $(S_1, S_2, S_3, S_3, S_4, S_1, S_2, S_4, S_1, S_2)$. Here, $W$ corresponds to the nondeterministic state transition diagram in Figure~\ref{fig:fsm1}}
\label{fig:fsm1_nondeterm_factor}
\end{figure}

We first perform an experiment to verify that the inferred values for the hidden transitions $H$ are correct for a short fully observed sequence of observations $X$.  Figure~\ref{fig:fsm1_nondeterm_X1} shows an image plot for  $X = \left[ \begin{array}{cccccccccc} s_1 & s_2 & s_3 & s_3 & s_4 & s_1 & s_2 & s_4 & s_1 & s_2 \end{array} \right]$, which represents a valid sequence of state transitions under the transition model in Figure~\ref{fig:fsm1}. We then perform inference on $H$ using the algorithm from Appendix~\ref{sec:simpleDyn1_inf_learn}, which is a special case of the general learning and inference algorithm specified in Algorithm~\ref{alg:inference_learning1}. The inference results are shown in Figure~\ref{fig:fsm1_nondeterm_factor} and correspond to the correct factorization.

\begin{figure}
\centering
\includegraphics[width=80ex]{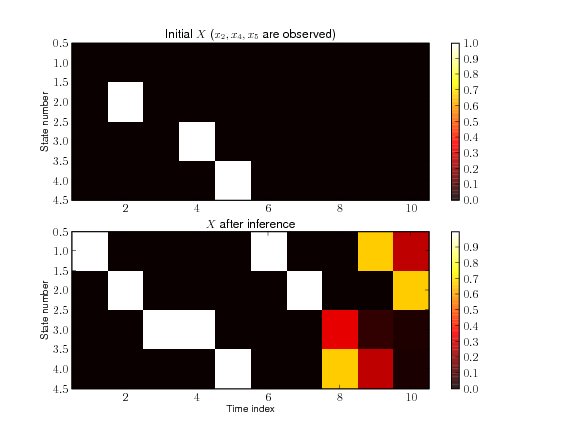}
\caption{The top image plot shows $X$ after initialization, corresponding to the partially observed state sequence: $(?, S_2, ?, S_3, S_4, ?, ?, ?, ?, ?)$, where "?" denotes a hidden state. The bottom image plot shows the inference results for $X$ in which the hidden variables have now been estimated. Observe that variables ${x_1, x_3, x_6, x_7}$ correspond to exactly one state and variables $x_8,x_9,x_{10}$ correspond to a superposition of states, since multiple solutions are possible for $x_8,x_9,x_{10}$.}
\label{fig:fsm1_nondeterm2_X1}
\end{figure}

\begin{figure}
\centering
\includegraphics[width=80ex]{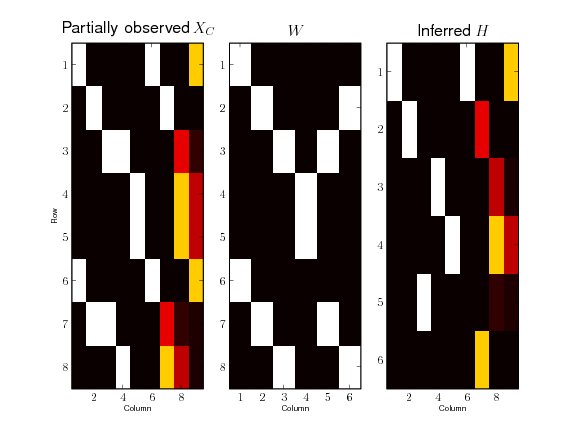}
\caption{An image plot of the matrices of the factorization $X_C = W H$, after solving for both $H$ and the hidden variables of $X$.}
\label{fig:fsm1_nondeterm2_factor}
\end{figure}

We now consider a partially observed sequence of state observations for $X$. Let

\begin{align}
X = \left[ \begin{array}{cccccccccc}? & s_2 & ? & s_3 & s_4 & ? & ? & ? & ? & ?\end{array} \right]
\label{eqn:x1PartiallyObserved1}
\end{align}

where the "?" state vectors denotes a hidden state. We initialize $H$ and all hidden variables in $X$ to small positive random values and run the inference algorithm to convergence. Figure~\ref{fig:fsm1_nondeterm2_X1} shows $X$ before and after performing inference. An exact factorization is found and is shown in Figure~\ref{fig:fsm1_nondeterm2_factor}. We observed that repeated runs always converged to an exact solution, although different (but valid) result was obtained each time. Observe that some of the hidden variables correspond to exactly one state, while others correspond to a superposition of states, such that the column sums are equal. We do not explicitly normalize the columns to have equal column sums in the inference procedure, but rather, the equality follows from the choice of $W$ and the observed variables. Looking at the transition diagram in Figure~\ref{fig:fsm1}, we see that the inferred variables with a single state component (i.e., $x_t: t \in  \{1, 3, 6, 7\}$) occur at the time slices for which there is exactly one possible state that satisfies the transition model, given the observed variables in $X$. For example, time slice 2 has an observed state $x_2 = s_2$, corresponding to $S_2$. From the transition diagram we see that the only state that can transition to $S_2$ is $S_1$ and therefore the hidden state at time slice 1 must correspond to state $S_1$. The variables with multiple distinct state components (i.e., $x_t, t \in  \{ 8,9,10\}$) correspond to time slices for which there are multiple possible transitions that satisfy the transition model. We can think of these time slices simultaneously being in multiple distinct states. For example, a valid solution for hidden state $x_8$ can consist of any additive combination of state vectors $s_3, s_4$ such that the column sum of $x_8$ is 1.

\begin{figure}
\centering
\includegraphics[width=80ex]{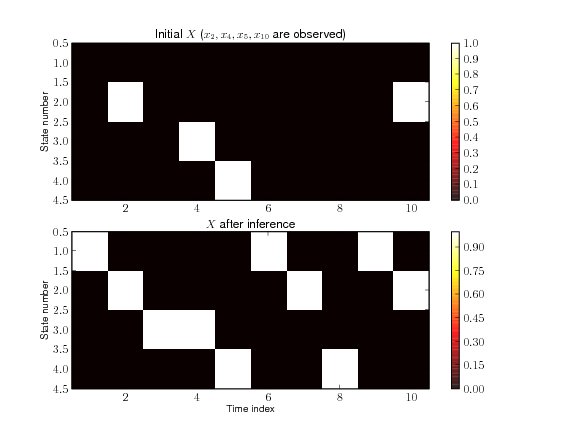}
\caption{The top image plot shows $X$ after initialization, corresponding to the partially observed state sequence: $(?, S_2, ?, S_3, S_4, ?, ?, ?, ?, S_2)$, where "?" denotes a hidden state. The bottom image plot shows $X$ again after running the inference algorithm to infer the hidden states. In this case, there is only one possible configuration of hidden states that is constant with the observed states and with the underlying transition model.}
\label{fig:fsm1_nondeterm3_X1}
\end{figure}

\begin{figure}
\centering
\includegraphics[width=80ex]{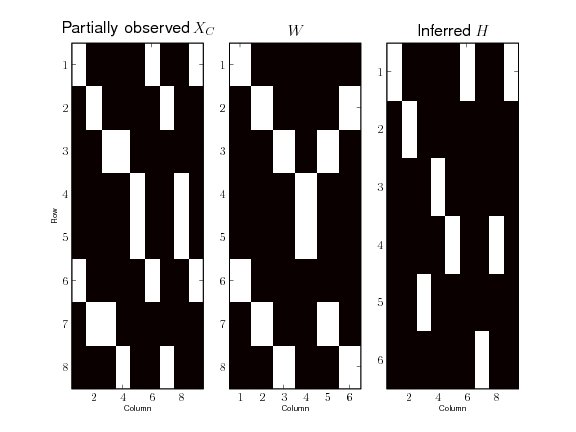}
\caption{An image plot of the matrices of the factorization $X_C = W H$, after solving for $H$. This is for the case where $X$ corresponds to the paritally observed sequence: $(?, S_2, ?, S_3, S_4, ?, ?, ?, ?, S_2)$. In this case, there is only one possible configuration of hidden states that is constant with the observed states and with the underlying transition model.}
\label{fig:fsm1_nondeterm3_factor}
\end{figure}

Suppose that we now make the final time slice observed by setting $x_{10} = s_2$, corresponding to the partially observed sequence $(?, S_2, ?, S_3, S_4, ?, ?, ?, ?, S_2)$. In this case, there is only one possible configuration of hidden states that is constant with the observed states and with the underlying transition model. Figures~\ref{fig:fsm1_nondeterm3_X1} and \ref{fig:fsm1_nondeterm3_factor} show the corresponding inference results. We observed that repeated runs of the algorithm always converged to the correct solution.

\begin{figure}
\centering
\includegraphics[width=80ex]{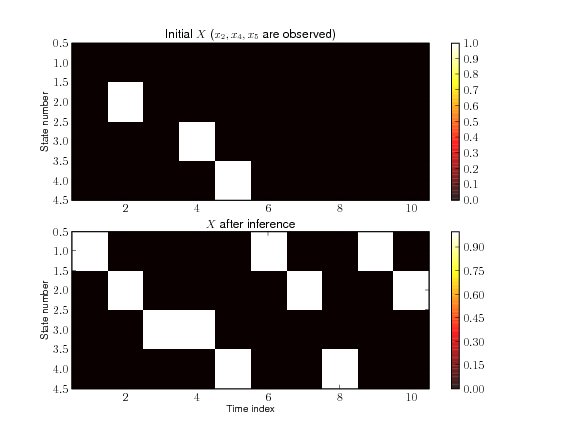}
\caption{The top image plot shows $X$ after initialization, corresponding to the partially observed state sequence: $(?, S_2, ?, S_3, S_4, ?, ?, ?, ?, ?)$, where "?" denotes a hidden state. The bottom image plot shows $X$ again after running the inference algorithm to infer the hidden states. Sparse NMF with sparseness = 0.1 was used. We observe that multiple solutions are possible (since $x_{10}$ is now hidden), but the sparsity constraint leads to solutions in which only 1 state is active in any given time slice.}
\label{fig:fsm1_nondeterm4_X1}
\end{figure}

\begin{figure}
\centering
\includegraphics[width=80ex]{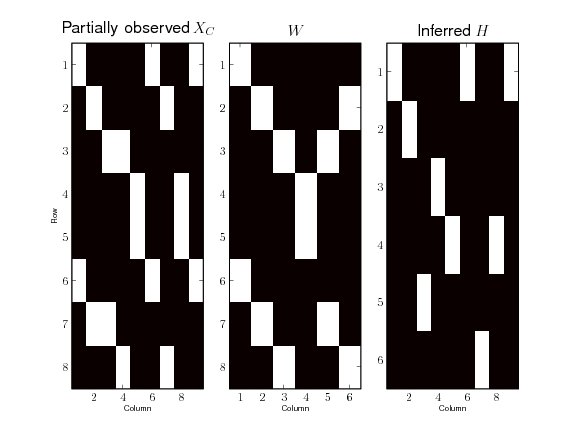}
\caption{An image plot of the matrices of the factorization $X_C = W H$, after solving for $H$. Note that $X$ is equal to the upper half submatrix of $X_C$ and corresponds to the state sequence $(?, S_2, ?, S_3, S_4, ?, ?, ?, ?, ?)$.  Sparse NMF with sparseness = 0.1 was used. We observe that multiple solutions are possible  (since $x_{10}$ is now hidden), but the sparsity constraint leads to solutions in which only 1 state is active in any given time slice.}
\label{fig:fsm1_nondeterm4_factor}
\end{figure}

In the case where multiple valid solutions are possible for the hidden variables, it is interesting to consider modifying the inference algorithm to attempt to find sparse solutions. One possibility that is simple to implement and seems to provide robust solutions consists of replacing the standard NMF update steps of the basic inference algorithm specified in Algorithm~\ref{alg:inference_learning1} with corresponding sparse NMF update steps. For these experiments, we chose to use the nonsmooth non-negative matrix factorization (nsNMF) algorithm \cite{nsNMF2006} which is described in Appendix~\ref{appendix_sparse_nmf} for reference. Using the partially observed sequence from Equation~\ref{eqn:x1PartiallyObserved1} and the modified inference algorithm with an nsNMF sparseness value of 0.1, we obtain the results shown in Figures~\ref{fig:fsm1_nondeterm4_X1} and \ref{fig:fsm1_nondeterm4_factor}. Repeated runs of the algorithm appear to produce distinct solutions such that each inferred hidden state $x_t$ has only a single nonzero component, corresponding to a single active state in any given time slice. For sparseness values much lower than 0.1, we observe that a solution consisting of a superposition of states can still occur, but most of the weight tends to be concentrated in a single state for each time slice.

\begin{figure}
\centering
\includegraphics[width=80ex]{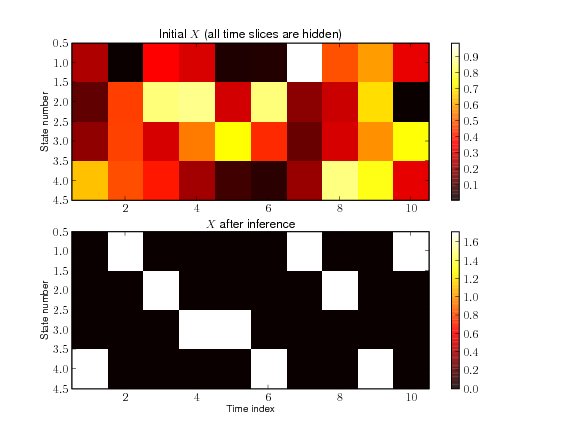}
\caption{The top image plot shows $X$ after being initialized to random positive values. The bottom image plot shows $X$ again after convergence of the inference algorithm, using sparse NMF with sparseness = 0.1}
\label{fig:fsm1_nondeterm5_X1}
\end{figure}

\begin{figure}
\centering
\includegraphics[width=80ex]{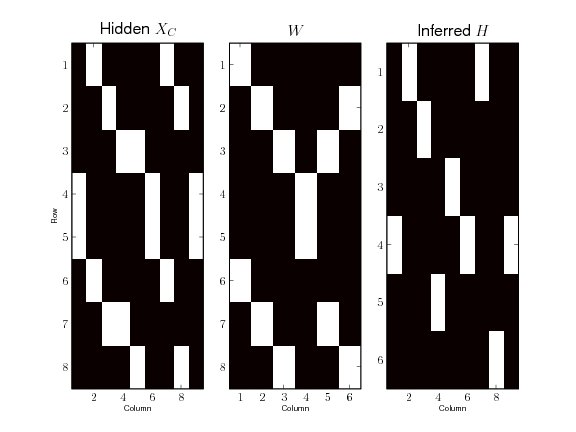}
\caption{An image plot of the matrices of the factorization $X_C = W H$, after solving for $X$ and $H$, which were initialized to random positive values. Even though all model variables are hidden, the sparse NMF updates caused the inference algorithm to converge to a sparse solution corresponding to an valid state sequence under the state transition diagram in Figure~\ref{fig:fsm1} in which only a single state is active in any given time slice.}
\label{fig:fsm1_nondeterm5_factor}
\end{figure}

We now perform an experiment to see what will happen if all model variables are made hidden (i.e., $X$ and $H$ are initialized to positive random values), and we perform inference using the sparse NMF updates. The top plot of Figure~\ref{fig:fsm1_nondeterm5_X1} shows $X$ after being initialized to random positive values uniformly distributed between 0 and 1. The bottom plot shows the result of one run of the inference algorithm using a nsNMF sparseness value of 0.1. Figure~\ref{fig:fsm1_nondeterm5_factor} shows the resulting factorization, which converged to an exact solution. We observe that repeated runs of the inference algorithm produce factorizations corresponding to a randomly sampling of the underlying transition model. This interesting result only appears to occur when we use the sparse inference algorithm. The non-sparse algorithm also converges to an exact factorization, but the inferred $X$ and $H$ then tend to consist of a superposition of many states in any given time slice.

\subsubsection{Learning the transition model from training data}
\label{sec:learn_tran_model_from_data}

We now attempt to learn a transition model from training data. The training data will consist of an observed sequence $X$, and $H$ will be hidden. The model parameters $W$ are therefore unknown and will be learned from the observed sequence $X$. We first consider the case where the input sequence $X$ is an elementary sequence, so that any two adjacent state vectors $x_t, x_{t+1}$ can be represented by a single transition basis vector in $W$. That is, any state vector $x_t$ contains only one nonzero component and therefore corresponds to exactly one state in the transition model. We let $X$ consist of a sequence of state basis vectors, such as:

\begin{align}
X = \left[ \begin{array}{ccccc}s_1 & s_2 & s_3 & s_3 & \dots \end{array} \right]
\end{align}

where the states are constrained to evolve according to the transition diagram in Figure~\ref{fig:fsm1}. The training sequence is generated as follows: We choose the initial state $x_1$ randomly with uniform probability. When multiple outgoing transitions from a state are possible, the next state vector is chosen randomly with equal probability from the set of allowable transitions. For all experiments in this section, we use a training sequence length of 1000. 



\begin{figure}
\centering
\includegraphics[width=80ex]{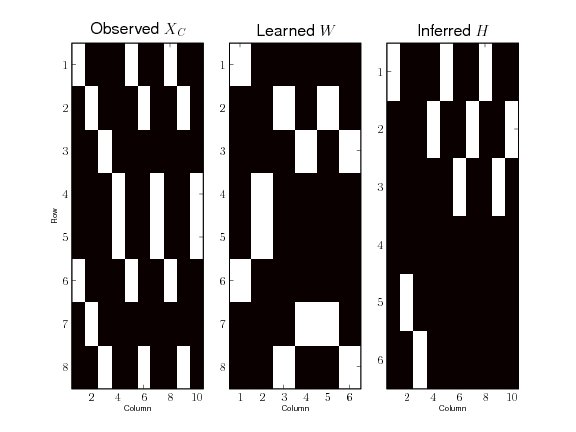}
\caption{An image plot of the matrices of the factorization $X_C = W H$, after learning $W$ and solving for $H$, which were initialized to random positive values. $W$ was constrained to have 6 transition basis vectors. The training sequence $X$ consisted of state basis vectors that evolved according to the transition diagram in Figure~\ref{fig:fsm1}.}
\label{fig:fsm1_learnW_1component_6basis_factor}
\end{figure}

\begin{figure}
\centering
\includegraphics[width=80ex]{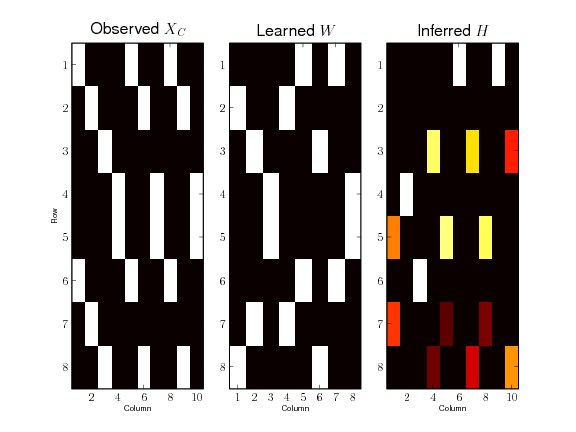}
\caption{An image plot of the matrices of the factorization $X_C = W H$, after learning $W$ and solving for $H$, which were initialized to random positive values. $W$ was constrained to have 6 transition basis vectors. Since there are two more columns than necessary to represent the transition model, observe that there are two duplicate columns in the learned $W$. The training sequence $X$ consisted of state basis vectors that evolved according to the transition diagram in Figure~\ref{fig:fsm1}.}
\label{fig:fsm1_learnW_1component_8basis_factor}
\end{figure}

Recall that inference and learning is performed jointly, so that in the process of learning $W$, we also end up with an estimate for $H$ (the transition activations for the training sequence). The column count of $W$ is a free parameter that must be specified before performing learning. Since we already know the transition model that was used to generate the training data, we know that we must specify at least 6 columns for $W$ in order to have any hope of achieving an exact factorization. We will therefore specify that $W$ have 6 columns. We then run the inference and learning algorithm from Appendix~\ref{sec:simpleDyn1_inf_learn} to convergence. Figure~\ref{fig:fsm1_learnW_1component_6basis_factor} shows the inference and learning results on a length 10 sub-sequence of $X$. Observe that the learned $W$ contains the 6 correct transition basis vectors. However, we have observed that it can sometimes take a few runs of the algorithm in order to find an exact factorization. When the column count of $W$ was then increased slightly to 8, we found that the inference and learning algorithm always converged to an exact factorization. Figure~\ref{fig:fsm1_learnW_1component_8basis_factor} shows the inference and learning results for a $W$ with 8 columns. Note that although the transition basis vectors are correct, there are now some duplicate columns in $W$. This is due to the normalization step in which the columns of $W$ are normalized to have unit sum as part of the NMF left (learning) update step. 

\begin{figure}
\centering
\includegraphics[width=80ex]{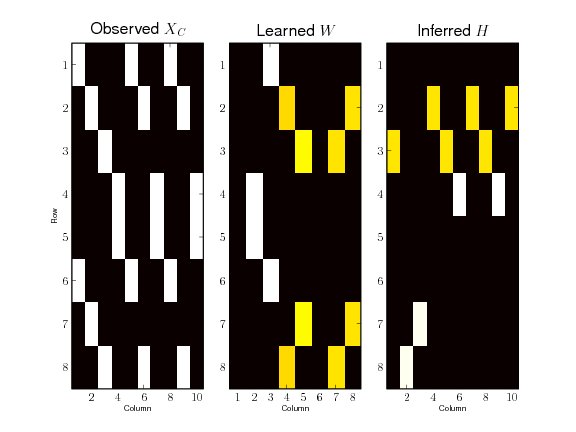}
\caption{An image plot of the matrices of the factorization $X_C = W H$, after learning $W$ and solving for $H$. $W$ was constrained to have 6 transition basis vectors. There are two more columns than necessary to represent the transition model, but we remove the normalization step from the $W$ update and set the nsNMF sparseness value to 0.1. The learning algorithm then learns only the 6 required columns. That is, we see that two columns of the learned $W$ are zero-valued.}
\label{fig:fsm1_learnW_1component_8basis_factorSparse}
\end{figure}

We have observed that the number of duplicate columns in $W$ can be reduced if we use sparse NMF updates in the inference and learning algorithm and also remove the column sum normalization step from the NMF left update step. We instead normalize each column of $W$ so the the upper and lower sub-columns (corresponding to $W_1$ and $W_2$) are constrained have equal sum, which may be 0. This allows unneeded columns of $W$ to tend towards zero during the learning updates. Using an nsNMF sparseness value of 0.1, we obtain the results in Figure~\ref{fig:fsm1_learnW_1component_8basis_factorSparse}. We observe that even though it was specified that $W$ have two more columns than required to represent the transition model, the learned $W$ contains only the required 6 columns and two zero-valued columns. Note also that since we have omitted the step of normalizing all columns to have sum equal to one, the nonzero columns of $W$ no longer sum to 1. If we wish for the nonzero columns to sum to the same value (e.g., 1), we could consider adding column normalization for the nonzero columns as a post-processing step. We have also observed that the use of the nsNMF algorithm leads to a small approximation error so that the resulting factorization is no longer exact.

\begin{figure}
\centering
\includegraphics[width=80ex]{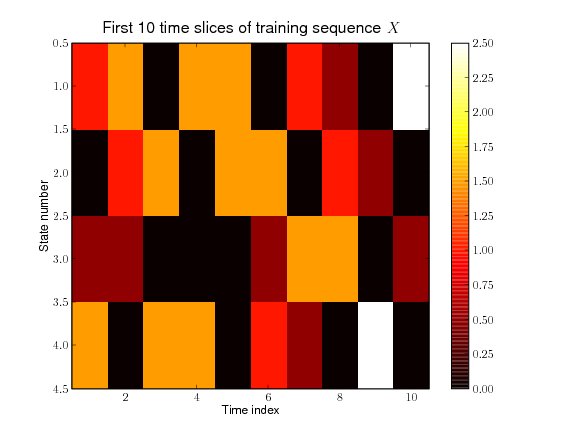}
\caption{An image plot showing the first 10 time slices of the length 1000 training sequence $X$ consisting of an additive mixture of 3 elementary state transition sequences.}
\label{fig:fsm1_learnW_mixture_component_X1}
\end{figure}

\begin{figure}
\centering
\includegraphics[width=80ex]{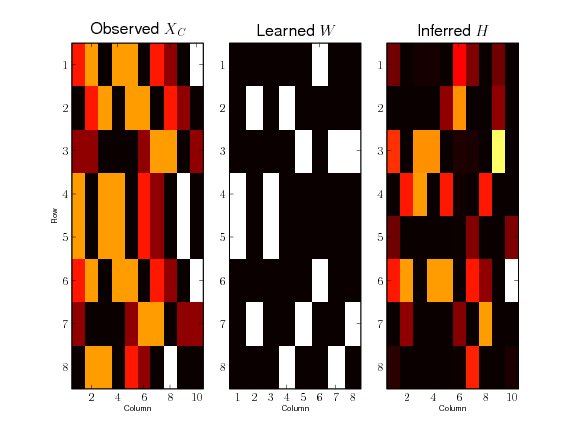}
\caption{An image plot of the matrices of the factorization  $X_C = W H$, after learning $W$ and solving for $H$ where the $X$ training sequence consists of an additive mixture of three state transition sequences. $W$ was specified to have 8 transition basis vectors. The underlying transition model used to generate the training data corresponds to the transition diagram in Figure~\ref{fig:fsm1}.}
\label{fig:fsm1_learnW_mixture_component_factor}
\end{figure}

We now consider the more challenging learning problem in which the training sequence consists of an additive mixture of realizations of the transition model in Figure~\ref{fig:fsm1}. We construct the training sequence $X$ as the following additive mixture of three scaled elementary state transition realization sequences:

\begin{align}
X = 0.5*Xelem_1 + 1.0*Xelem_2 + 1.5*Xelem_3
\end{align}

where the $Xelem_i$ are independent elementary state sequences, each consisting of a sequence of state basis vectors that conform to the transition model in Figure~\ref{fig:fsm1}. Figure~\ref{fig:fsm1_learnW_mixture_component_X1} shows the first 10 time slices of the training sequence. Note that each time slice of $X$ now corresponds to a state vector that represents a superposition of three states. Given this training sequence, we perform inference and learning using 8 columns for $W$. The algorithm converges to an exact factorization, which is shown in Figure~\ref{fig:fsm1_learnW_mixture_component_factor} and we see that the underlying transition model was learned, even though the training data consisted of an additive mixture of elementary state sequences. Thus, even though a realization of the transition model was not presented individually as training data, the model still was able to learn an underlying transition model that explained the mixture training sequence.

\begin{figure}
\centering
\includegraphics[width=80ex]{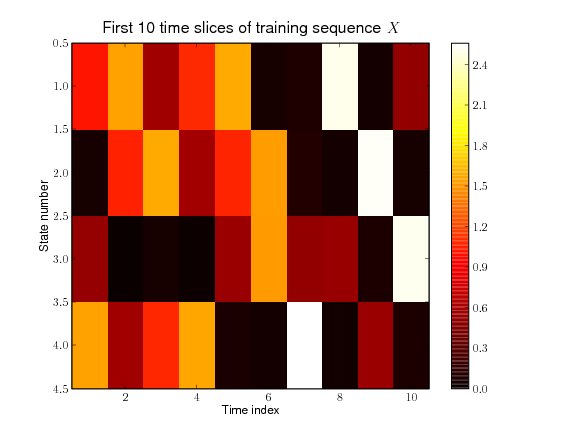}
\caption{An image plot showing the first 10 time slices of the length 1000 training sequence $X$ consisting of an additive mixture of 3 elementary state transition sequences and a noise component.}
\label{fig:fsm1_learnW_mixture_noise_component_X1}
\end{figure}

\begin{figure}
\centering
\includegraphics[width=80ex]{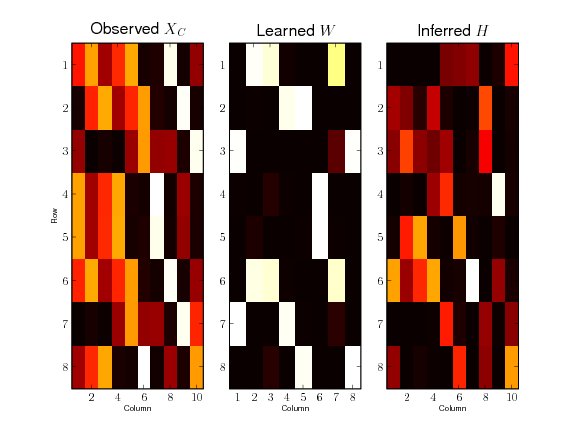}
\caption{An image plot of the matrices of the matrices of the factorization $X_C = W H$, after learning $W$ and solving for $H$ where the training sequence $X$ consists of an additive mixture of three state transition sequences plus a noise component. $W$ was specified to have 8 transition basis vectors.  The underlying transition model used to generate the training data corresponds to the transition diagram in Figure~\ref{fig:fsm1}.}
\label{fig:fsm1_learnW_mixture_noise_component_factor}
\end{figure}

We now add a noise component to the training sequence so that:

\begin{align}
X = 0.5*Xelem_1 + 1.0*Xelem_2 + 1.5*Xelem_3 + \epsilon
\end{align}

where $\epsilon$ is a 4 x $L$ matrix of uniformly distributed noise between 0 and 0.1. Figure~\ref{fig:fsm1_learnW_mixture_noise_component_X1} shows a portion of the noisy training sequence. The inference and learning algorithm converges to the approximate factorization shown in Figure~\ref{fig:fsm1_learnW_mixture_noise_component_factor}. We see that the underlying transition model was still recovered in $W$, although with a small error. These experiments indicate that it is possible to learn an underlying transition model, given only noisy training data corresponding to an additive mixture of realizations of the underlying model.

\section{Hierarchical factored state model}
\label{sec:main_hierarchical_state_model}

In this section we present DPFNs for modeling sequential data with hierarchical structure. The general DPFN framework allows for the representation of dynamical models with complex hierarchical structure. We present multilevel hierarchical networks in which only certain combinations of upper level and lower level variables and state transitions may be simultaneously active. We show how such networks provide for the modeling of factored hierarchical state transition models.

The models that we present in this section can be thought of as two copies of the network from Figure Figure~\ref{fig:1layerDynamic} to represent multiple levels in a hierarchy, along with an additional coupling module to enforce desired coupling constraints between pairs of variables within a time slice. As a concrete example, we present a hierarchical state model for a regular expression and present empirical results.

\subsection{Model}
\label{sec:dyn2level}
\begin{figure}
\centering
\includegraphics[width=60ex]{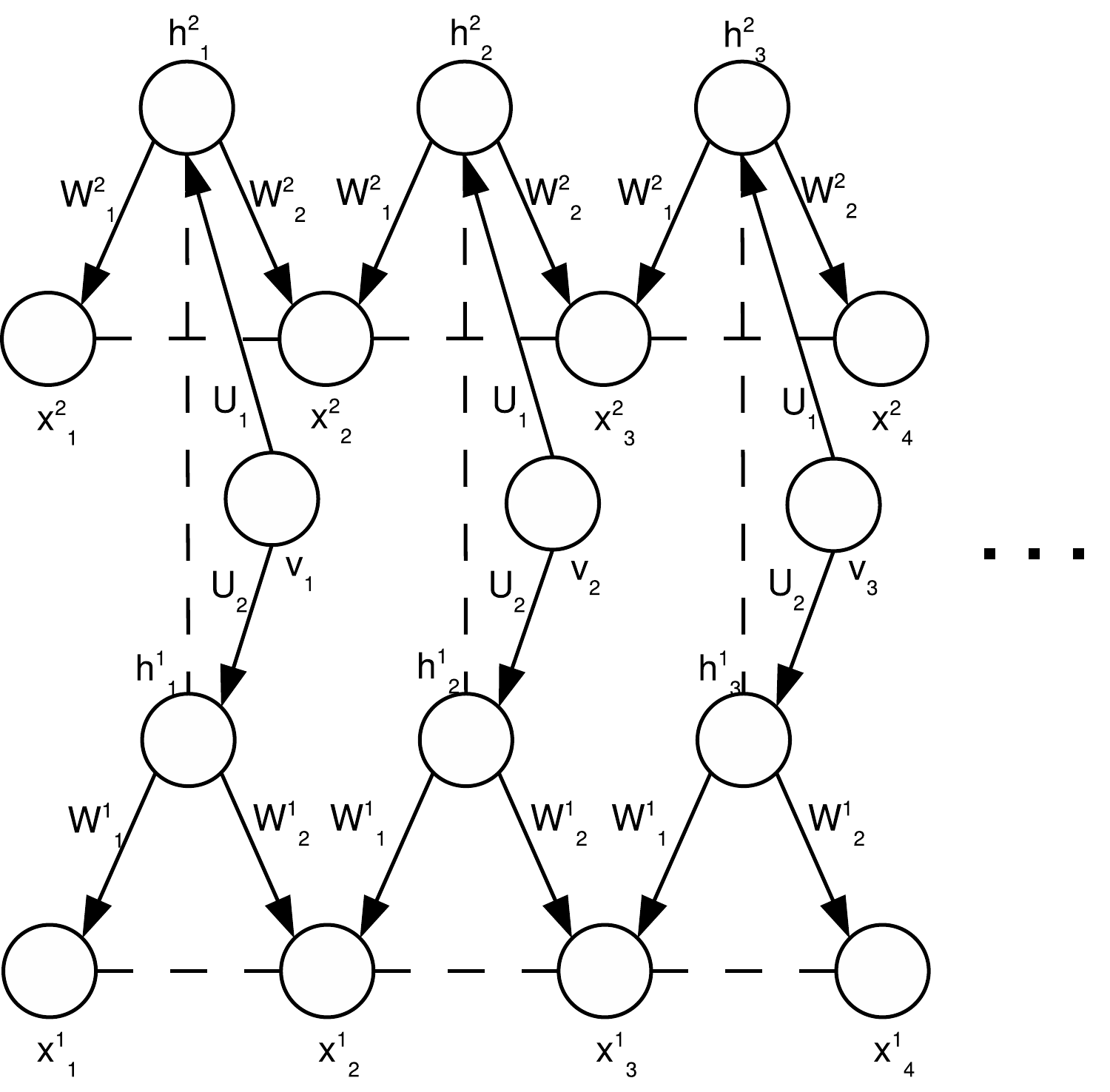}
\caption{A DPFN with a coupled 2-level transition model. The level 2 state variables are denoted by $x^2_t$ and the level 1 state variables are denoted by $x^1_t$. The level 2 transition variables $h^2_t$ are coupled to the level 1 transition variables $h^1_t$. The first four time slices of the $T$ time slice network are shown. The dashed lines represent (optional) forced factored co-activations, which are enforced if all sub-columns of the parameter matrices are normalized to have equal column sums.}
\label{fig:2layerDynamic}
\end{figure}

Consider a 2 level network in which each level represents a state transition model. One can think of each of the levels as having a corresponding state transition diagram. The active state transitions in the 2 levels are then coupled so that only certain combinations of level 2 and level 1 state transitions may be simultaneously active. Figure~\ref{fig:2layerDynamic} shows a 2-level network for modeling a factored hierarchical state model with two levels in the transition hierarchy. The model immediately extends to an arbitrary number of levels. The level 2 state variables are denoted by \{$x^2_t: t = 1, \dots, T$\} and the level 1 state variables are denoted by $\{x^1_t: t=1, \dots, T\}$. The level 2 transition variables $\{h^2_t: t=1, \dots, T-1 \}$ are coupled to the level 1 transition variables $\{h^1_t: t = 1, \dots, T -1\}$ via the coupling variables $\{v_t: t = 1, \dots, T -1\}$.

Note the symmetry between levels 1 and 2. One can think of the level 1 and level 2 transition models as executing independently, except for where the coupling constraints forbid it by constraining the combinations of level 1 and level 2 state transitions that can be simultaneously activated. Any two consecutive time slices ($t, t+1$) of Figure~\ref{fig:2layerDynamic} correspond to the following three factorizations:

For level 1, we have:
\begin{align}
\left[ \begin{array}{c}
x^1_t \\
x^1_{t+1} \end{array} \right] =&\ W^1 h^1_t
\nonumber \\
 =&\ \left[ \begin{array}{c}
W^1_1 \\
W^1_2 \end{array} \right] h^1_t
\end{align}

For level 2, we have:
\begin{align}
\left[ \begin{array}{c}
x^2_t \\
x^2_{t+1} \end{array} \right] =&\ W^2 h^2_t
\nonumber \\
 =&\ \left[ \begin{array}{c}
W^2_1 \\
W^2_2 \end{array} \right] h^2_t
\end{align}

For the level 1 to level 2 transition coupling, we have:
\begin{align}
\left[ \begin{array}{c}
h^2_t \\
h^1_t \end{array} \right] =&\ U v_t
\nonumber \\
 =&\ \left[ \begin{array}{c}
U_1 \\
U_2 \end{array} \right] v_t
\end{align}

For $T$ time slices, we then have the  following three matrix factorization euqations:

For level 1, we have:
\begin{align}
\left[ \begin{array}{ccccc}
x^1_1 & x^1_2 & x^1_3 & \dots & x^1_{T-1} \notag \\
x^1_{2} & x^1_{3} & x^1_{4} & \dots & x^1_{T} \end{array} \right] =&\ W^1 \left[ \begin{array}{ccccc} h^1_1 & h^1_2 & h^1_3 & \dots & h^1_{T-1} \end{array} \right] \\
=&\ \left[ \begin{array}{c}
W^1_1 \\
W^1_2 \end{array} \right] \left[ \begin{array}{ccccc} h^1_1 & h^1_2 & h^1_3 & \dots & h^1_{T-1} \end{array} \right]
\end{align}

which we can write more concisely as:

\begin{align}
\label{eq:dynamic1_fact1}
X^1_c =& W^1 H^1 \notag \\
=& \left[ \begin{array}{c}
W^1_1 \\
W^1_2 \end{array} \right] H^1
\end{align}

For level 2, we have:
\begin{align}
\left[ \begin{array}{ccccc}
x^2_1 & x^2_2 & x^2_3 & \dots & x^2_{T-1} \notag \\
x^2_{2} & x^2_{3} & x^2_{4} & \dots & x^2_{T} \end{array} \right] =&\ W^2 \left[ \begin{array}{ccccc} h^2_1 & h^2_2 & h^2_3 & \dots & h^2_{T-1} \end{array} \right] \\
=&\ \left[ \begin{array}{c}
W^2_1 \\
W^2_2 \end{array} \right] \left[ \begin{array}{ccccc} h^2_1 & h^2_2 & h^2_3 & \dots & h^2_{T-1} \end{array} \right]
\end{align}

which we can write more concisely as:

\begin{align}
\label{eq:dynamic1_fact2}
X^2_c =& W^2 H^2 \notag \\
=& \left[ \begin{array}{c}
W^2_1 \\
W^2_2 \end{array} \right] H^2
\end{align}

For the level 1 to level 2 transition coupling, we have:
\begin{align}
\left[ \begin{array}{ccccc}
h^2_1 & h^2_2 & h^2_3 & \dots & h^2_{T-1} \notag \\
h^1_1& h^1_3 & h^1_4 & \dots & h^1_{T-1} \end{array} \right] =&\ U \left[ \begin{array}{ccccc} v_1 & v_2 & v_3 & \dots & v_{T-1} \end{array} \right] \\
=& \left[ \begin{array}{c}
U_1 \\
U_2 \end{array} \right] \left[ \begin{array}{ccccc} v_1 & v_2 & v_3 & \dots & v_{T-1} \end{array} \right]
\end{align}

Defining $V = \left[ \begin{array}{ccccc} v_1 & v_2 & v_3 & \dots & v_{T-1} \end{array} \right]$, we can then write the above factorization more concisely as:

\begin{align}
\label{eq:dynamic1_fact3}
\left[ \begin{array}{c}
H^2 \\
H^1 \end{array} \right] =& U V \notag \\
=& \left[ \begin{array}{c}
U_1 \\
U_2 \end{array} \right] V
\end{align}

The columns of $U$ define the coupling basis. The positive components of each column $u_i$ specify components of $h^2_t$ and $h^1_t$ that can be co-activated. That is, the stacked vector of $h^2_t$ and $h^1_t$ is representable as a non-negative combination of the columns of $U$, with $v_t$ giving the encoding in terms of the basis columns. A learning and inference algorithm for this network can be obtained by applying Algorithm~\ref{alg:inference_learning1} to the system of factorizations (\ref{eq:dynamic1_fact1}),(\ref{eq:dynamic1_fact2}), and (\ref{eq:dynamic1_fact3}). Example pseudocode is presented in Appendix~\ref{sec:inf_learn_dyn2level}.

\begin{figure}
\centering
\includegraphics[width=60ex]{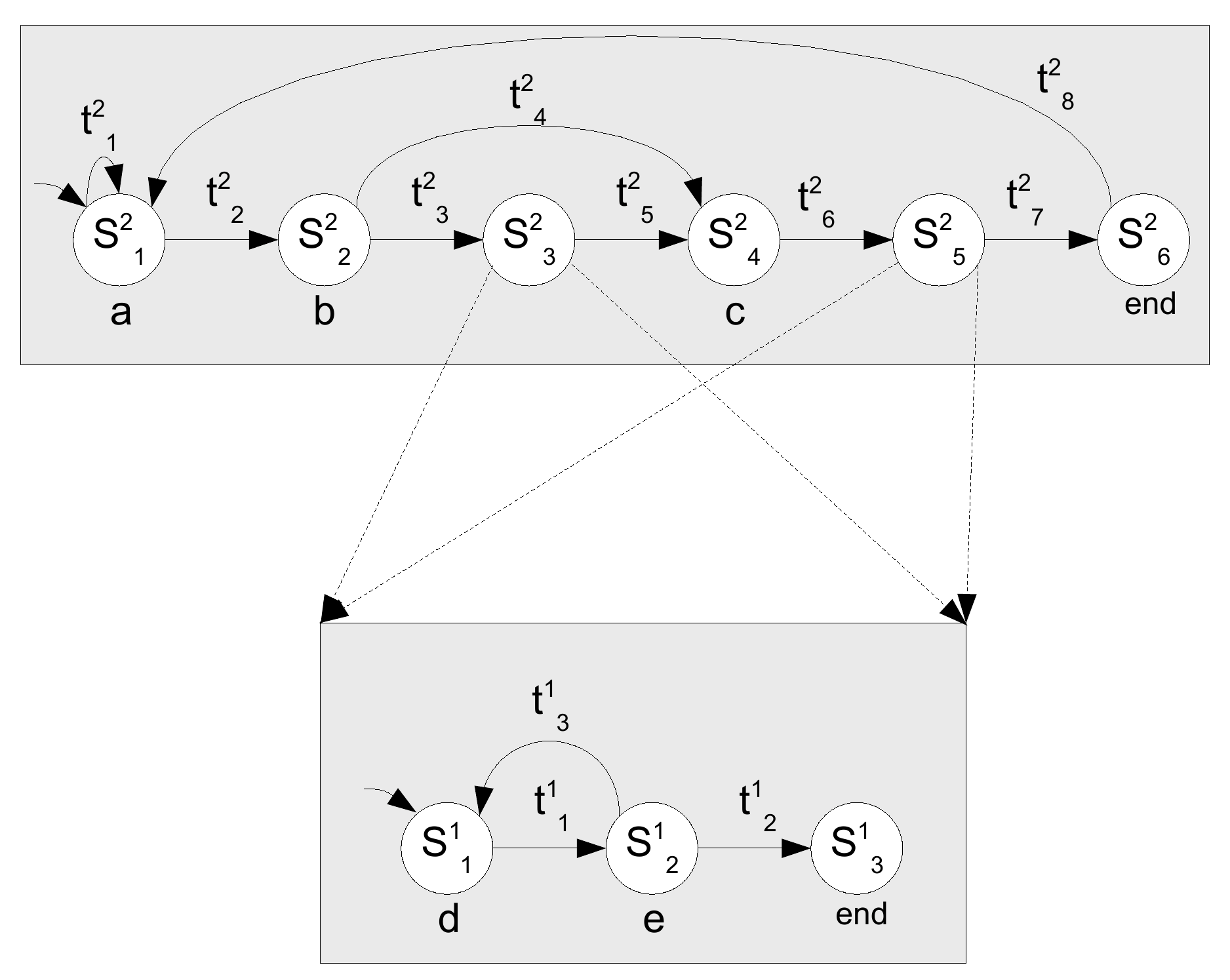}
\caption{A state transition diagram for the regular expression ``a+b(de)*c(de)+''.}
\label{fig:hfsm1_crop}
\end{figure}
\begin{figure}
\centering
\includegraphics[width=60ex]{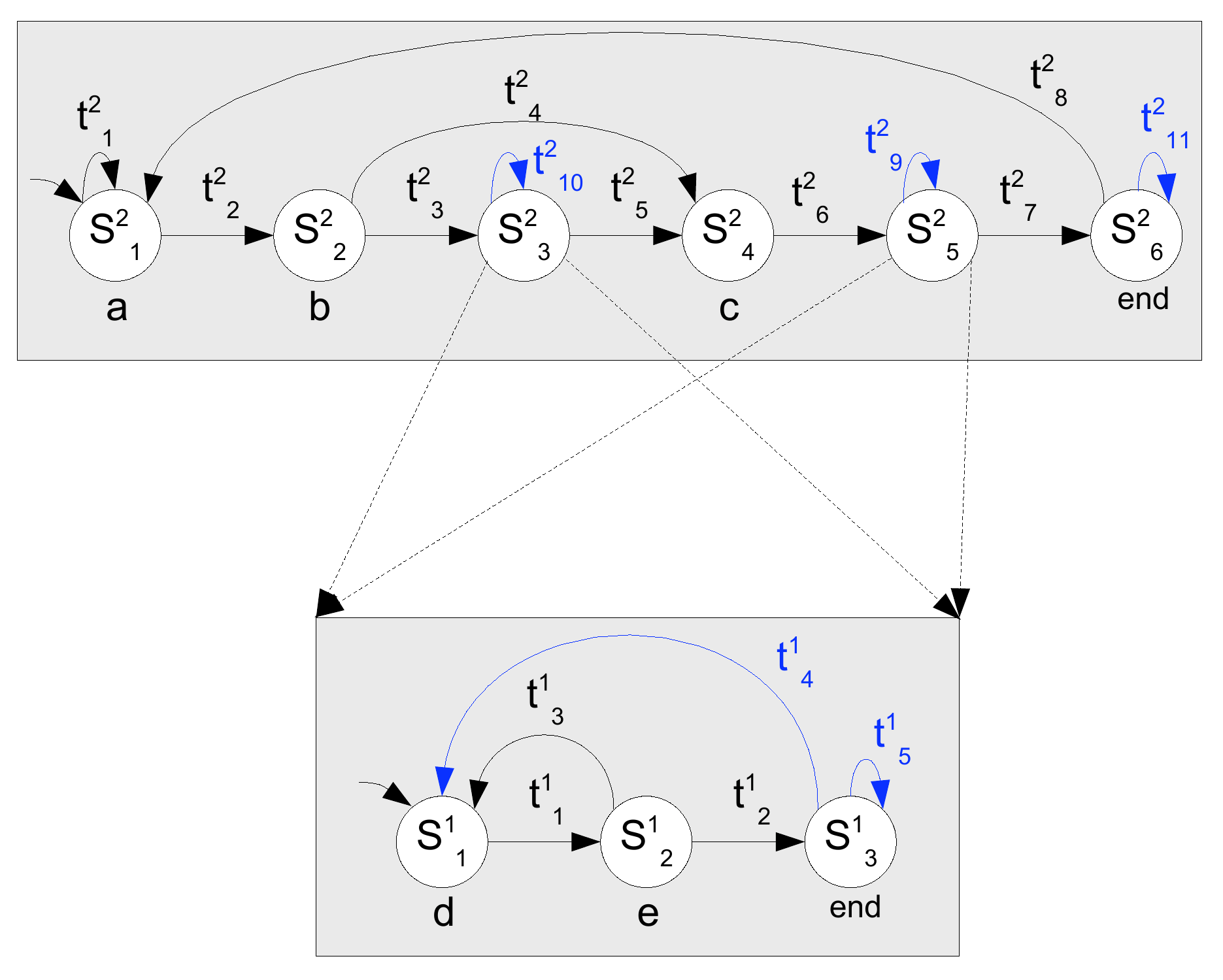}
\caption{The state transition diagram for the regular expression ``a+b(de)*c(de)+''. The implicit transitions are shown in blue.}
\label{fig:hfsm1_detailed_crop}
\end{figure}

We now consider a regular expression example similar in spirit to the one used by Murphy in \cite{Murphy_Thesis}. Murphy showed how a Hierarchical hidden Markov model (HHMM) \cite{murphy01linear} could be used to model the hierarchical transition model corresponding to a regular expression. In this section, we show how a regular expression can be modeled in a somewhat analogous manner by using DPFN with hierarchical structure. We stress that there are many significant differences, however. For example, in a HHMM, each possible state transition is assigned a probability, which is not the case here for the DPFN. Also, in a HHMM, the hidden states are discrete-valued, whereas they are non-negative continuous-valued here. Figure~\ref{fig:hfsm1_crop} shows a state transition diagram for the DPFN in Figure~\ref{fig:2layerDynamic} that models the regular expression ``a+b(de)*c(de)+''. Under this transition model, an exact state factorization is only possible if the input sequence satisfies the given regular expression. This regular expression specifies that we have one or more repetitions of ``a'' followed by ``b'' followed by zero or more repetitions of ``de'' followed by ``c'' followed by one or more repetitions of ``de.'' We allow the expression to repeat after reaching the end state. Observe that states $S^2_3$ and $S^2_5$ of the top level transition model (which we refer to as the level 2 transition model) refine to the same lower level transition model (the level 1 transition model). Some states (the ones with letters under them) are \emph{production states} that correspond to the production of the corresponding letters. The initial state is $S^2_1$, and states $S^2_6$ and $S^1_3$ denote explicit ends states. Although explicit end states are used here, they are not needed in general, as we will see in the next section. The semantics is such that when a transition to a refining state, such as $S^2_3$ or $S^2_5$ occurs, the refinement executes while the parent remains in the same state. We will choose the basis vectors for the coupling activations $U$ so that at any time step, the sum of active state values in level 2 is equal to the corresponding sum of active state values in level 1. For example, the transition $t^2_3$ correspond to a transition from $S^2_2$ to $S^2_3$ in level 2 and a simultaneous transition from the off/end state to the starting state $S^1_1$ in level 1. The next transition must correspond to a transition from $S^1_1$ (production of ``d'' to $S^1_2$ (production of ``e''), while level 2 remains in the parent state $S^2_3$. The hierarchical transition diagram places implicit constraints on transitions that may simultaneously occur (co-occur) in levels 1 and 2. For example, the level 2 transition $t^2_5$ can co-occur with the level 1 transition $t^1_2$. An implicit self-loop transition from $S^2_3$ to $S^2_3$ can co-occur with level 1 transitions $t^1_1$ and $t^1_3$. Figure~\ref{fig:hfsm1_detailed_crop} shows the same transition diagram, annotated to include the implicit transitions. Table~\ref{table:trans_coupling} shows the transition coupling constraints that specify the pairs of transitions in levels 1 and 2 that can co-occur. For example, note that when the refinement is not producing a ``d'' or ``e'' the end state remains in a self-loop.

\begin{table}
\begin{center}
\begin{tabular}{| c ||c | c | c | c | c | c | c | c | c | c | c | c | c |}
    \hline
    Level 2 transition & $t^2_3$ & $t^2_{10}$ & $t^2_{10}$ & $t^2_5$ & $t^2_6$ & $t^2_9$ & $t^2_9$ & $t^2_7$ & $t^2_{11}$ & $t^2_8$ & $t^2_1$ & $t^2_2$ & $t^2_4$ \\ \hline
    Level 1 transition & $t^1_4$ & $t^1_1$ & $t^1_3$ & $t^1_2$  & $t^1_4$  & $t^1_1$ & $t^1_3$ & $t^1_2$ & $t^1_5$ & $t^1_5$ & $t^1_5$ & $t^1_5$ & $t^1_5$ \\ \hline
\end{tabular}
 \caption{Transition coupling constraints for the example hierarchical transition model. Each column specifies a level 2 transition and a level 1 transition that can co-occur.}
 \label{table:trans_coupling}
 \end{center}
 \end{table}

The level 1 transition model, level 2 transition model, and coupling constraints from the state transition diagram directly map into the corresponding parameters $W^1$, $W^2$, and $U$ as follows. The $W^1$ and $W^2$ matrices are constructed following the procedure outlined in Section~\ref{sec:fact_state_tran_model}. Each $w^1_i$ is the vertical concatenation of the two state basis vectors associated transition $t^1_i$. $W^1$ is then constructed as the horizontal concatenation of the transition basis vector columns \{$w^1_i$\}. We then have:

\begin{align}
W^1 =& \left[ \begin{array}{ccccc} w^1_1 & w^1_2 & w^1_3 & w^1_4 & w^1_5\end{array} \right] \notag \\
=&\ \left[ \begin{array}{c}
W^1_1 \\
W^1_2 \end{array} \right] \notag\\
=&\ \left[ \begin{array}{ccccc}
s^1_1 & s^1_2 & s^1_2 & s^1_3 & s^1_3 \\
s^1_2 & s^1_3 & s^1_1 & s^1_1 & s^1_3 \end{array} \right] \notag \\
=& \left[ \begin{array}{cccccc}
1 & 0 & 0 & 0 & 0 \\
0 & 1 & 1 & 0 & 0 \\
0 & 0 & 0 & 1 & 1 \\ \hline
0 & 0 & 1 & 1 & 0 \\
1 & 0 & 0 & 0 & 0 \\
0 & 1 & 0 & 0 & 1 \end{array} \right] 
\label{eqn:W_1hfsm}
\end{align}

Note that $W^1_1$ is given by:

\begin{align}
W^1_1 =& \left[ \begin{array}{ccccc} s^1_1 & s^1_2 & s^1_2 & s^1_3 & s^1_3 \end{array} \right] \notag \\
=& \left[ \begin{array}{cccccc}
1 & 0 & 0 & 0 & 0 \\
0 & 1 & 1 & 0 & 0 \\
0 & 0 & 0 & 1 & 1\end{array} \right] 
\end{align}

and $W^1_2$ is given by:

\begin{align}
W^1_2 =& \left[ \begin{array}{ccccc} s^1_2 & s^1_3 & s^1_1 & s^1_1 & s^1_3 \end{array} \right] \notag \\
=& \left[ \begin{array}{cccccc}
0 & 0 & 1 & 1 & 0 \\
1 & 0 & 0 & 0 & 0 \\
0 & 1 & 0 & 0 & 1\end{array} \right] 
\end{align}

We construct $W^2$ as:

\begin{align}
W^2 =& \left[ \begin{array}{ccccccccccc} w^2_1 & w^2_2 & w^2_3 & w^2_4 & w^2_5 & w^2_6 & w^2_7 & w^2_8 & w^2_9 & w^2_{10} & w^2_{11} \end{array} \right] \notag \\
=&\ \left[ \begin{array}{c}
W^2_1 \\
W^2_2 \end{array} \right] \notag\\
=&\ \left[ \begin{array}{ccccccccccc}
s^2_1 & s^2_1 & s^2_2 & s^2_2 & s^2_3 & s^2_4 & s^2_5 & s^2_6 & s^2_5 & s^2_3 & s^2_6 \\
s^2_1 & s^2_2 & s^2_3 & s^2_4 & s^2_4 & s^2_5 & s^2_6 & s^2_1 & s^2_5 & s^2_3 & s^2_6 \end{array} \right] \notag \\
=& \left[ \begin{array}{ccccccccccc}
1 & 1 & 0 & 0 & 0 & 0 & 0 & 0 & 0 & 0 & 0 \\
0 & 0 & 1 & 1 & 0 & 0 & 0 & 0 & 0 & 0 & 0 \\
0 & 0 & 0 & 0 & 1 & 0 & 0 & 0 & 0 & 1 & 0 \\
0 & 0 & 0 & 0 & 0 & 1 & 0 & 0 & 0 & 0 & 0 \\
0 & 0 & 0 & 0 & 0 & 0 & 1 & 0 & 1 & 0 & 0 \\
0 & 0 & 0 & 0 & 0 & 0 & 0 & 1 & 0 & 0 & 1 \\ \hline
1 & 0 & 0 & 0 & 0 & 0 & 0 & 1 & 0 & 0 & 0 \\
0 & 1 & 0 & 0 & 0 & 0 & 0 & 0 & 0 & 0 & 0 \\
0 & 0 & 1 & 0 & 0 & 0 & 0 & 0 & 0 & 1 & 0 \\
0 & 0 & 0 & 1 & 1 & 0 & 0 & 0 & 0 & 0 & 0 \\
0 & 0 & 0 & 0 & 0 & 1 & 0 & 0 & 1 & 0 & 0 \\
0 & 0 & 0 & 0 & 0 & 0 & 1 & 0 & 0 & 0 & 1 \end{array} \right] 
\label{eqn:W_2hfsm}
\end{align}
 
Likewise, $W^2_1$ corresponds to the upper sub-matrix of $W^2$ and $W^2_2$ corresponds to the lower sub-matrix.

We construct the transition coupling matrix $U$ as follows. We start with the $Q$ pairs of level 2 - level 1 couplings specified in Table~\ref{table:trans_coupling}. We construct $U$ from this table such that the $i$'th column of $U$ corresponds to the transition pair in the $i$'th column of the table. For the level 2 transitions, we replace each transition $t^2_j$ in the table with the column vector $u_{upper_j} \in \mathbb{R}^{R2}$ that contains a 1 in the component corresponding to the row of $H^2$ that activates the corresponding transition in $W^2$. Likewise, for the level 1 transitions, we replace each transition $t^1_j$ in the table with the column vector $u_{lower_j} \in \mathbb{R}^{R1}$ that contains a 1 in the component corresponding to the row of $H^1$ that activates the corresponding transition in $W^1$. We then form the $j$'th column $u_j$ of $U$ as the vertical concatenation of $u_{upper_j}$ on top of $u_{lower_j}$ so that we have for the $(R2 + R1)$ x $Q$ matrix U:

\begin{align}
U =& \left[ \begin{array}{ccccc} u_1 & u_2 & u_3 & \dots & u_Q \end{array} \right] \notag \\
=& \left[ \begin{array}{c} U_1 \\
U_2 \end{array} \right] \notag \\
=& \left[ \begin{array}{ccccc} u^2_{upper_1} & u^2_{upper_2} &u^2_{upper_3} & \dots & u^2_{upper_Q} \\
u^1_{lower_1} & u^1_{lower_2} & u^1_{lower_3} & \dots & u^1_{lower_Q} \end{array} \right]
\end{align}

where the $R2$ x $Q$ submatrix $U_1$ is given by:

\begin{align}
U_1 = \left[ \begin{array}{ccccc} u_{upper_1} & u_{upper_2} &u_{upper_3} & \dots & u_{upper_Q} \end{array} \right]
\end{align}

and the $R1$ x $Q$ submatrix $U_2$ is given by:

\begin{align}
U_2 = \left[ \begin{array}{ccccc} u_{lower_1} & u_{lower_2} & u_{lower_3} & \dots & u_{lower_Q} \end{array} \right]
\end{align}

From Table~\ref{table:trans_coupling}, we then have the following for $U$:

\begin{align}
\label{eq:U_coupling}
U = \left[ \begin{array}{ccccccccccccc}
0 &  0 &  0 &  0 &  0 &  0 &  0 &  0 &  0 &  0 &  1 &  0 &  0 \\
 0 &  0 & 0 &  0 &  0 &  0 &  0 &  0 &  0 &  0 &  0 &  1  &  0 \\
 1 &  0 &  0 &  0 &  0 &  0 &  0 &  0 &  0 &  0 &  0 &  0 &  0 \\
 0 &  0 &  0 &  0 &  0 &  0 &  0 &  0 &  0 &  0 &  0 &  0 &  1 \\
 0 &  0 &  0 &  1 &  0 &  0 &  0 &  0 &  0 &  0 &  0 &  0 &  0 \\
 0 &  0 &  0 &  0 &  1 &  0 &  0 &  0 &  0 &  0 &  0 &  0 &  0 \\
 0 &  0 &  0 &  0 &  0 &  0 &  0 &  1 &  0 &  0 &  0 &  0 &  0 \\
 0 &  0 &  0 &  0 &  0 &  0 &  0 &  0 &  0 &  1 &  0 &  0 &  0 \\
 0 &  0 &  0 &  0 &  0 &  1 &  1 &  0 &  0 &  0 &  0 &  0 &  0 \\
 0 &  1 &  1 &  0 &  0 &  0 &  0 &  0 &  0 &  0 &  0 &  0 &  0 \\
 0 &  0 &  0 &  0 &  0 &  0 &  0 &  0 &  1 &  0 &  0 &  0 &  0 \\ \hline
 0 &  1 &  0 &  0 &  0 &  1 &  0 &  0 &  0 &  0 &  0 &  0 &  0 \\
 0 &  0 &  0 &  1 &  0 &  0 &  0 &  1 &  0 &  0 &  0 &  0 &  0 \\
 0 &  0 &  1 &  0 &  0 &  0 &  1 &  0 &  0 &  0 &  0 &  0 &  0 \\
 1 &  0 &  0 &  0 &  1 &  0 &  0 &  0 &  0 &  0 &  0 &  0 &  0 \\
 0 &  0 &  0 &  0 &  0 &  0 &  0 &  0 &  1 &  1 &  1 &  1 &  1 \end{array} \right] 
\end{align}

Note that $U^1$ activates $H^2$ (activations for $W^2$) and $U^2$ activates $H^1$ (activations for $W^1$)

\subsection{Empirical results}

In this section, we perform inference on the network in Figure~\ref{fig:2layerDynamic} using the parameter matrices for the regular expression hierarchical transition diagram in Figure~\ref{fig:hfsm1_detailed_crop}. We use parameter matrices  $W^1$, $W^2$, and $U$ from Equations (\ref{eqn:W_1hfsm}),(\ref{eqn:W_2hfsm}), and (\ref{eq:U_coupling}) respectively.

We choose to model a length 10 sequence for ease of diplay of the results. We let the observed set $X_E$ consist of $\{x^2_2 = s^2_2, x^2_7 = s^2_4\}$. We let the hidden set $X_H$ consist of all other model variables. That is, we are only given that in time slice 2, the level 2 state is $S^2_2$ (production state ``b'') in Figure~\ref{fig:hfsm1_detailed_crop} and that in time slice 7, the level 2 state is $S^2_4$ (production state ``c''). We are given no information about the level 1 states, transitions, or coupling variable states. However, even given this limited information, the constraints impose by the hierarchical transition model result in a single possible solution for the hidden variables in time slices $t=1,\dots,9$. Multiple solutions (due to multiple outgoing transitions) are possible for time slice 10. The reader can verify this by plugging in the observed states and studying the transition diagram to trace the patch of allowable transitions.

We now perform inference using an instance of the general inference and learning algorithm given in Appendix~\ref{sec:inf_learn_dyn2level}. All hidden variables are initialized to random positive values, and the algorithm is run for 500 iterations, which is sufficient for convergence to an exact factorization (RMSE less than $10^-4$) for all three factorization equations. We observed that repeated runs always converged, although a different solution for time slice 10 was reached in each case (since multiple solutions are possible for this time slice).

\begin{figure}
\centering
\includegraphics[width=80ex]{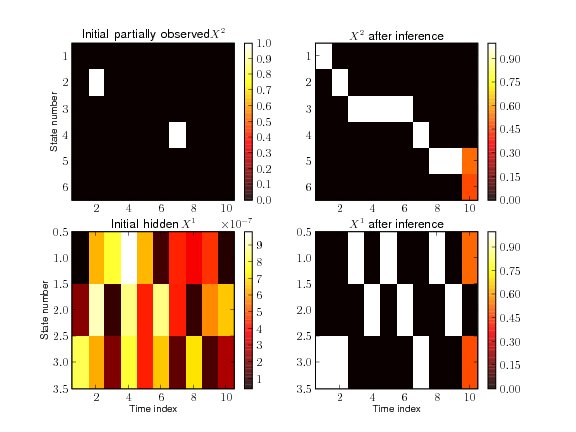}
\caption{The initialized $X^2$ is shown in the upper left. Model variables $x^2_2 = s^2_2$ and $x^2_7 = s^2_4$ were observed and all other model variables were hidden and estimated by the inference algorithm. The initialized $X^1$ is shown in the lower left. All hidden variables were initialized to random positive values. The inference results for $X^1$ and $X^2$ are shown in the rightmost image plots. We note that given the two observed variables, only one possible solution exists for time slices $t=1,\dots,9$. Multiple solutions exist for the final time slice $t=10$.}
\label{fig:hdyn1_X1_X2_results}
\end{figure}

Figure~\ref{fig:hdyn1_X1_X2_results} shows both the initialized and inferred values for $X^1$ and $X^2$.  Model variables $x^2_2 = s^2_2$ and $x^2_7 = s^2_4$ were observed and all other model variables were hidden and estimated by the inference algorithm. Note that the hidden values of $X^2$ (all time slices except $t=2,7$) appear black in the figure because the initialized random values are very small (order of $10^-6$) compared to the maximum value of the observed time slices in the figure. The inference results for $X^1$ and $X^2$ are shown in the rightmost image plots. We see that the inference algorithm successfully found the single possible solution for time slices $t=1,\dots,9$. In time slice 10 there are two possible solution states for both $x^1_{10}$ and $x^2_{10}$ and we see that the inferred values represent a mixture of the two possible state configurations. That is, in time slice 9, the inferred state is $x^2_9 = s^2_5$ and $x^1_9 = s^1_2$. According to the transition diagram in Figure~\ref{fig:hfsm1_detailed_crop}, we see that in the next time slice (slice 10), we can either remain in level 2 state $S^2_5$ while level 1 transitions to $S^1_1$, or both level 2 and level 1 can transition to their respective end states: $S^2_6$ and $S^1_3$. We observe that the inferred state values represent a superposition of these two possibilities. We also note that when sparse NMF updates were used in inference, the solution tended to chose one alternative or the other, but not a superposition of the two.

\begin{figure}
\centering
\includegraphics[width=80ex]{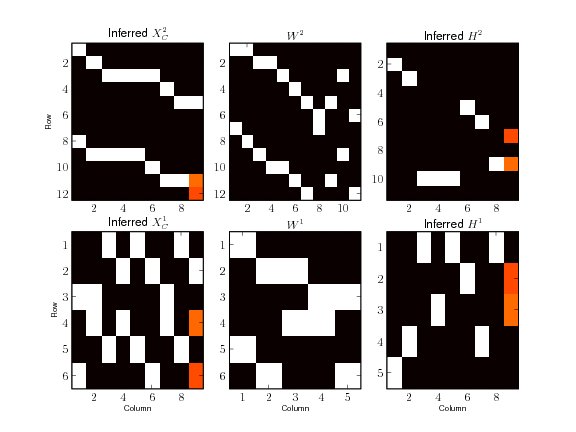}
\caption{The top three matrices show the inference results for $X^2_C$ and $H^2$ in the factorization $X^2_C = W^2 H^2$. The bottom three matrices show the inference results for $X^1_C$ and $H^1$ in the factorization $X^1_C = W^2 H^1$. Model variables $x^2_2$ and $x^2_7$ were observed and all other model variables were hidden and inferred by the inference algorithm. Parameter matrices $W^1$ and $W^2$ are given in Equations (\ref{eqn:W_1hfsm}) and (\ref{eqn:W_2hfsm}), respectively.}
\label{fig:hdyn1_X_W_H_factor}
\end{figure}

\begin{figure}
\centering
\includegraphics[width=80ex]{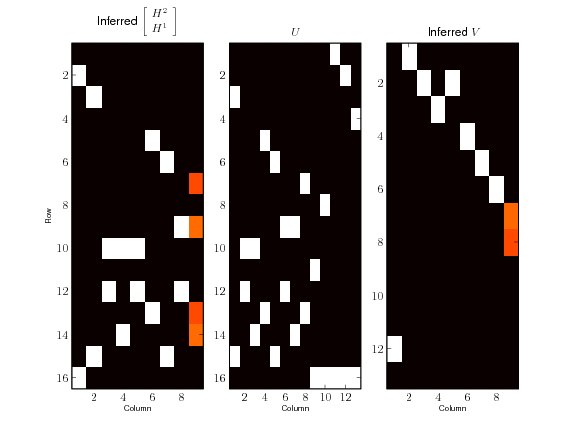}
\caption{An image plot of the inference results for the matrices in the transition coupling factorization. All variables $H^1, H^2, V$ were hidden and inferred. Parameter matrix $U$ is given in Equation (\ref{eq:U_coupling}).}
\label{fig:hdyn1_H_U_V_factor}
\end{figure}

Figure~\ref{fig:hdyn1_X_W_H_factor} shows the inference results for $X^2_C$ and $H^2$ in the factorization $X^2_C = W^2 H^2$ in the top three image plots. The bottom three image plots show the inference results for $X^1_C$ and $H^1$ in the factorization $X^1_C = W^2 H^1$.  Figure~\ref{fig:hdyn1_H_U_V_factor} shows an image plot of the inference results for the matrices in the transition coupling factorization.

\section{Multiple target tracking model}
\label{sec:target_tracking}

We now consider a multiple target tracking model that is motivated by the problem of tracking auditory objects in computational auditory scene analysis (CASA) \cite{ref:bregman}. This example serves to motivate the potential usefulness of our approach to problems such as CASA. Recall that a DPFN supports the representation of an additive combination of realizations of underlying dynamical process models. Such a representation seems to fit very naturally within CASA problems, since the sound pressure waves that arrive at the eardrum are well modeled as an additive mixture of scaled versions of the sound pressure waves produced by each auditory object alone. If it turns out to be possible to use a DPFN to model auditory objects such as musical instruments, environmental noises, speech, etc., then the linearity property would imply that any additive mixture of the various supported auditory objects at various sound levels would then be representable by the model as well. Of course, this does not imply that any particular inference and/or learning algorithm will successfully be able to deal with such a mixture, but the fact that an additive mixture is representable at all seems to be quite powerful. Our approach has the desirable property that nowhere in our model do we need to explicitly specify anything about the maximum number of simultaneous targets/objects or the allowable magnitude range (volume levels) for targets.

We will specify a dynamical model for a single target, and then observe how a scaled mixture of multiple targets is then automatically supported. This illustrates how the factored representation of a dynamical system allows for a compact parameter set, while still being expressive enough to support multiple simultaneous scaled realizations of the underlying process model. All parameters in the target model will be manually specified here for ease of illustration, although in principle one could attempt to learn them from training data as well. We therefore perform inference only in the results that follow. We will also work with synthetic target observation data, rather than actual audio recordings, again for ease of explanation and illustration. It is hoped that after reading this paper, the interested reader will find it straightforward to come up with ideas for developing more complex extensions of these models that may then be applied to practical problems involving real-world data sets.

The problem setup is as follows. We are given a sequence of non-negative observation column vectors, one for each time slice. When concatenated horizontally, they form an image such that the time axes runs from left to right. The vertical axis then corresponds to sensor position or frequency bin index, depending on the interpretation. We therefore use position and frequency bin interchangeably in the following discussion. In the context of CASA, each observation vector can be considered a time slice of a magnitude spectrogram such that the components of an observation vector represent the short-time spectra of an audio signal.  In the CASA interpretation, we can then think of the observation image as a magnitude time frequency image, such as a spectrogram.

The observations may also be potentially corrupted by additive non-negative noise. Our goal is to infer the types and positions of all targets given only the target observation sequence. We may also be interested in predicting future target trajectories and/or inferring missing observations data.

Specifically, we wish to model a target that has the following properties:
\begin{enumerate}

\item A new target can come come into existence (become alive) at any time instant. 

\item Each target has a certain localized \emph{pattern signature}. That is, its position/frequency evolves in time according to an underlying transition model. Target classes are distinguished by their position signatures. For example, one can think of two distinct musical instruments: one produces a certain type of vibrato regardless of the note pitch, while the other instrument produces a short chirp followed by a more steady pitch.

\item A target's pattern signature is modeled as being independent of the target's overall position/location/frequency bin index. Thus, we model a target's location as a position shift or frequency shift of the pattern signature. We only allow a target's overall position to change at certain constrained points in the position signature transition model. In the context of CASA, a the target's overall location might correspond to the pitch of a note played by an instrument, which is independent of the instrument's pattern signature (e.g., each note is performed with a vibrato).

\item A target may cease to exist (die) after some number of time slices which may or may not be deterministic. 

\item When a target comes into existence, it can take on any positive magnitude value, which remains constant during the time that the target is alive. This corresponds to an auditory object, such as an instrument note event, that can begin sounding at an arbitrary sound level or loudness and remains at the same sound level until the note ends. It is possible to relax this constraint, but we enforce it in this example.

\item Any number of targets may be simultaneously present, each potentially having a distinct magnitude. This corresponds to multiple auditory objects simultaneously occurring, such as several instruments performing together, or several people talking at the same time, for example.

\end{enumerate}

\subsection{Model}

\begin{figure}
\centering
\includegraphics[width=60ex]{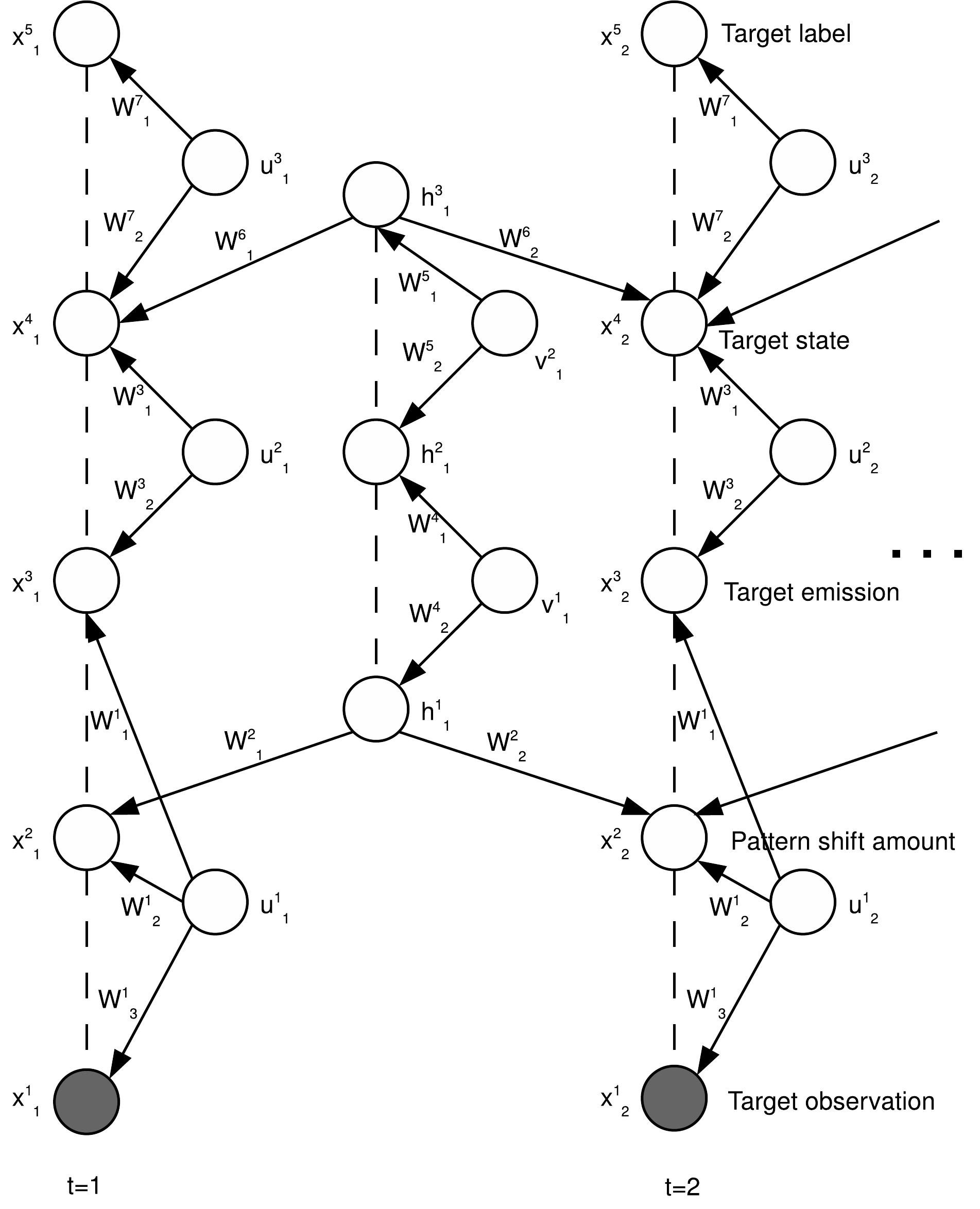}
\caption{The DPFN for the the multiple target tracker. The target observations variables $x^1_t, t=1...T$ are observed and all other variables are hidden. The first two time slices are shown.}
\label{fig:targetTracking1}
\end{figure}

Figure~\ref{fig:targetTracking1} shows the DPFM for the target tracker. $x^1_t, t=1...T$ are observed and all other variables are hidden. We now describe the role of each of the 7 factorization equations that specify this model. We make use of the matrix notation developed in Section \ref{sec:fact_state_tran_model} so that the factorizations may be expressed concisely.

\subsubsection{Target state transition model}

The target state transition model corresponds to the factorization:
\begin{align}
\label{eq:target_state_transition_model}
X^4_c =& W^6 H^3 \notag \\
=& \left[ \begin{array}{c}
W^6_1 \\
W^6_2 \end{array} \right] H^3
\end{align}

\begin{figure}
\centering
\includegraphics[width=80ex]{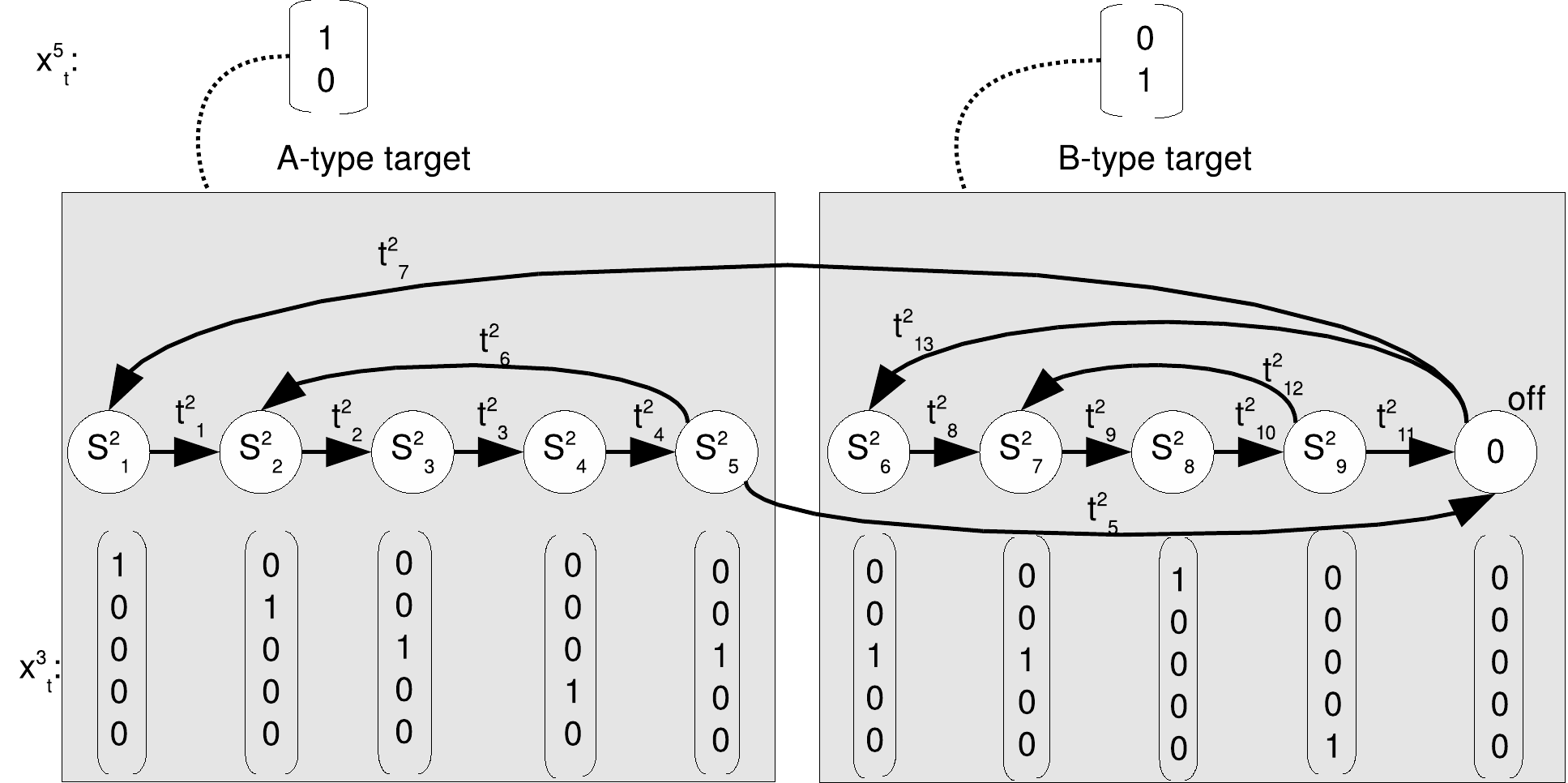}
\caption{The state transition diagram for the target state variables $x^4_t$. The corresponding state emission vectors $x^3_t$ are also shown.}
\label{fig:fsm_target_tracker1}
\end{figure}

Figure~\ref{fig:fsm_target_tracker1} shows the target state transition diagram. Observe that states $S^2_1$ through $S^2_5$ correspond to a ``A-type'' target and states $S^2_6$ through $S^2_9$ correspond to a ``B-type'' target. The initial state when a ``A-type'' target comes into existence is $S^2_1$, and the minimum duration (i.e., target lifetime) is 5 time slices, since the target must pass consecutively through states $S^2_1$ through $S^2_5$ before it is possible for the target to die (i.e., transition to the 0/off state). We see that when the current state is $S^2_5$ the following state can be $S^2_2$ or the 0 state. Recall that this is not a probabilistic model. Thus, multiple outgoing transitions from a given state can be thought of as transition possibilities that are equally likely. The particular transition (or superposition of transitions) that is chosen during inference depends on the values of the observed variables as well as any other constraints imposed by other parts of the model that are coupled to the transition model. The state transition model for a ``B-type'' target is similar but only consists of 4 states. We construct $W^6$ directly by inspection of the state transition diagram using the procedure described in Section \ref{sec:fact_state_tran_model}. Note that there are 9 states and 13 transitions. The ``off'' state is not an explicit state, and is represented by the 9-dimensional 0 vector in the transition basis matrix. Thus $W^6$ has size 18 x 13. The state basis vectors \{$s^2_i$\} are therefore 9-dimensional column vectors, as are the target state variables \{$x^4_t$\}. Each $w^6_i$ is the vertical concatenation of the two basis state vectors associated transition $t^2_i$. $W^6$ is then constructed as the horizontal concatenation of the transition basis vector columns \{$w^1_i$\}. We then have for $W^6$:

\begin{align}
W^6 =& \left[ \begin{array}{ccccccccccccc} w^6_1 & w^6_2 & w^6_3 & w^6_4 & w^6_5 & w^6_6 & w^6_7 & w^6_8  & w^6_9 & w^6_{10} & w^6_{11} & w^6_{12} & w^6_{13} \end{array} \right] \notag \\
=&\ \left[ \begin{array}{c}
W^6_1 \\
W^6_2 \end{array} \right] \notag\\
=&\ \left[ \begin{array}{ccccccccccccc}
s^2_1 & s^2_2 & s^2_3 & s^2_4 & s^2_5  & s^2_5 & 0      & s^2_6 & s^2_7 & s^2_8 & s^2_9 & s^2_9 & 0\\
s^2_2 & s^2_3 & s^2_4 & s^2_5 & 0      & s^2_2 & s^2_1  & s^2_7 & s^2_8 & s^2_9 & 0     & s^2_7 & s^2_6\end{array} \right] \notag \\
\label{eqn:W_6target}
\end{align}

\subsubsection{Target label to target state coupling}

The target label to target state coupling corresponds to the factorization:

\begin{align}
\label{eq:target_label_to_target_state_coupling}
\left[ \begin{array}{c}
X^5 \\
X^4 \end{array} \right] =& W^7 U^3 \notag \\
=& \left[ \begin{array}{c}
W^7_1 \\
W^7_2 \end{array} \right] U^3
\end{align}

This module performs the task of assigning a target type label $x^5_t$ to a target state $x^4_t$. From Figure~\ref{fig:fsm_target_tracker1} we see that there are two distinct target types. We let $x^5_t$ denote the 2-dimensional column vector such that the first (top) component denotes the strength of an ``A-type'' target and the second (bottom) component denotes the strength of a ``B-type'' target. We will choose the basis columns $U$ such that 

We wish to add coupling constraints so that the ``A-type'' basis vector $s^3_1 = (1,0)^T$ can co-occur with any of the basis state vectors \{$s^2_1, s^2_2, s^2_3, s^2_4, s^2_5$\} and the ``B-type'' basis vector $s^3_2 = (0,1)^T$ can co-occur with any of the basis state vectors \{$s^2_6, s^2_7, s^2_8, s^2_9$\}. We construct each column of $W^7$ as the vertical concatenation of an $s^3_i$ on top of and $S^2_j$ that can co-occur. Noting that the column ordering is arbitrary, we have for the 7 x 9 matrix $W^7$:

\begin{align}
W^7 =& \left[ \begin{array}{c} W^7_1 \\
W^7_2 \end{array} \right] \notag \\
=& \left[ \begin{array}{ccccccccc} s^3_1 & s^3_1 & s^3_1 & s^3_1 & s^3_1 & s^3_2 & s^3_2 & s^3_2 & s^3_2 \\
s^2_1 & s^2_2 & s^2_3 & s^2_4 & s^2_5 & s^2_6 & s^2_7 & s^2_8 & s^2_9 \end{array} \right]
\end{align}

\subsubsection{Target state to target emission coupling}

The target state to target emission coupling corresponds to the factorization:

\begin{align}
\label{eq:target_state_to_target_emission_coupling}
\left[ \begin{array}{c}
X^4 \\
X^3 \end{array} \right] =& W^3 U^2 \notag \\
=& \left[ \begin{array}{c}
W^3_1 \\
W^3_2 \end{array} \right] U^2
\end{align}

This module performs the task of coupling the target state variable $x^4_t$ to the corresponding production (emission) variable $x^3_t \in \mathbb{R}^{patternSize}$, where $patternSize = 5$ in this example. Figure~\ref{fig:fsm_target_tracker1} shows the emission basis vector $s^e_i \in \mathbb{R}^{patternSize}$ below the corresponding state $S^2_i$. If a given state $S^2_i$ is present in $x^4_t$ with magnitude $\alpha$, then we wish for the corresponding emission variable $s^e_i$ to also be present in $x^3_t$ with magnitude $\alpha$ and vice versa. We construct each column of $W^3$ as the vertical concatenation of a basis state vector $s^2_i$ on top of the corresponding emission basis vector $s^e_i$. The 10 x 9 matrix $W^3$ is then given as:

\begin{align}
W^3 =& \left[ \begin{array}{c} W^3_1 \\
W^3_2 \end{array} \right] \notag \\
=& \left[ \begin{array}{ccccccccc} s^2_1 & s^2_2 & s^2_3 & s^2_4 & s^2_5 & s^2_6 & s^2_7 & s^2_8 & s^2_9 \\
s^e_1 & s^e_2 & s^e_3 & s^e_4 & s^e_5 & s^e_6 & s^e_7 & s^e_8 & s^e_9 \end{array} \right]
\end{align}

\subsubsection{Pattern translation transition model}

The pattern translation transition model corresponds to the factorization:

\begin{align}
\label{eq:pattern_translation_transition_model}
X^2_c =& W^2 H^1 \notag \\
=& \left[ \begin{array}{c}
W^2_1 \\
W^2_2 \end{array} \right] H^1
\end{align}

\begin{figure}
\centering
\includegraphics[width=80ex]{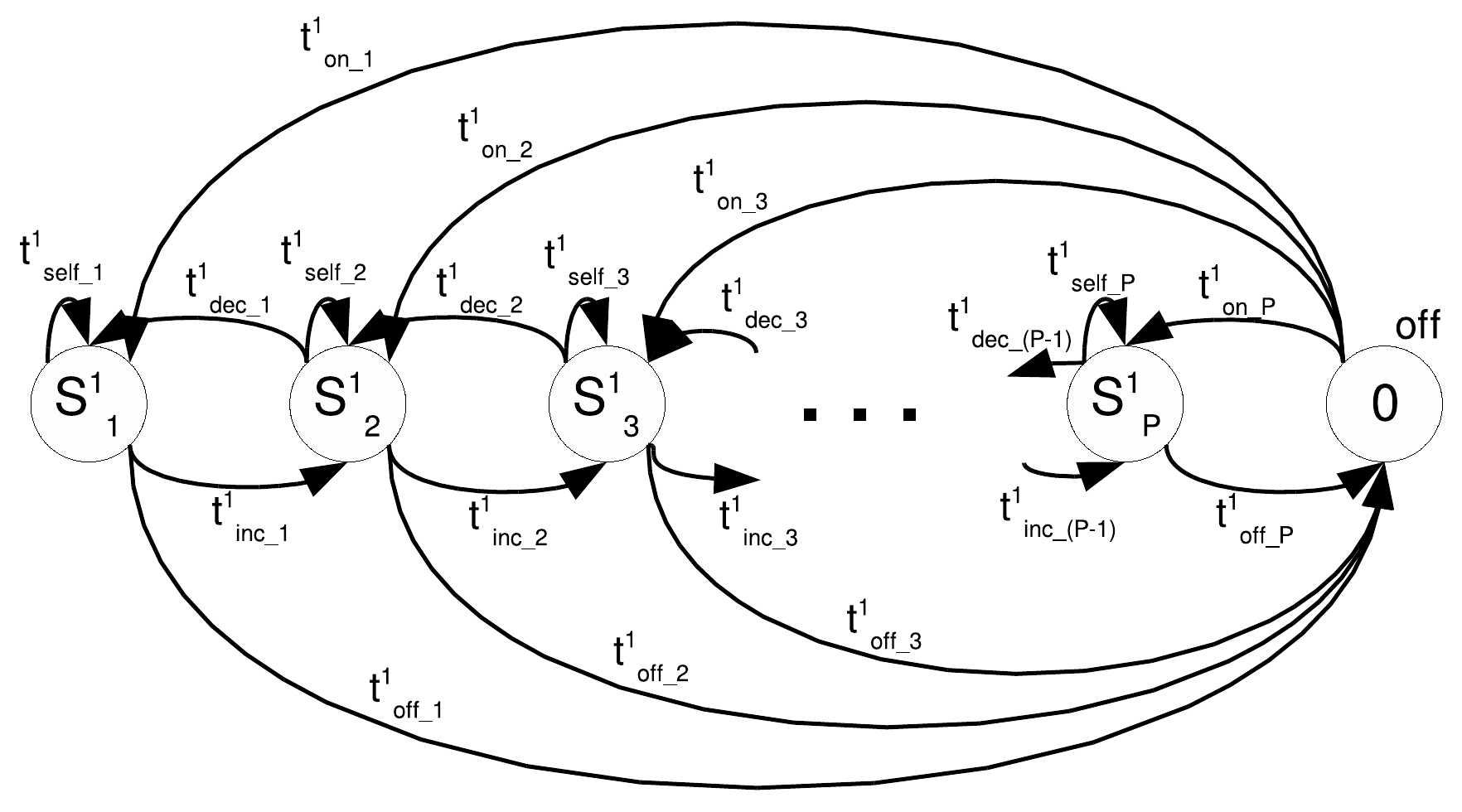}
\caption{The state transition diagram for the position shift variables $x^2_t$ with $P$ possible distinct position shift amounts.}
\label{fig:targetTrackingPosition}
\end{figure}

This module represents the transition model for the amount by the target emission pattern is translated (i.e., target position) between time steps $t$ and $t+1$. Figure~\ref{fig:targetTrackingPosition} shows the state transition diagram for the position shift variables $x^2_t$ with $P$ possible distinct position shift amounts. Note that $x^2_t$ is therefore a $P$ dimensional column vector, with the $i$'th component corresponding to the magnitude of $S^1_i$ at time $t$. Each state $S^1_i$ corresponds to a distinct target position. The index $i$ representing the downward translation amount that the target pattern will be translated when it appears in the observation vector $x^1_t$. The state $S^1_1$ corresponds to the minimum downward translation amount, so that the target emission pattern appears as the top $patternSize$ components of $x^1_t$. The state $S^1_P$ corresponds to the maximum downward translation amount, so that the target emission pattern appears as the bottom $patternSize$ components of $x^1_t$.

We specify in the state transition diagram that a transition $t_{on_i}$ from the off state at time $t$ to to any of the $P$ possible position shift amounts $S^1_i$ at time $t+1$ is allowed. Let $t_{on}$ denote the set \{$t_{on_i}, i = 1, \dots, P$\} of all ``on'' transitions. We specify that a transition $t_{off_i}$ from any of the $P$ possible positions $S^1_i$ to the off state is allowed. Let $t_{off}$ denote the set \{$t_{off_i}, i = 1, \dots, P$\} of all ``off'' transitions.  A self-loop transition $t_{self_i}$ from any position state $S^1_i$ to itself is allowed. Let $t_{self}$ denote the set \{$t_{self_i}, i = 1, \dots, P$\} of all ``self-loop'' transitions.  Finally, we also allow a transition $t_{inc_i}$ from state $S^1_i$ to $S^1_{i+1}$, as well as a transition $t_{dec_i}$ from state $S^1_{i+1}$ to $S^1_{i}$. Let $t_{inc}$ denote the set \{$t_{inc_i}, i = 1, \dots, P-1$\} of all ``position increment'' transitions. Let $t_{dec}$ denote the set \{$t_{dec_i}, i = 1, \dots, P-1$\} of all ``position decrement'' transitions. Let $t_{change}$ denote the union of $t_{inc}$ and $t_{dec}$, which is the set of all ``position change'' transitions. The complete set of all position transitions is then $Tpos =\{t_{self}, t_{change}, t_{on}, t_{off}\}$.

We construct $W^2$ directly by inspection of the state transition diagram using the procedure described in Section \ref{sec:fact_state_tran_model}. Using our convention for basis state vectors, state $S^1_i$ in the transition diagram corresponds to a 9-dimensional column vector $s^1_i$ such that the $i$'th component is 1 and all other components are 0.  The ``off'' state is not an explicit state, and is represented by the 9-dimensional 0 vector in the transition basis matrix. Since there are $P$ states, the state basis vectors \{$s^1_i$\} are 9-dimensional column vectors. Each column of the $2 P$ x $|Tpos|$ matrix $W^2$ then corresponds to the vertical concatenation of the two basis state vectors associated with each allowable transition in $Tpos$. Again, note that the column ordering of the transition basis vectors is arbitrary. For example, the column of $W^2$ corresponding to transition $t_{inc_2}$ would consist of the vertical concatenation of $s^1_2$ on top of $s^1_3$.

\subsubsection{Target state transition to position shift transition coupling}

We now place coupling constraints on the target state transition and pattern translation transition models. Without these constraints, it would be possible for a target to have a positive state magnitude but a zero-valued corresponding position magnitude and vice versa. However, we wish for these magnitudes to be identical for a given target. Specifically, we add coupling constraints so that a target state ``on'' transition co-occurs with a target position ``on'' transition, and likewise for the ``off'' transitions. We also constrain the target position value so that a pattern position change (pattern translation) can co-occur with the repetition transitions ($t^2_6, t^2_{13}$) for the target pattern, but is otherwise disallowed.

Although it would be possible to couple the target state transitions $h^3_t$ directly with the target position transitions $h^1_t$, each of the transitions $t^2_i$ would then have couplings for each of the $P$ positions, resulting in a potentially large number of coupling basis vectors. We can reduce the number of required coupled transitions by introducing an intermediate high-level position transition variable $h^2_t$. The target state transition variable $h^3_t$ is coupled to $h^2_t$ via the factorization:

\begin{align}
\label{eq:h3Toh2CouplingTarget}
\left[ \begin{array}{c}
H^3 \\
H^2 \end{array} \right] =& W^5 V^2 \notag \\
=& \left[ \begin{array}{c}
W^5_1 \\
W^5_2 \end{array} \right] V^2
\end{align}

and $h^2_t$ is coupled to the target position transition variable $h^1_t$ via the factorization:

\begin{align}
\label{eq:h2Toh1CouplingTarget}
\left[ \begin{array}{c}
H^2 \\
H^1 \end{array} \right] =& W^4 V^1 \notag \\
=& \left[ \begin{array}{c}
W^4_1 \\
W^4_2 \end{array} \right] V^1
\end{align}

Let $h^2_t$ correspond to a 4-dimensional column vector such that each component corresponds to a distinct type of position transition. Specifically, we let $h^2_t = a_1 = (1,0,0,0)^T$ represent ``remaining at the same position.'' We let   $h^2_t = a_2 = (0,1,0,0)^T$ represent ``position change.'' We let   $h^2_t = a_3 = (0,0,1,0)^T$ represent ``position turn on.''  We let   $h^2_t = a_4 = (0,0,0,1)^T$ represent ``position turn off.''

Table~\ref{table:trans_coupling_target_tran_abstract_pos} shows the transition coupling constraints between $h^3_t$ and $h^2_t$. We construct the parameter matrix $W^5$ from this table as follows. We see from \ref{eqn:W_6target} that setting $h^3_t$ equal to the standard unit basis vector $e_i \in \mathbb{R}^9$ will activate transition $t^2_i$. We then construct the $j$'th column of $W^5$ as the vertical concatenation of $e^j$ on top of $a_k$ such that $e^j$ and $a_k$ correspond to the $j$'th target state transition and high-level position transition in the table, respectively. We then arrive at the following 9 x 15 matrix for $W^5$:

\begin{align}
W^5 =& \left[ \begin{array}{ccccccccccccccc} e_6 & e_6 & e_{12} & e_{12} & e_{7} & e_{13} & e_5 & e_{11} & e_1 & e_2 & e_3 & e_4 & e_8 & e_9 & e_{10}\\
a_2 & a_1 & a_2 & a_1 & a_3 & a_3 & a_4 & a_4 & a_1 & a_1& a_1& a_1& a_1& a_1& a_1 \end{array} \right]
\end{align}

\begin{table}
\begin{center}
\begin{tabular}{| c ||c | c | c | c | c | c | c | c |}
    \hline
    Target state transition & $t^2_6$ & $t^2_6$ & $t^2_{12}$ & $t^2_{12}$ & $t^2_7$ & $t^2_{13}$ & $t^2_5$  & $t^2_{11}$ \\ \hline
    High-level position transition & $t_{change}$ & $t_{self}$ & $t_{change}$ & $t_{self}$  & $t_{on}$  & $t_{on}$ & $t_{off}$  & $t_{off}$ \\ \hline
\end{tabular}

\begin{tabular}{| c ||c | c | c | c | c | c | c | }
    \hline
    Target state transition & $t^2_1$ & $t_2$ & $t_3$ & $t_4$ & $t_8$ & $t_9$ & $t_{10}$ \\ \hline
    High-level position transition & $t_{self}$ & $t_{self}$ & $t_{self}$ & $t_{self}$ & $t_{self}$ & $t_{self}$ & $t_{self}$ \\ \hline
\end{tabular}
 \caption{Transition coupling constraints between the target state transitions $h^3_t$ and the high-level position transitions $h^2_t$. Each column specifies a target state transition and high-level position transition that can co-occur.}
 \label{table:trans_coupling_target_tran_abstract_pos}
 \end{center}
 \end{table}

We construct $W^4$ as follows. Note that the top 4 rows of $W^4$ correspond to the submatrix $W^4_1$, which contain the basis activations for $h^2_t$. The bottom submatrix $W^4_2$ of $W^4$ contains the activations for $h^1_t$. Note that setting $h^1_t$ equal to the standard unit basis vector $e_i \in \mathbb{R}^{|Tpos|}$ activates the $i$'th column of $W^2$, corresponding to a position transition in $Tpos =\{t_{self}, t_{change}, t_{on}, t_{off}\}$. For each $i \in 1 , \dots, |Tpos|$, we add a corresponding column $w^4_i$ to $W^4$ as follows. If the $h^1_t = e_i$ activates a transition in $W^2$ in the set $t_{self}$, let $w^4_i$ equal the vertical concatenation of $a_1$ on top of $e_i$. Otherwise, if $h^1_t = e_i$ activates a transition in $W^2$ in the set $t_{change}$, let $w^4_i$ equal the vertical concatenation of $a_2$ on top of $e_i$. Otherwise, if $h^1_t = e_i$ activates a transition in $W^2$ in the set $t_{on}$, let $w^4_i$ equal the vertical concatenation of $a_3$ on top of $e_i$. Otherwise, if $h^1_t = e_i$ activates a transition in $W^2$ in the set $t_{off}$, let $w^4_i$ equal the vertical concatenation of $a_4$ on top of $e_i$.

\subsubsection{Target pattern translation coupling}

The target pattern translation coupling corresponds to the factorization:

\begin{align}
\label{eq:target_pattern_translation_coupling}
\left[ \begin{array}{c}
X^3 \\
X^2 \\
X^1 \end{array} \right] =& W^1 U^1 \notag \\
=& \left[ \begin{array}{c}
W^1_1 \\
W^1_2 \\
W^1_3 \end{array} \right] U^1
\end{align}

where $W^1_1$ has $patternSize$ rows, $W^1_2$ has $P$ rows, and $W^1_3$ has $P -1 + patternSize$ rows.

We wish for the target emission vectors to appear translated (vertically shifted) in the observation vectors. An observation vector $x^1_t$ has dimension greater than the target emission vector $x^3_t$ so that $x^3_t$ may appear as a subvector in $x^1_t$. The number of components by which $x^3_t$ is translated is specified by the pattern shift amount $x^2_t$. If component $i$ of $x^3_t$ has value $\alpha$ and component $j$ of $x^2_t$ also has value $\alpha$ then we wish for component $k = i + j$ of $x^1_t$ to also have value $\alpha$. We then construct $W^1$ as follows. For each combination of $i \in 1, \dots, dimension(x^3_t)$ and $j \in 1, \dots, dimension(x^2_t)$ add a column $q_{i_j}$ to $W^1$. The subcolumn of $q_{i_j}$ corresponding to $W^1_1$ has only the $i$'th component equal to 1 and all other components 0-valued. The subcolumn of $q_{i_j}$ corresponding to $W^1_2$ has only the $j$'th component equal to 1 and all other components 0-valued. The subcolumn of $q_{i_j}$ corresponding to $W^1_3$ has only the $k$'th component equal to 1 and all other components 0-valued. 

\begin{figure}
\centering
\includegraphics[width=100ex]{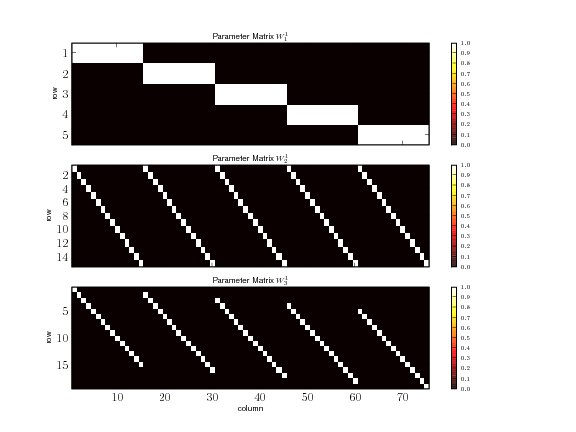}
\caption{Submatrices $W^1_1$, $W^1_2$, and $W^1_3$ of $W^1$.}
\label{fig:W1W2W3}
\end{figure}

Figure~\ref{fig:W1W2W3} shows an image plot of the submatrices of $W^1$ for the case where $patternSize = 5$, $dimension(x^2_t) = 15$, and $dimension(x^1_t) = P -1 + patternSize = 19$. Note that $W^1$ has $patternSize P = 75$ columns. Note that each column in $W^1_1, W^1_2, W^1_3$ has column sum equal to 1 so that $x^3_t$, $x^2_t$, and $x^1_t$ are modeled as having equal column sums. In our model, $x^1_t$ is observed and all other variables are hidden. Note that if a given component of $x^1_t$ has some positive value, then there are typically many basis columns in $W^1$ that can explain it, meaning that many solutions for $x^2_t, x^3_t$ are generally possible. However, the other factorizations in the model will add enough additional constraints that a unique solution for the hidden variables can still be possible.

A learning and inference algorithm for this network can be obtained by applying Algorithm~\ref{alg:inference_learning1} to the system of factorizations Equations (\ref{eq:target_state_transition_model}) - (\ref{eq:target_pattern_translation_coupling}). Appendix~\ref{sec:learn_inf_target_tracking} shows the psuedocode for the algorithm. Since all parameters matrices are known, we run the algorithm in inference-only mode with learning disabled in the following section. 

\subsection{Empirical Results}

We now present empirical results for the target tracking model. We make use of the parameter matrices $\{W^i\}$ presented in the previous section so that the model parameters can be considered known. We present inference-only results for the case where the observation sequence $X^1$ is either fully or partially observed.

\subsubsection{Clean target observations}
\label{sec:clean_target_obs}
\begin{figure}
\centering
\includegraphics[width=100ex]{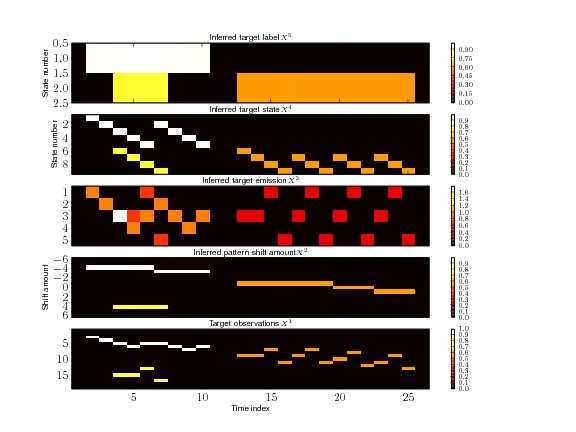}
\caption{Inference results for $X^2$, $X^3$, $X^4$, $X^5$. Note that only the bottom $X^1$ is observed and consist of clean target observations. All other variables are hidden and were correctly inferred.}
\label{fig:hdyn2_clean_X5X4X3X2X1}
\end{figure}

We first consider the case where $X^1$ is fully observed and consists of noiseless observations. All other model variables are hidden and will be inferred. The bottom image plot in Figure~\ref{fig:hdyn2_clean_X5X4X3X2X1} shows the target observations. There are three target objects represented: two A-type targets and one B-type target. Observe that there are two targets present during time slices $t = 4,5,6,7$. The targets have magnitudes of 1.0, 0.8, and 0.6. These target objects were chosen such that they conform to the target model and therefore are exactly representable under the model. We used $P = 15$ position values for these experiments, yielding a pattern shift amount from -7 to 7. That is, $P = 8$ corresponds to the 0 shift amount where a pattern is in the center of the observation vector. 


We used the inference algorithm described in Appendix~\ref{sec:learn_inf_target_tracking}, which is a special case of the general inference and learning algorithm presented in Algorithm~\ref{alg:inference_learning1}. The top four image plots in Figure~\ref{fig:hdyn2_clean_X5X4X3X2X1} show the inference results for $X^2$, $X^3$, $X^4$, $X^5$. We note that only $X^1$ was observed, so that higher-level variables $U^1$, $U^2$, $U^3$, $H^1$, $H^2$, $H^3$, $V^1$, and $V^2$ were also hidden, but are not shown here since they are less relevant.

We found that 10000 iterations were sufficient for all factorization equations to converge completely (RMSE less than 10e-4), which took approximately 1 minute to run on an a PC with a 3.0 GHz Intel Core2 Duo CPU. Given the observed sequence, exactly 1 solution is possible. All results for this model were obtained by running the inference algorithm for 10000 iterations. We observed that on repeated runs, the inference algorithm always converged to the correct exact solution (within our RMSE tolerance). We see that the observed sequence $X^1$ was successfully decomposed into the target label ($X^5$), target state ($X^4$), target emission pattern ($X^3$), pattern shift amount ($X^2$), and the other higher-level hidden variables which are not shown for space reasons.

\subsubsection{Noisy target observations}

\begin{figure}
\centering
\includegraphics[width=100ex]{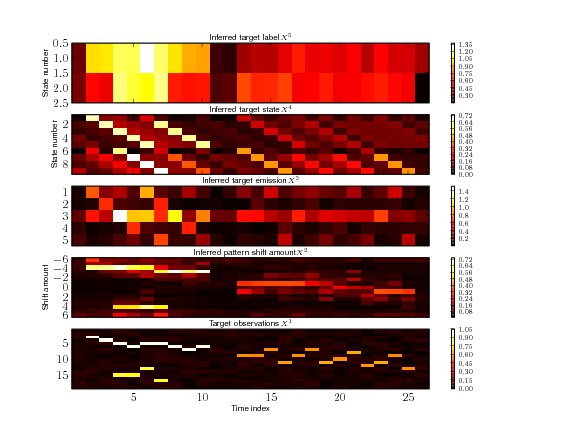}
\caption{Inference results for $X^2$, $X^3$, $X^4$, $X^5$. Note that only the bottom $X^1$ is observed and consist of target observations corrupted by additive noise.}
\label{fig:hdyn2_noisy_X5X4X3X2X1_0_sparseness}
\end{figure}

We now add some noise to the previous clean target observations to see how inference will perform. The bottom image plot in Figure~\ref{fig:hdyn2_noisy_X5X4X3X2X1_0_sparseness} shows the observed noisy target observations. Noise consisting of uniformly distributed values in the range $[0,0.06]$ was added to the clean target observations from the previous section to form the noisy target observations $X^1$. The maximum noise amount is thus one tenth of the smallest target magnitude.


Figure~\ref{fig:hdyn2_noisy_X5X4X3X2X1_0_sparseness} shows the inference results  for $X^2$, $X^3$, $X^4$, $X^5$. As before, the results for the other variables in the model are not shown for space reasons. We observe that the inferred values also appear somewhat noisy, since the model is also trying to explain the noise in the target observations. We observed that the visual cleanness of the results appears to degrade gradually as more noise is added. Small amounts of noise are tolerated quite well. A noise amount less than 0.02 or so produces visually clean results.

\subsubsection{Noisy target observations using sparse inference}

Given the previous results for the case of noisy observations, we might like for the inference algorithm to attempt to ignore as much of the noise as possible while still accounting for the non-noise components. We modify the inference algorithm so that the NMF update steps are replaced by corresponding sparse NMF update steps using the nsNMF algorithm \cite{nsNMF2006}, which is describe in Appendix~\ref{appendix_sparse_nmf}. We have observed empirically that the quality of the inference results seems to be improved if the sparseness parameter value is gradually increased during the inference procedure. For these results, the sparseness value was increased linearly from 0 at iteration 5000 to 0.05 at iteration 10000.

\begin{figure}
\centering
\includegraphics[width=100ex]{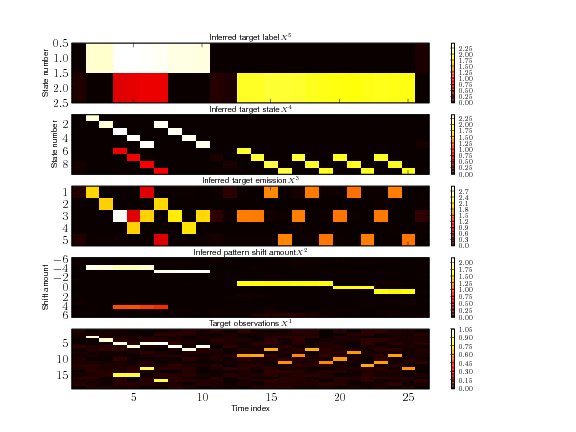}
\caption{Inference results for $X^2$, $X^3$, $X^4$, $X^5$. Note that only the bottom $X^1$ is observed and consist of target observations corrupted by additive noise. Gradually increasing sparseness was used in the inference algorithm.}
\label{fig:hdyn2_noisy_X5X4X3X2X1_gradual_sparseness}
\end{figure}

The input observations are generated with the same targets plus additive noise as in the previous section. Figure~\ref{fig:hdyn2_noisy_X5X4X3X2X1_gradual_sparseness} shows the inference results, where the bottom subplot shows the noisy target observations for reference. We observe that the inference results are visually less cluttered with noise compared to the previous non-sparse inference case. However, we observe that the magnitudes of the inference results are off by approximately a factor of 2, so that some renormalization would be needed as a post-processing step.

\subsubsection{Noisy target observations using modified sparse inference}

As an experiment, we now modify the previous sparse inference procedure so that the observation sequence $X^1$ is made to be hidden (i.e., all model variables are now hidden) during the final several inference iterations. The motivation for this is that after basically reaching convergence on the noisy observations data, the model has settled on a somewhat sparse solution. However, the noise in $X^1$ cannot be ignored completely, and so the model tries to find a configuration of the hidden variables that best explain the noise to some extent. However, if we stop updating $X^1$ on each iteration with the actual observations, the model can then be free to ignore the noise and converge to a sparser solution. We need to be careful, though. If the model is allowed to run a large number of iteration with $X^1$ now hidden, it may gradually ``forget'' $X^1$ and settle on some other configuration of variables.

We again supply the same target observations corrupted by additive noise as the previous examples. Sparseness was 0 for iterations 0 through 5000, and then gradually increased linearly reaching a maximum sparseness of 0.05 at iteration 9980. For the final 20 iterations, $X^1$ was made hidden. This was accomplished by simply disabling the re-copying of the observations data into the corresponding $X^1$ local variable during inference. Thus, on each iteration, the down propagated values for $X^1$ would be reused on the successive up propagation. 

\begin{figure}
\centering
\includegraphics[width=100ex]{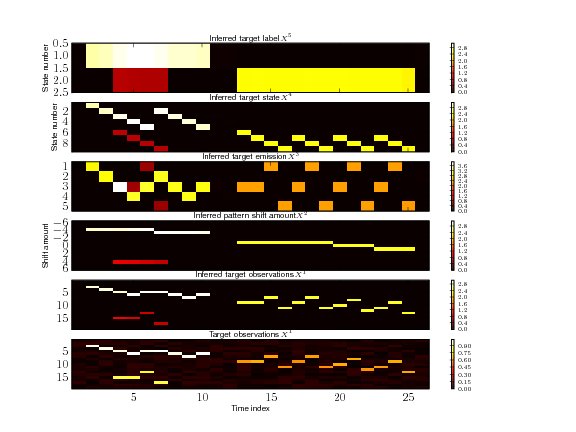}
\caption{Inference results for $X^1$, $X^2$, $X^3$, $X^4$, $X^5$. Gradually increasing sparseness was used. Here $X^1$ was observed for the first 9980 iterations and then made to be hidden for the last 20 iterations. Note that since $X^1$ was hidden for several iteration, the inferred values for $X^1$ here differ from those of the actual observed $X^1$ in the bottom image plot.}
\label{fig:hdyn2_noisy_X5X4X3X2X1}
\end{figure}

Figure~\ref{fig:hdyn2_noisy_X5X4X3X2X1} shows the inference results for $X^1$, $X^2$, $X^3$, $X^4$, $X^5$. Note that the inference results appear quite clean. However, the relative magnitudes of two of the targets are not correct. Note that the lowest-magnitude target in the figure corresponds to an to an inferred target with with magnitude greater than one of the other inferred targets.

\subsubsection{Prediction}

We now consider the case where some of the target observations are hidden. We will then infer their values are part of the inference procedure, along with the other hidden variables. We first consider the case where time slice 25 of the clean $X^1$ from Section~\ref{sec:clean_target_obs} is hidden. The regular inference algorithm was used (i.e.,the usual non-sparse NMF updates were used).

\begin{figure}
\centering
\includegraphics[width=100ex]{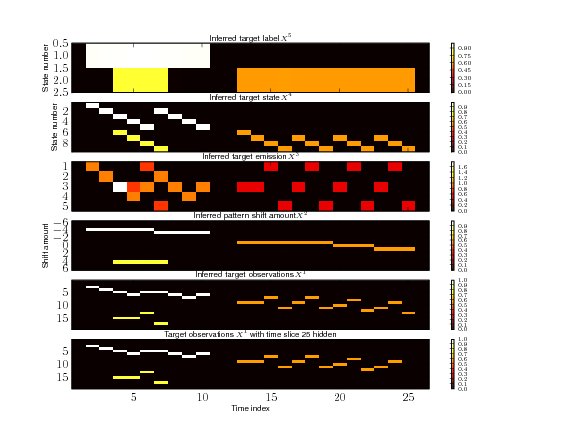}
\caption{Inference results for $X^1$, $X^2$, $X^3$, $X^4$, $X^5$. $x^1_{25}$ was hidden and all other time slices of $X^1$ are observed. We observe that the inference results are the same as for the case of a fully observed $X^1$ since only a single solution is possible.}
\label{fig:hdyn2_clean_X5X4X3X2X1_prediction1}
\end{figure}

Figure~\ref{fig:hdyn2_clean_X5X4X3X2X1_prediction1} shows the inference results for $X^1$, $X^2$, $X^3$, $X^4$, $X^5$. We observe that the inferred values for $x^1_{25}$ are the same for the case of fully observed $X^1$. This makes sense since the the transition model in Figure~\ref{fig:targetTracking1} specifies that target state 9 ($x^4_{25}$) must follow target state 8 ($x^4_{24}$) and a pattern shift is not allowed during this transition (i.e., $X^2$ must remain in the same state). Thus, given the observed time slices of $X^1$ before and after 25, there is only one possible configuration of variable values for time slice 25 that satisfies the model, and our inference algorithm successfully solved for it.

\begin{figure}
\centering
\includegraphics[width=100ex]{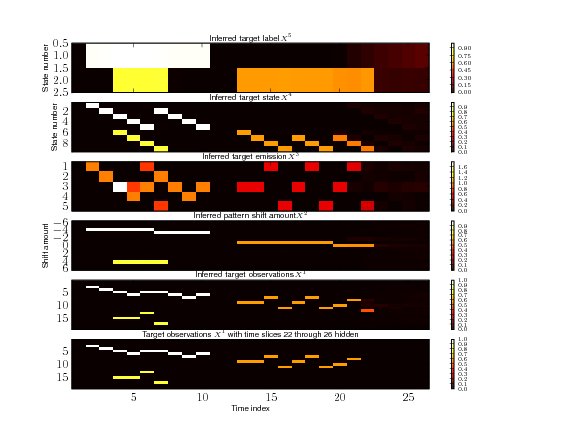}
\caption{Inference results for $X^1$, $X^2$, $X^3$, $X^4$, $X^5$. Time slices 22 through 26 of $X^1$ are hidden. We observe that the infered values for the final 5 time slices now correspond to a superposition of states since multiple solutions are possible.}
\label{fig:hdyn2_clean_X5X4X3X2X1_prediction2}
\end{figure}

We now set the last 5 time slices of $X^1$ ($x^1_{22}, x^1_{23}, x^1_{24}, x^1_{25}, x^1_{26}$) to be hidden and run the inference algorithm. Figure~\ref{fig:hdyn2_clean_X5X4X3X2X1_prediction2} shows the corresponding inference results. We observe that the infered values for the final 5 time slices now correspond to a superposition of states since multiple solutions are possible.

In performing these experiments, we observed that the inference procedure appears to be quite robust, since for the noiseless case, complete convergence (to a very small RMSE) was achieved on every run that we performed. Even in the noisy case, the noise in the inferred results appeared to increase gradually in proportion to the noise in the observation sequence. A possible drawback is that a large number of iterations were needed in order for the algorithm to converge to a solution. We have not yet explored how the number of iterations to convergence varies with network size or graphical structure.


\section{Sparse hierarchical sequential data model}
\label{sec:hierarchicalSeqDecomp}

In this section we consider a multilevel hierarchical DPFN for sequential data in which the activations required in order to explain a given sequence (or non-negative superposition of sequences) become more sparse we we ascend the hierarchy. We then present learning and inference results on magnitude spectrogram data.

\subsection{Model}
\label{sec:sparseHierarchicalModel}

In the sequential data models presented in the previous sections, we modeled $x_t$ as being simultaneously representable as a both a linear function of parent $h_{t-1}$ and another linear function of parent $h_t$. Thus, a nonzero value of the child node $x_t$ implied nonzero values for all of its parents and vice versa. In this section, we consider an alternative network structure for sequential data in which the value of a child node $x_t$ is modeled as sum of linear functions of multiple parents. Under this representation, a nonzero value of a child node implies that at least one parent node must also have a nonzero value and vice versa. Thus, a nonzero value of a single parent node can explain a sequence of possibly several nonzero child node values.

\begin{figure}
\centering
\includegraphics[width=60ex]{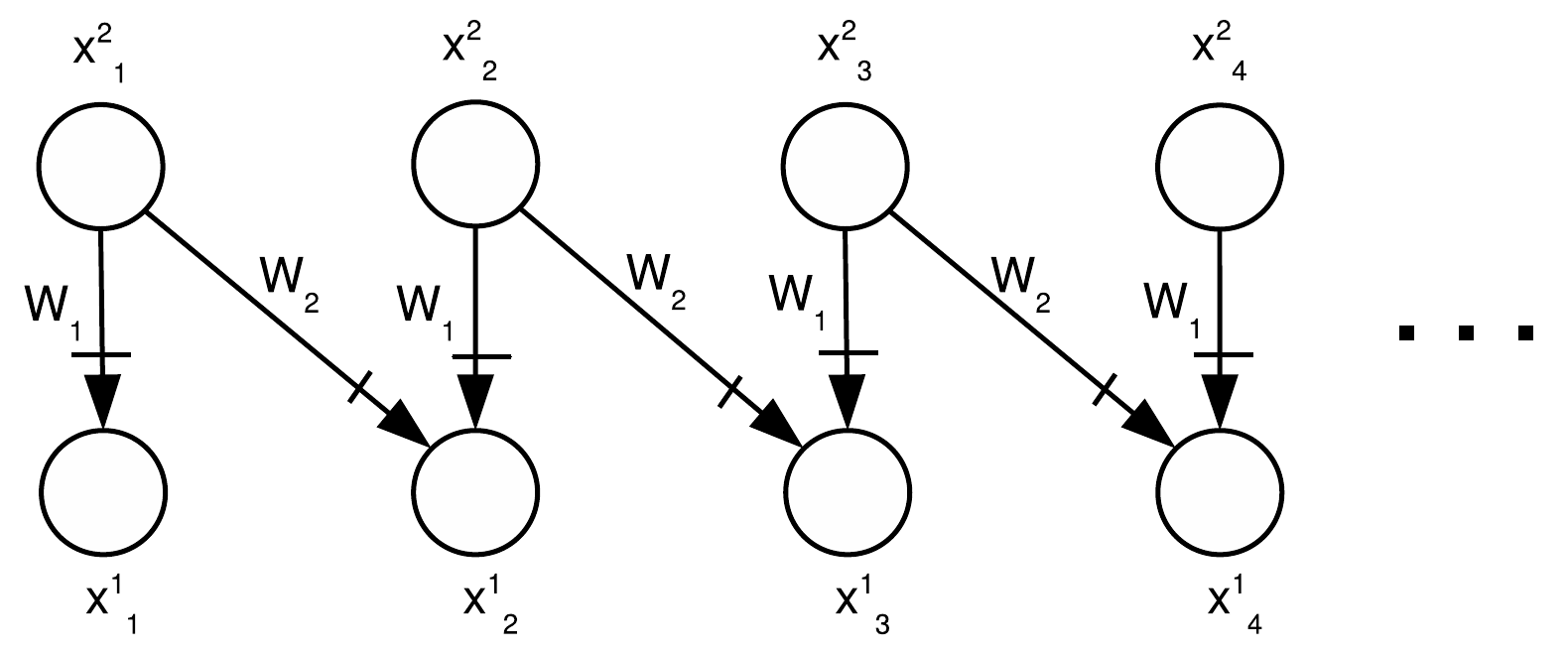}
\caption{A 2-level DPFN for a sparse hierarchical sequence model.}
\label{fig:1hiddenLayerRep2}
\end{figure}

Figure~\ref{fig:1hiddenLayerRep2} shows a DPFN with two layers. We note that Non-Negative Matrix Factor Deconvolution \cite{NMFParis} appears to be a special case of this network.  The variables are non-negative vectors such that $x^i_t \in \mathbb{R}^{M_i}$.  Typically, $\{x^1_t\}$ would correspond to the observed data sequence and $\{x^2_t\}$ would be hidden. A node $x^1_t$ with parents $x^2_{t-1}$ and $x^2_t$ corresponds to the factorization:

\begin{align}
x^1_t =& W_1 x^2_t + W_2 x^2_{t-1} \notag \\
=& \left[ \begin{array}{cc} W_1 & W_2 \end{array} \right] \left[ \begin{array}{c} x^2_t\\
x^2_{t-1} \end{array} \right] 
\end{align}

For time slices $x^1_1 \dots x^1_T$ we then have:

\begin{align}
\left[ \begin{array}{ccccc} x^1_1 & x^1_2 & x^1_3 & \dots & x^1_T \end{array} \right] = \left[ \begin{array}{cc} W_1 & W_2 \end{array} \right] \left[ \begin{array}{ccccc}
x^2_1 & x^2_2 & x^2_3 & \dots & x^2_T \\
x^2_0 & x^2_1 & x^2_2 & \dots & x^2_{T-1} \end{array} \right] 
\end{align}

where we note that $x^2_0 = 0$. Denoting the right-most matrix above as $X^2_S$, we can then express the factorization more concisely as:

 \begin{align}
 X^1 =&  \left[ \begin{array}{cc} W_1 & W_2 \end{array} \right] X^2_S\\
=& W X^2_S
 \end{align}

Note the duplicated variables in $X^2_S$. Since $x^2_j$ appears in two consecutive columns of $X^2_S$, we see that activating $x^2_j$ will correspond to the $j$'th column of $W_1$ being activated in time slice $j$ and the $j$'th column of $W_2$ being activated in time slice $j+1$. Thus, one can think of corresponding columns of $W_1$ and $W_2$ as comprising a transition basis pair. 

We might consider normalizing $W$ so that all columns sum to 1. Alternatively, we could consider normalizing $W$ so that the sum of the corresponding columns of all the $W_j$ sum to 1. That is, we can view the matrix consisting of the j'th column of each consecutive $W_i$ as a basis sequence.

\begin{figure}
\centering
\includegraphics[width=60ex]{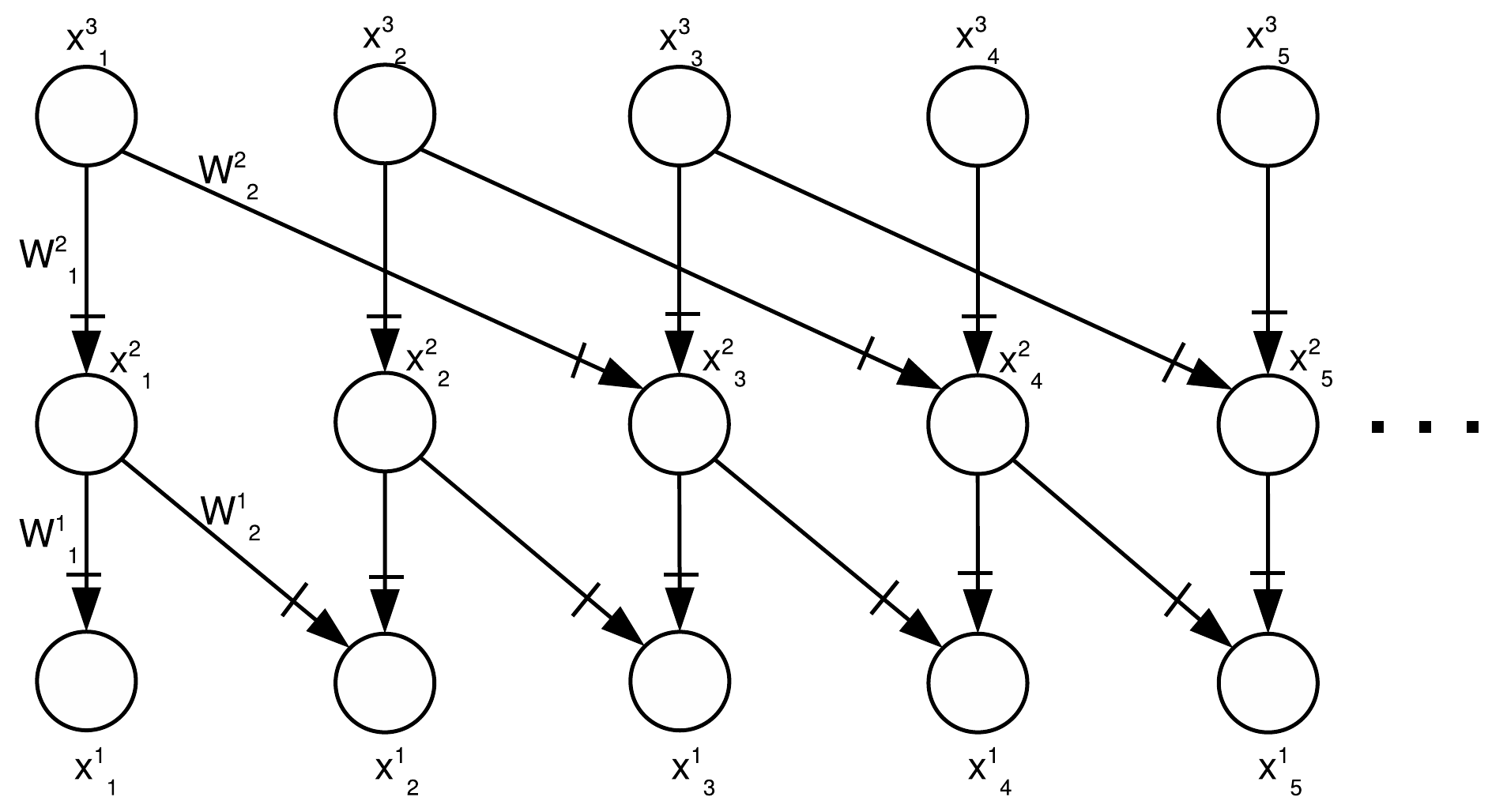}
\caption{A 3-level DPFN for a sparse hierarchical sequence model. The first five time slices are shown.}
\label{fig:2hiddenLayerRep2V2}
\end{figure}

Figure~\ref{fig:2hiddenLayerRep2V2} shows a 3-level network. A sequence of nonzero values $x^1_t, x^1_{t+1}, x^1_{t+2}, x^1_{t+3}$ in level 1 can be represented by  two nonzero values $x^2_t, x^2_{t+2}$ in level 2, and by a single nonzero value of $x^3_t$ in level 3. That is, a single nonzero value $x^{i+1}_t$ in level $i+1$ represents a pair of consecutive values in level $i$. If viewed in a generative setting, a single nonzero level 3 variable $x^1_1$ can cause the level 2 variables $x^2_1, x^2_3$ to be nonzero, which in turn can cause the level 1 variables $x^1_1, x^1_2, x^1_3, x^1_4$ to be nonzero.

Suppose that the level 1 variables $x^1_1, \dots, x^1_T$ are observed. Provided that the parameter matrices $W^1 = \left[ \begin{array}{cc} W^1_1 & W^1_2 \end{array} \right]$ and $W^2 = \left[ \begin{array}{cc} W^2_1 & W^2_2 \end{array} \right]$ contain sufficient basis columns to represent the possible observation pairs $(x^1_t, x^1_{t+1})$, a given sequence can be represented such that only every other level 2 variable is nonzero, and only every 4th level 3 variable is nonzero. Thus, as we ascend the hierarchy, the activations become increasingly sparse.

We represent each level 1 variable in Figure~\ref{fig:2hiddenLayerRep2V2} as:

\begin{align}
x^1_t =& W^1_1 x^2_t + W^1_2 x^2_{t-1} \notag \\
=& \left[ \begin{array}{cc} W^1_1 & W^1_2 \end{array} \right] \left[ \begin{array}{c} x^2_t\\
x^2_{t-1} \end{array} \right] 
\end{align}

Likewise, we represent each level 2 variable as:

\begin{align}
x^2_t =& W^2_1 x^3_t + W^2_2 x^3_{t-2} \notag \\
=& \left[ \begin{array}{cc} W^2_1 & W^2_2 \end{array} \right] \left[ \begin{array}{c} x^3_t\\
x^3_{t-2} \end{array} \right] 
\end{align}

For a network with $T$ time slices, we then have the following factorizations:

\begin{align}
\left[ \begin{array}{ccccc} x^1_1 & x^1_2 & x^1_3 & \dots & x^1_T \end{array} \right] = \left[ \begin{array}{cc} W^1_1 & W^1_2 \end{array} \right] \left[ \begin{array}{ccccc}
x^2_1 & x^2_2 & x^2_3 & \dots & x^2_T \\
x^2_0 & x^2_1 & x^2_2 & \dots & x^2_{T-1} \end{array} \right] 
\end{align}

Note that $x^2_0 = 0$.

\begin{align}
\left[ \begin{array}{ccccc} x^2_1 & x^2_2 & x^2_3 & \dots & x^2_T \end{array} \right] = \left[ \begin{array}{cc} W^2_1 & W^2_2 \end{array} \right] \left[ \begin{array}{ccccc}
x^3_1 & x^3_2 & x^3_3 & \dots & x^3_T \\
x^3_{-1} & x^3_0 & x^3_1 & \dots & x^3_{T-2} \end{array} \right] 
\end{align}

Note that $x^3_{-1}  = x^3_0 = 0$.

The network in Figure~\ref{fig:2hiddenLayerRep2V2} extends immediately to networks with an arbitrary number of levels, arbitrary numbers of parent variables, and arbitrary separation (in time slices) between consecutive child nodes in any given level. Consider a network with $L$ levels and such that the variables $x^{i+1}_t$ in level $i+1$ each have $p$ child nodes with a separation of $q$ time slices between nodes. Let $x^{i+1}_t$ at each time slice $t$ be an $M_{i+1}$ dimensional column vector and let $x^i_t$ at each time slice $t$ be $M_i$ dimensional column vector. For each time slice $t$ and $i \in \{1, \dots, L-1\}$, $x^i_t$ is then expressed by the factorization:

\begin{align}
x^i_t =& W^i_1 x^{i+1}_t + W^i_2 x^{i+1}_{t-q} + W^i_3 x^{i+1}_{t - 2 q} + \dots + W^i_p x^{i+1}_{t- (p-1) q}\\
=& \left[ \begin{array}{ccccc} W^i_1 & W^i_2 & W^i_3 & \dots & W^i_p \end{array} \right] \left[ \begin{array}{c} x^{i+1}_t\\
x^{i+1}_{t-q}\\
x^{i+1}_{t - 2 q}\\
\vdots\\
x^{i+1}_{t- (p-1) q} \notag \\
 \end{array} \right] 
\end{align}

For a network with $T$ time slices, we then have the following factorization equation for each $i \in \{1, \dots, L-1\}$, $x^i_t$:

\begin{align}
\left[ \begin{array}{ccccc} x^i_1 & x^i_2 & x^i_3 & \dots & x^i_T \end{array} \right] = \left[ \begin{array}{ccccc} W^i_1 & W^i_2 & W^i_3 & \dots & W^i_p \end{array} \right] \left[ \begin{array}{ccccc}
x^{i+1}_1 & x^{i+1}_2 & x^{i+1}_3 & \dots & x^{i+1}_T \\
x^{i+1}_{1-q} & x^{i+1}_{2-q} & x^{i+1}_{3-q} & \dots & x^{i+1}_{T-q} \\
x^{i+1}_{1-2 q} & x^{i+1}_{2-2 q} & x^{i+1}_{3-2 q} & \dots & x^{i+1}_{T-2 q}\\
\vdots & \vdots & \vdots & \vdots & \vdots \\
x^{i+1}_{1-(p-1) q} & x^{i+1}_{2-(p-1) q} & x^{i+1}_{3-(p-1) q} & \dots & x^{i+1}_{T-(p-1) q} \\
 \end{array} \right] 
\end{align}

All $x^{i+1}_j$ such that $j < 1$ are zero-valued.

Letting 
\begin{align}
X^i = \left[ \begin{array}{ccccc} x^i_1 & x^i_2 & x^i_3 & \dots & x^i_T \end{array} \right]
\end{align}

\begin{align}
W^i = \left[ \begin{array}{ccccc} W^i_1 & W^i_2 & W^i_3 & \dots & W^i_p \end{array} \right]
\end{align}

\begin{align}
X^{i+1}_S = \left[ \begin{array}{ccccc}
x^{i+1}_1 & x^{i+1}_2 & x^{i+1}_3 & \dots & x^{i+1}_T \\
x^{i+1}_{1-q} & x^{i+1}_{2-q} & x^{i+1}_{3-q} & \dots & x^{i+1}_{T-q} \\
x^{i+1}_{1-2 q} & x^{i+1}_{2-2 q} & x^{i+1}_{3-2 q} & \dots & x^{i+1}_{T-2 q}\\
\vdots & \vdots & \vdots & \vdots & \vdots \\
x^{i+1}_{1-(p-1) q} & x^{i+1}_{2-(p-1) q} & x^{i+1}_{3-(p-1) q} & \dots & x^{i+1}_{T-(p-1) q}\\
\end{array} \right]
\end{align}
 
we can express the factorization more concisely as:
 
 \begin{align}
\label{eq:sparse_hierarchical_fact}
 X^i = W^i X^{i+1}_S
 \end{align}

Note that $X^i$ has size $M_i$ by $T$. Each $W^i_j$ has $M_{i+1}$ columns. The matrix $W^i$ has size $M_i$ by $ M_{i+1}$. The matrix $X^{i+1}_S$ has size $p M_{i+1}$ by $T$.
For example, a network with $L = 2$ levels, $p = 3$ children and $q = 2$ seperation will correspond to the factorization:

\begin{align}
\left[ \begin{array}{ccccccccc} x^1_1 & x^1_2 & x^1_3 & x^1_4 & x^1_5 & x^1_6 & x^1_7 & \dots & x^1_T \end{array} \right] = \left[ \begin{array}{ccc} W^1_1 & W^1_2 & W^1_3 \end{array} \right] \left[ \begin{array}{ccccccccc}
x^2_1 & x^2_2 & x^2_3 & x^2_4 & x^2_5 & x^2_6 & x^2_7 & \dots & x^2_T \\
0 & 0 & x^2_1 & x^2_2 & x^2_3 & x^2_4 & x^2_5 & \dots & x^2_{T-2} \\
0 & 0 & 0 & 0 & x^2_1 & x^2_2 & x^2_3 & \dots & x^2_{T-4} \\
\end{array} \right] 
\end{align}

A learning and inference algorithm for a network with $L$ levels can be obtained by applying Algorithm~\ref{alg:inference_learning1} to the system of factorizations equations (\ref{eq:sparse_hierarchical_fact}). Appendix~\ref{sec:sparse_hierarchical_inf_alg} shows the pseudocode for the algorithm.

\subsection{Empirical results}

\begin{figure}
\centering
\includegraphics[width=100ex]{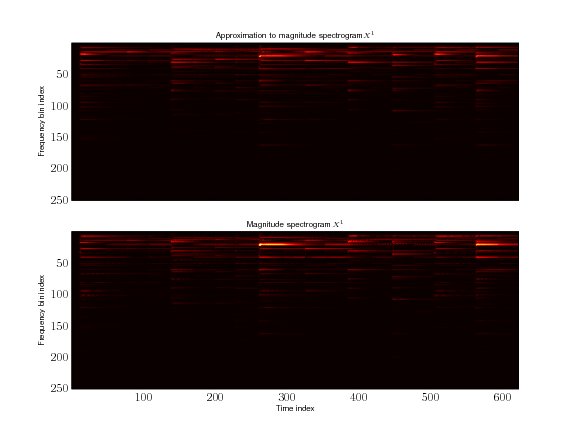}
\caption{Actual vs approximated spectrogram $X^1$ for the 0 sparseness case.}
\label{fig:actual_vs_inferred_spectrogram_0sparseness}
\end{figure}

We now present learning and inference results on magnitude spectrogram data. We let the lowest level variables $X^1$ correspond to a magnitude spectrogram such that $x^1_t$ represents time slice $t$ of a spectrogram. For this experiment, we use a 15 second audio clip from Beethoven's Piano Sonata No. 8 in C minor, op. 13, Adagio cantabile, performed by the author. The corresponding spectrogram, with 512 frequency bins and 622 time slices (41 time slices per second) is shown in the bottom plot of Figure~\ref{fig:actual_vs_inferred_spectrogram_0sparseness}.

We make use of a DPFN with the following network parameters: We use a 4-level network with $X^1$ observed and $X^2$, $X^3$, and $X^4$ hidden. Let $x^2_t \in \mathbb{R}^{50}$ have $p^2 = 4$ child nodes and a separation of $q^2 = 1$ time slice. Let $x^3_t \in \mathbb{R}^{40}$ have $p^3 = 4$ child nodes and a separation of $q^3 = 4$ time slices. Let $x^4_t \in \mathbb{R}^{40}$ have $p^4 = 4$ child nodes and a separation of $q^4 = 16$ time slices.

We use the learning and inference algorithm described in Appendix~\ref{sec:sparse_hierarchical_inf_alg}, which is a special case of the general algorithm described in Algorithm~\ref{alg:inference_learning1}. We start by setting $X^1$ to the spectrogram data values and initializing all other variables and model parameters $W^i$ to random positive values. We have observed that model convergence can sometimes be improved by performing the parameter learning update steps more frequently on parameter matrices that are closer to the observed data in the graphical model. In this case, we perform $W^1$ more frequently than $W^2$, and perform $W^2$ updates more frequently then $W^3$ and so on. For each level, $W^i$ was normalized so that the sum of the corresponding columns of all the $W^i_j$ sum to 1.

\begin{figure}
\centering
\includegraphics[width=100ex]{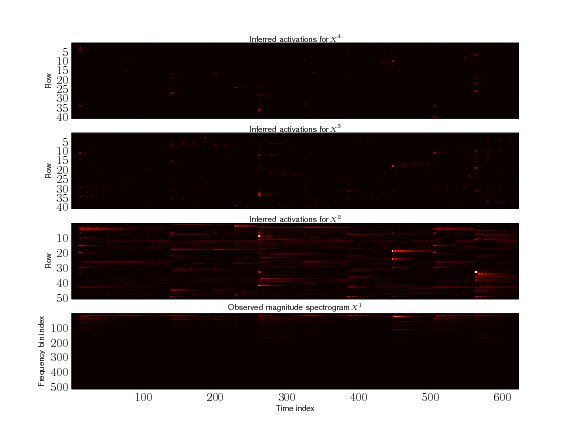}
\caption{Observed spectrogram $X^1$ and inferred activations for $X^2$, $X^3$, and $X^4$. Sparseness was 0.}
\label{fig:inferred_spectrogram_0sparseness}
\end{figure}

Figure~\ref{fig:inferred_spectrogram_0sparseness} Shows the observed spectrogram $X^1$ and inferred activations for $X^2$, $X^3$, and $X^4$ after running the learning and inference algorithm for 800 iterations. Note that the activations become progressively more sparse as we ascend the hierarchy from $X^2$ to $X^4$, even though sparse NMF update steps were not used in the inference and learning algorithm. 

Figure~\ref{fig:actual_vs_inferred_spectrogram_0sparseness} shows the actual observed spectrogram as well as the approximation to the spectrogram obtained by using only the inferred top-level activations $X^4$ and propagating them downward to form an approximation to $X^1$. The resulting RMSE was 0.07. Only the first 250 frequency bins are shown since the higher frequency bins have negligible energy.

\begin{figure}
\centering
\includegraphics[width=100ex]{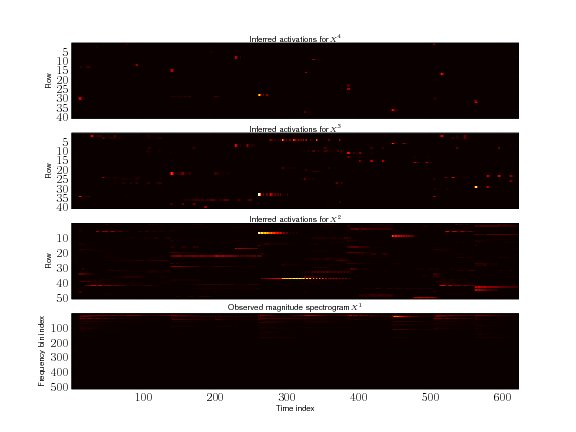}
\caption{Observed spectrogram $X^1$ and inferred activations for $X^2$, $X^3$, and $X^4$. Sparseness was gradually increased to a value of 0.2 at 800 iterations.}
\label{fig:inferred_spectrogram_sparse}
\end{figure}

Figure~\ref{fig:inferred_spectrogram_sparse} Shows the observed spectrogram $X^1$ and inferred activations for $X^2$, $X^3$, and $X^4$ for the case where sparse NMF updates were used in the inference and learning algorithm. Here, we ran the inference and learning algorithm for 400 iterations with 0 sparseness, and then increased the nsNMF sparseness parameter gradually from 0 to 0.2 at 800 iterations.

\begin{figure}
\centering
\includegraphics[width=100ex]{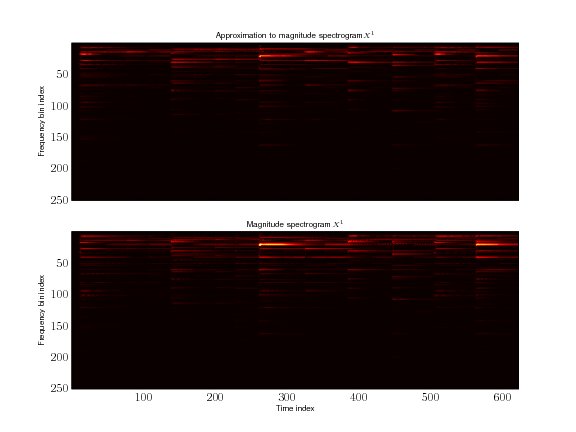}
\caption{Actual vs approximated spectrogram $X^1$ for the 0 sparseness case.}
\label{fig:actual_vs_inferred_spectrogram_sparse}
\end{figure}

Figure~\ref{fig:actual_vs_inferred_spectrogram_sparse} shows the actual observed spectrogram as well as the approximation to the spectrogram obtained by using only the inferred top-level activations $X^4$ and propagating them downward to form an approximation to $X^1$. The resulting RMSE was 0.2. Only the first 250 frequency bins are shown since the higher frequency bins have negligible energy. This is for gradually increasing sparse NMF. We observe that the results using sparse NMF updates appear slightly more sparse than those using the non-sparse NMF updates. However, the reconstruction error using sparse updates is also higher.

We feel that it may be interesting to consider applying this or a similar DPFN to problems where a sparse hierarchical representation of sequential data could be useful, such as compression, noise reduction, and transcription.

\section{A DPFN for language modeling}
\label{sec:main_language_model}

In this section we propose a data model for the sequence of words in a text document, but do not present any corresponding empirical results. In this model, we perform dimensionality reduction by extracting low dimensional feature vectors corresponding to each distinct word. We then make use of a transition model for the feature vectors. By performing dimensionality reduction to extract features, we can use a much more compact transition model than would be possible by operating directly on the words themselves. The idea of extracting low-dimensional word features in order to model sequential word data in text documents and make predictions of future words was used in \cite{bengioy-journal}.

\begin{figure}
\centering
\includegraphics[width=60ex]{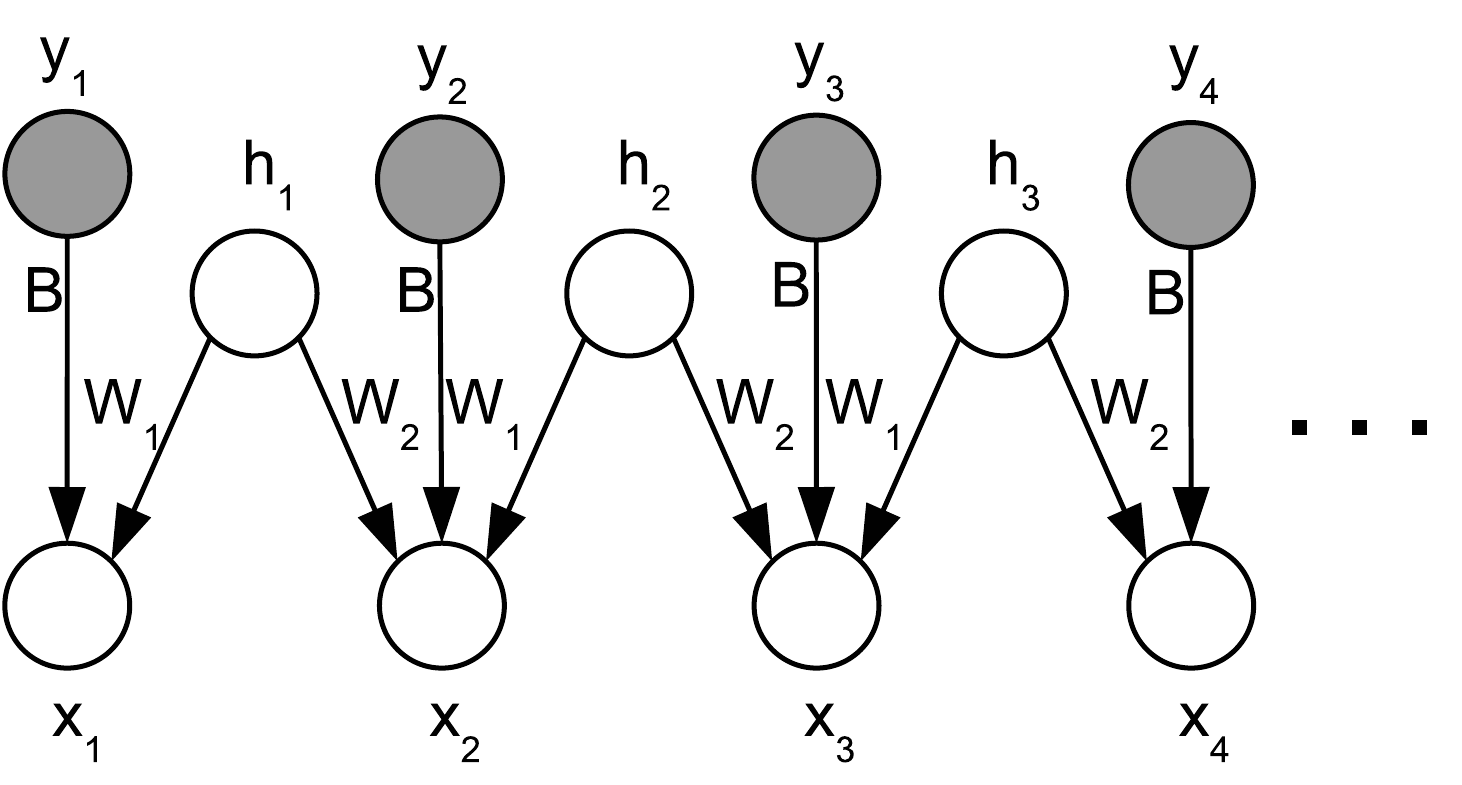}
\caption{A DPFN representing a transition model over word features. The observed $y_t$ denotes a word vector, $x_t$ denotes a word feature vector, and $h_t$ denotes a word feature transition. The first four time slices are shown.}
\label{fig:wordFeatureMode}
\end{figure}

Figure~\ref{fig:wordFeatureMode} shows the graphical model. In this model, the observed \{$y_t : t = 1, \dots, T$\} correspond to the sequence of words that appear in some text document. Suppose our vocabulary size is $N$ words. Then let $y_t$ denote an $N$ dimensional vector such that the $t$'th word $i \in \{1, \dots, N\}$ in the document is represented by setting the $i$'th component of $y_t$ to 1 and all other components to zero.  This model can also be used for prediction of future words, given a sequence of words up to the present $t_p$ by simply considering the word vector nodes for future time slices \{$y_{t_p}, y_{{t_p}+1}, \dots$\} to be hidden.

We perform dimensionality reduction to extract low-dimensional word features $x_t$ by left multiplying each word vector $y_t$ by a \emph{word feature matrix} $B$ so that we have:

\begin{align}
x_t = B y_t
\end{align}

The non-negative matrix $B$ then represents the word feature matrix, where the $i,j$'th component represents the amount of feature $i$ contained by word $j$. One might consider normalizing the columns of $B$ to have unit column sum. Letting $Y = \left[ \begin{array}{ccccc} y_1 & y_2 & y_3 & \dots & y_T \end{array} \right]$, we then have the following matrix factorization for the word features:

\begin{align}
X = B Y
\label{eqn:word_feature_factorization}
\end{align}

The factorization equations that describe this model are exactly the same as those for the model in Figure~\ref{fig:1layerDynamic} with the additional above factorization for the word features in terms of the word vectors. Thus, the complete system of factorizations for this model can be expressed as the two matrix factorizations (\ref{eqn:key_factorization}) and (\ref{eqn:word_feature_factorization}).

Suupose, for example, that the word ``cat'' has the features ``animal'' and ``has claws.'' The feature transition model (parametrized by $W_1$, $W_2$) could possibly have distinct basis transitions containing the ``animal'' and ``has claws'' features. For a word such as ``cat'' that has both features, both basis transitions could be simultaneously activated due to the non-negative superposition property (Section~\ref{sec:model_specification}). Thus, a given word sequence could be explained such that each word has possibly multiple features, and that multiple distinct basis transitions in the feature transition model could simultaneously be activated to explain the words in a sequence as an additive combination of basis feature transitions. One could then think of the word feature transition model as operating on the individual word features independently and simultaneously such that the observed or inferred word sequence is consistent with them. We feel that such an additive factored language model might be a powerful way of modeling word and/or character sequences. However preliminary empirical results suggest that additional sparseness and/or orthogonality constraints might need to be imposed in order to obtain meaningful solutions.

\section{Discussion}

We have presented a framework for modeling non-negative data that retains the linear representation of NMF, while providing for more structured representations using (non-probabilistic) graphical models. The graphical model provides a visual representation of the factored correlation constraints between variables. We have presented algorithms for performing the inference and learning that leverage existing non-negative matrix factorization (NMF) algorithms. Since the algorithms are inherently parallel, we expect that optimized implementations for hardware such as multi-core CPUs and GPUs will be possible. These algorithm have the advantage of being quite straightforward to implement, even for relatively complex networks structures. We expect that they can be understood and implemented by anyone with a basic understanding of computer programming, matrix algebra, and graph theory, and require no background in probability theory. 

We presented empirical results for several example networks that illustrated the robustness of the inference and learning algorithms. Our goal was to provide a sufficient variety of model structures so that the interested reader will hopefully be able to extend this work and apply it to interesting and practical applications. We observed that in all models in which noiseless synthetic data sets were used, learning and/or inference converged to an essentially exact factorization, corresponding to a correct solution for the hidden variables. In the case of additive noise, the quality of the solutions tended to degrade gradually as the noise level was increased. In particular, in Section~\ref{sec:learn_tran_model_from_data} we observed the somewhat surprising result that it was possible to learn an underlying transition model even when the training sequence contained only an additive mixture of realizations of the underlying model. In our magnitude spectrogram modeling example, we observed that the inference and learning solution appeared to correspond to a hierarchical sparse decomposition, with the solution sparseness increasing as we ascend the model hierarchy. As with existing NMF algorithms, however, there is no guarantee that convergence to a global optimum for any particular cost function will occur. We did observe good convergence properties in our empirical results, however. We also proposed a PFN for language modeling and plan to experiment with related models as future research. 

It is still unknown how well models using this framework will perform on hard real-world problems such as speech recognition, language modeling, and music transcription, and this will be an interesting area of future research. It could also be interesting to explore the use of dynamic networks that are directional in time, such as DPFNs with a similar state space model to an HMM or HHMM, for example.  Source code for implementing the models in this paper will be made available by the author.

\section{Acknowledgments}

I would like to thank Michael I. Jordan for reviewing a draft of this paper and offering helpful suggestions.

\appendix

\section{Algorithms for Non-negative Matrix Factorization}
\label{appendix_nmf}

In this section we present the non-negative matrix factorization (NMF) algorithms that were used to implement the examples in this paper. NMF is a method for constructing an approximate factorization of a matrix subject to non-negativity constraints. It was originally proposed by \cite{paatero_1994} as \emph{positive matrix factorization}. An observed non-negative $M$ by $T$ matrix $X \in \mathbb{R}^{\geq 0, M \text{x} T}$ is factored as :

\begin{align}
X \approx W H
\end{align}

 where the $M$ by $R$ matrix $W \in \mathbb{R}^{\geq 0, M \text{x} R}$ and the $R$ by $T$ matrix $H \in \mathbb{R}^{\geq 0, R \text{x} T}$ are the non-negative factor matrices.

NMF can be viewed as a linear basis decomposition in which the columns of $X$ represent the observed variables, the columns of $W$ represent the basis vectors, and the columns of $H$ represent the corresponding encoding variables. Let $x_i$, $w_i$, and $h_i$ denote the $i$'th columns of $X$, $W$, and $H$, respectively. Let $h_i(j)$ denote the $j$'th component in the column vector $h_i$ (i.e., $H_{j i}$). Then we have :

\begin{align}
x_i \approx& W h_i \notag \\
=& \left[ \begin{array}{cccc} w_1 w_2 \dots w_R\end{array} \right] h_i \label{eqn:nmf_linear_comb} \\
=& w_1 h_i(1) + w_2 h_i(2) + \dots + w_R h_i(R) \notag
\end{align}

Each column $x_i, i=1 \dots T$ represents an observed variable that is modeled as a non-negative linear combination of the $R$ non-negative basis vectors in $W$. The activations of the hidden variable $h_i$ specify the columns of $W$ that will combine additively to form $x_i$. Since only additive combinations are possible, NMF can be interpreted as a parts-based representation.

\cite{Lee_seung} developed simple and robust multiplicative update rules for minimizing the reconstruction error between $X$ and  $W H$ using a Frobenius norm cost function and also a generalized KL divergence cost function given by:

\begin{align}
D(X || W H) = \sum_{i,j} (X_{i j} log \frac{X_{i j}}{(W H)_{i j}} - X_{i j} + (W H)_{i j})
\label{eq:kl_div}
\end{align}

The NMF factorization is then given by:
\begin{align*}
\min_{W, H}
D(X || W H) \\
\text{s.t. } W, H \geq 0
\end{align*}

From the above cost function, \cite{Lee_seung} derived the following multiplicative update rules which are guaranteed to converge to a local minimum of the KL divergence cost function $D(X || H)$:
\begin{eqnarray}
\label{eqn:lee_seung_HUpdate}
H &\leftarrow& H \otimes \frac{W^T \frac{X}{W H}} { W^T 1_{X}} \label{eqn:HUpdate} \\
\label{eqn:lee_seung_WUpdate}
W &\leftarrow& W \otimes \frac{ \frac{X}{W H} H^T} {1_X H^T} \label{eqn:WUpdate}
\end{eqnarray}

where $X \otimes Y$ denotes component-wise multiplication, the division is component-wise, and $1_X$ is a matrix of ones of the same size as $X$. The $W$ and $H$ are typically initialized to random positive values. We will refer to the $W$ update step as the \emph{learning NMF update} or \emph{left NMF update} step and we will refer to the $H$ update step as the \emph{inference NMF update} or \emph{right NMF update} step in this paper. In our data sets, we have observed that both the Frobenius norm and the KL divergence cost functions tend to produce qualitatively similar results. However, we have observed that convergence tends to be faster when using the KL divergence cost function. We only consider the KL divergence cost function in this paper.

In an implementation of these update rules, one must be careful to avoid division be zero. We prevent division by zero by adding a small positive value $\epsilon$ to the numerator and denominator before performing the division operation, where $\epsilon$ is small compared to the largest value in the numerator or denominator matrices. In our empirical results sections, we used $\epsilon = 0.00001$. This leads to the following update rules:

\begin{eqnarray}
\label{eqn:right_nmf_update}
H &\leftarrow& H \otimes \frac{W^T \frac{X + \epsilon 1_X}{W H + \epsilon 1_X} + \epsilon 1_H} { W^T 1_{X} + \epsilon 1_H} \label{eqn:HUpdateEpsilon} \\
\label{eqn:left_nmf_update}
W &\leftarrow& W \otimes \frac{ \frac{X + \epsilon 1_X}{W H + \epsilon 1_X} H^T + \epsilon 1_W} {1_X H^T + \epsilon 1_W} \label{eqn:WUpdateEpsilon}
\end{eqnarray}

\section{Sparse NMF}
\label{appendix_sparse_nmf}

Classical NMF tends to produce somewhat sparse solutions already. However, there may be cases in which we wish to control the amount of sparseness. There are many algorithms for performing NMF with sparse factorizations, such as \cite{hoyer_sparse_NMF}, \cite{cichockiNMF}, \cite{nsNMF2006}. We have implemented the nonSmooth NMF (nsNMF) algorithm \cite{nsNMF2006} and used it where we want sparse solutions since it seems to perform well and is straightforward to implement. However, it is possible that other sparse NMF algorithms might also perform similarly well.

The nsNMF method uses the factorization

\begin{eqnarray}
X = W S H \label{eqn:nsNMF}
\end{eqnarray}

where $S \in \mathbb{R}^{R \text{ x } R}$ is a symmetric smoothing matrix defined as

\begin{eqnarray}
S = (1 - \theta) I_S + \frac{\theta}{R} 1_S
\end{eqnarray}

where $I_S \in \mathbb{R}^{R \text{x} R}$ is an identity matrix and $a_S \in \mathbb{R}^{R \text{ x } R}$ is a matrix of ones. The sparseness parameter $\theta$ satisfies $0 \leq \theta \leq 1$. As described in \cite{nsNMF2006}, the nsNMF update rules can use the update rules from Equations (\ref{eqn:lee_seung_HUpdate}),(\ref{eqn:lee_seung_WUpdate}) with the following substitutions: In the $W$ update rule, replace $H$ with $S H$. In the $H$ update rule, replace $W$ with $W S$.  However, we used the update rules from Equations (\ref{eqn:right_nmf_update}),(\ref{eqn:left_nmf_update}) instead for numerical performance reasons. In \cite{nsNMF2006}, it was proposed to normalize the columns of $W$ to sum to 1. However, in some cases we have observed that this additional constraint can actually result in less sparse solutions, since this constraint disallows zero-valued columns of $W$. Note that nsNMF reduces to the standard NMF update rules when $\theta  = 0$.

\section{Learning and inference algorithm 2}
\label{appendix_inf_learn_2}

Algorithm~\ref{alg:inference_learning2} and the corresponding procedure in Algorithm~\ref{alg:averageLevel} show the pseudocode for a slightly modified inference and learning algorithm for a general PFN that uses a slightly different way of computing the variable averages than was used in Algorithm~\ref{alg:inference_learning1}. In computing the mean values of the variables, the new mean for the model variables is computed from the the complete set of local variables. Thus, after performing the value propagation step, the updated values of all lower level variables in the model are a function of both the top level variables and the values from the previous inference step. In this algorithm, when a local variable $v^j_k$ is updated, the new mean value is computed over all of the $v^j_k$ associated with $x_{f(j,k)}$, whereas in the previous algorithm (\ref{alg:inference_learning1}), the new mean value is computed over possibly only a proper subset of the $v^j_k$ associated with $x_{f(j,k)}$ (i.e., over the $v^j_k$ associated with the current level $l$). We have observed good convergence using this algorithm, although it can take longer to converge than algorithm 1. We did not perform any extensive comparison with algorithm 1.

\begin{algorithm}
\caption{Perform inference and learning}
\label{alg:inference_learning2}
\begin{tabbing}
Initialize hidden variables to random positive values \\
// Main loop \\
{\bf repe}\={\bf at}  \\
\> // Bottom-to-top inference and learning \\
\> {\bf for} \= $l$ = 1 to $L -1$ \\
\> \> upStep($l$) \\
\> \> averageLevel($l$) \\
\> {\bf end} \\
\> // Top-to-bottom value propagation \\
\> {\bf for} $l$ = $L -1$ downto 1 \\
\> \> downStep($l$) \\
\> \> averageLevel($l$) \\
\> {\bf end} \\
{\bf until} convergence
\end{tabbing}
\end{algorithm}

\begin{algorithm}
\caption{averageLevel() procedure}
\label{alg:averageLevel}
\begin{tabbing}
// Let $X_l$ denote the set of model variables \{$x^l_1, x^l_2, \dots, x^l_{LevelCount_l}$\} corresponding to the level $l$ nodes.\\
// Let $FactorSystem_l$ denote the subset of the factorization equations from (\ref{eq:factorization_equation333})\\
// \{$eq_j: $ such that $v^j_0$ in $eq_j$ corresponds to an $x^l_i \in X_l$\}.  \\
// Let $duplicationSet(i,l)$ = \{$(j,k) : eq_j \in FactorSystem$ and $v^j_k$ corresponds to $x^l_i$\}. \\
//  Let $duplicationCount(i,l)$ = $|duplicationSet(i,l)|$.\\
{\bf aver}\={\bf ageLevel}($l$) \\
\> {\bf for} \= $i$ = 1 to $LevelCount_l$ \\
\> \> // For each (child) variable $x^l_i \in X_l$ \\
\> \> {\bf if} \=$x^l_i$ is hidden \\
\> \> \> meanValue = $\frac{1}{duplicationCount(i,l)} \sum_{j \in duplicationSet(i,l)} v^j_k$ \\
\> \> \> {\bf for} \= {\bf each} $(j,k) \in duplicationSet(i,l)$ \\
\> \> \> \> $v^j_k \gets meanValue$ \\
\> \> \> {\bf end} \\
\> \> \> $x^l_i \gets meanValue$\\
\> \> {\bf else if} $x^l_i$ is observed \\
\> \> \> {\bf for each} $(j,k) \in duplicationSet(i,l)$ \\
\> \> \> \> $v^j_k \gets x^l_i $ \\
\> \> \> {\bf end} \\
\> \> {\bf end} \\
\> {\bf end} \\
{\bf end} 
\end{tabbing}
\end{algorithm}

\section{Learning and inference algorithm for the network in Section~\ref{sec:fact_state_tran_model}}
\label{sec:simpleDyn1_inf_learn}

We now present a learning and inference algorithm for the network in Figure~\ref{fig:1layerDynamic}. The model corresponds to the variables \{$x_t : t = 1, \dots, T$\} and \{$h_t, t = 1, \dots, T-1$\}. Let us start by partitioning these variables into a hidden set $X_H$ and an observed set $X_E$. From equation (\ref{eqn:key_factorization}) we see that the $T-1$ vector factorizations that describe the network can be compactly represented as a single matrix factorization equation.

The learning and inference algorithm from Section~\ref{sec:inference_and_learning} then corresponds to the pseudocode shown in Algorithm~\ref{alg:inference_learning_dynamic1}. We include the learning step in the algorithm, although it can be disabled if $W$. We start by initializing the components of $X$ and $H$ corresponding to $X_E$ to the observed values, and initializing the components corresponding to $X_H$ to random positive values. We then iterate the bottom-to-top inference and learning followed by top-to-bottom value propagation until convergence. The components of $X$ and $H$ can be thought of as the model variables. The components of $X_C$ can be thought of as the local variables $v^j_0$ corresponding to the model variables in $X$. Note from Equation~(\ref{eqn:expanded_key_factorization}) that each model variable $x_t$ except $x_1$ and $x_T$ corresponds to two local variables in $X_C$. The components of $H$ can be thought of as serving as both model variables and local variables since no model variables $h_t$ are duplicated in $H$. If $W$ is to be learned then it is also initialized to random positive values.

\begin{algorithm}
\caption{Perform inference and learning for the network in Figure~\ref{fig:1layerDynamic}}
\label{alg:inference_learning_dynamic1}
\begin{tabbing}
// Main loop \\
{\bf repe}\={\bf at}  \\
\> // Bottom-to-top inference and learning \\
\> upStep($X_C,W,H$) \\
\> averageParents($H$) \\
\> // Top-to-bottom value propogation \\
\> downStep($X_C, W, H$) \\
\> averageChildren($X, X_C$) \\
{\bf until} convergence\\
// Procedure: upStep() \\
{\bf upSt}\={\bf ep}($W, H$) \\
\> {\bf if} \= learning is enabled \\
\> \> Learning update: Using $X_C = W H$, perform a left NMF update on $W$, using, e.g. (\ref{eqn:left_nmf_update}) \\
\> {\bf end} \\
\> Inference update: Using $X_C = W H$, perform a right NMF update on $H$, using, e.g. (\ref{eqn:right_nmf_update}) \\
{\bf end} \\
// Procedure: averageParents() \\
{\bf aver}\={\bf ageParents}($H$) \\
\> Overwrite all observed components in $H$ with the corresponding observed values from $X_E$. \\
{\bf end} \\
// Procedure: downStep() \\
{\bf down}\={\bf step}($X_C, W, H$) \\
\> $X_C \gets W H$ \\
{\bf end} \\
// Procedure: averageChildren() \\
{\bf aver}\={\bf ageChildren}($X, X_C$) \\
\> // Compute the mean value of each pair of components in $X_C$ that correspond to the same \\
\> // hidden model variable in $X$. Overwrite all hidden components in $X$ and $X_C$ with \\
\> // the computed mean values. \\
\> $(X_c, X) \gets meanValues(X_C)$ \\
\> Overwrite all observed components in $X$ and $X_C$ with the corresponding values in $X_E$. \\
{\bf end} \\
\end{tabbing}
\end{algorithm}

\section{Learning and inference algorithm for the network in Section~\ref{sec:dyn2level}}
\label{sec:inf_learn_dyn2level}

We now present a learning and inference algorithm for the network in Figure~\ref{fig:2layerDynamic}. A learning and inference algorithm for this network can be obtained by applying Algorithm~\ref{alg:inference_learning1} to the system of factorizations (\ref{eq:dynamic1_fact1}),(\ref{eq:dynamic1_fact2}), and (\ref{eq:dynamic1_fact3}). The psuedocode is shown in Algorithm~\ref{alg:inference_learning_dyn2level}.

Here, the $averageParents(<matrix>)$ procedure is analogous to the $averageParents()$ procedure of Algorithm~\ref{alg:inference_learning1}. In $averageParents(<matrix>)$, the mean value of each set of local variables in $<matrix>$ corresponding to the same model variable is computed. All hidden components in $<matrix>$ are then updated to the corresponding mean values and all observed components are updated to the corresponding values in $X_E$.

The $averageChildren(<matrix1>,<matrix2>)$ and $averageChildren(<matrix2>)$ procedures are analogous to the $averageChildren()$ procedure of Algorithm~\ref{alg:inference_learning1}. Here, <matrix1> corresponds to the model variables and <matrix2> corresponds to the local variables. The mean value of each set of local variables in $<matrix2>$ corresponding to the same model variable is computed. All hidden components in $<matrix1>$ and  $<matrix2>$ are then updated to the corresponding mean values and all observed components are updated to the corresponding values in $X_E$.

\begin{algorithm}
\caption{Perform inference and learning for the network in Figure~\ref{fig:2layerDynamic}}
\label{alg:inference_learning_dyn2level}
\begin{tabbing}
// Main loop \\
{\bf repe}\={\bf at}  \\
\> // Bottom-to-top inference and learning \\
\> upStep($X^1_C,W^1,H^1$) \\
\> averageParents($H^1$) \\
\> upStep($X^2_C,W^2,H^2$) \\
\> averageParents($H^2$) \\
\> upStep($\left[ \begin{array}{c}
H^2 \\
H^1 \end{array} \right],U,V$) \\
\> averageParents($V$) \\
\> // Top-to-bottom value propogation \\
\> downStep($\left[ \begin{array}{c}
H^2 \\
H^1 \end{array} \right],U,V$) \\
\> averageChildren($\left[ \begin{array}{c}
H^2 \\
H^1 \end{array} \right]$) \\
\> downStep($X^2_C,W^2,H^2$) \\
\> averageChildren($X^2, X^2_C$) \\
\> downStep($X^1_C,W^1,H^1$) \\
\> averageChildren($X^1, X^1_C$) \\
{\bf until} convergence\\
\end{tabbing}
\end{algorithm}

\section{Learning and inference algorithm for the target tracking network in Section~\ref{sec:target_tracking}}
\label{sec:learn_inf_target_tracking}

We now present a learning and inference algorithm for the target tracking network in Figure~\ref{fig:targetTracking1}. A learning and inference algorithm for this network can be obtained by applying Algorithm~\ref{alg:inference_learning1} to the system of factorizations () - (). The pseudocode is shown in Algorithm~\ref{alg:inference_learning_target_tracking}. Note that variables $\{x^1_t\}$ corresponding to $X^1$ are observed or partially observed and all other model variables were hidden in the experimental examples. However, the general form of the inference and learning algorithm remains the same regardless of any particular choice of variable partitioning into $X_H$ and $X_E$.

\begin{algorithm}
\caption{Perform inference and learning for the network in Figure~\ref{fig:2layerDynamic}}
\label{alg:inference_learning_target_tracking}
\begin{tabbing}
// Main loop \\
{\bf repe}\={\bf at}  \\
\> // Bottom-to-top inference and learning \\
\> upStep($\left[ \begin{array}{c}
X^3_{Copy1} \\
X^2 \\
X^1 \end{array} \right], W^1, U^1$) // Factorization~(\ref{eq:target_pattern_translation_coupling}) \\
\> averageParents($U^1$) \\
\> upStep($\left[ \begin{array}{c}
X^4_{Copy1} \\
X^3_{Copy2} \end{array} \right], W^3, U^2$) // Factorization~(\ref{eq:target_state_to_target_emission_coupling}) \\
\> averageParents($U^2$) \\
\> upStep($\left[ \begin{array}{c}
X^5 \\
X^4_{Copy2} \end{array} \right], W^7, U^3$) // Factorization~(\ref{eq:target_label_to_target_state_coupling}) \\
\> averageParents($U^3$) \\
\> upStep($X^2_C, W^2, H^1$) // Factorization~(\ref{eq:pattern_translation_transition_model}) \\
\> averageParents($H^1$) \\
\> upStep($X^4_C, W^6, H^3$) //  Factorization~(\ref{eq:target_state_transition_model}) \\
\> averageParents($H^3$) \\
\> upStep($\left[ \begin{array}{c}
H^2 \\
H^1 \end{array} \right], W^4, V^1$) //  Factorization~(\ref{eq:h2Toh1CouplingTarget}) \\
\> averageParents($V^1$) \\
\> upStep($\left[ \begin{array}{c}
H^3 \\
H^2_{Copy1} \end{array} \right], W^5, V^2$)  //  Factorization~(\ref{eq:h3Toh2CouplingTarget}) \\
\> averageParents($V^2$) \\
\> // Top-to-bottom value propogation \\
\> downStep($\left[ \begin{array}{c}
H^3 \\
H^2_{Copy2} \end{array} \right], W^5, V^2$)  //  Factorization~(\ref{eq:h3Toh2CouplingTarget}) \\
\> averageChildren($H^3$) \\
\> downStep($\left[ \begin{array}{c}
H^2 \\
H^1 \end{array} \right], W^4, V^1$) //  Factorization~(\ref{eq:h2Toh1CouplingTarget}) \\
\> averageChildren($H^1$) \\
\> averageChildren($H^2, H^2_{Copy1}, H^2_{Copy2}$) \\
\> downStep($X^4_C, W^6, H^3$) //  Factorization~(\ref{eq:target_state_transition_model}) \\
\> downStep($\left[ \begin{array}{c}
X^5 \\
X^4_{Copy2} \end{array} \right], W^7, U^3$) // Factorization~(\ref{eq:target_label_to_target_state_coupling}) \\
\> averageChildren($X^5$) \\
\> downStep($\left[ \begin{array}{c}
X^4_{Copy1} \\
X^3_{Copy2} \end{array} \right], W^3, U^2$) // Factorization~(\ref{eq:target_state_to_target_emission_coupling}) \\
\> averageChildren($X^4_C, X^4_{Copy1}, X^4_{Copy2}$) \\
\> downStep($X^2_C, W^2, H^1$) // Factorization~(\ref{eq:pattern_translation_transition_model}) \\
\> downStep($\left[ \begin{array}{c}
X^3_{Copy1} \\
X^2 \\
X^1 \end{array} \right], W^1, U^1$) // Factorization~(\ref{eq:target_pattern_translation_coupling}) \\
\> averageChildren($X^3_{Copy1}, X^3_{Copy2}$) \\
\> averageChildren($X^2 , X^2_C$) \\
\> averageChildren($X^1$) \\
{\bf until} convergence\\
\end{tabbing}
\end{algorithm}

\section{Learning and inference algorithm for the network in Section~\ref{sec:hierarchicalSeqDecomp}}
\label{sec:sparse_hierarchical_inf_alg}

We now present a learning and inference algorithm for a network with $L$ levels in Section~\ref{sec:hierarchicalSeqDecomp}. The network has $L$ levels, corresponding to layers $X^1_t, X^2_t, \dots, X^L_t$. Any values for the number of children nodes $p$ and node separation $q$ are allowed, and can even be distinct for each level. In the example in this paper $X^1$ is observed and all other $X^i$ are hidden, but any partitioning of the model variables into $X_H$ and $X_E$ is allowed.  A learning and inference algorithm for this network can be obtained by applying Algorithm~\ref{alg:inference_learning1} to the system of factorizations (\ref{eq:sparse_hierarchical_fact}). The pseudocode is shown in Algorithm~\ref{alg:inference_learning_sparse_hierarchical_decomp}.

\begin{algorithm}
\caption{Perform inference and learning for an $L$ level network in Section~\ref{sec:hierarchicalSeqDecomp}}
\label{alg:inference_learning_sparse_hierarchical_decomp}
\begin{tabbing}
// Main loop \\
{\bf repe}\={\bf at}  \\
\> // Bottom-to-top inference and learning \\
\> {\bf for} \= $l$ = 1 to $L -1$ \\
\> \> upStep($X^i, W^i, X^{i+1}_S$) // Factorization (\ref{eq:sparse_hierarchical_fact}) \\
\> \> averageParents($X^{i+1}_S, X^{i+1}$) \\
\> {\bf end} \\
\> // Top-to-bottom value propogation \\
\> {\bf for} $l$ = $L -1$ downto 1 \\
\> \> downStep($X^i, W^i, X^{i+1}_S$) // Factorization (\ref{eq:sparse_hierarchical_fact}) \\
\> \> averageChildren($X^i, X^i_S$) \\
\> {\bf end} \\
{\bf until} convergence
\end{tabbing}
\end{algorithm}

\bibliographystyle{spiebib.bst}

\bibliography{vogel_hsl}

\end{document}